\documentclass{article}

\usepackage{PRIMEarxiv}

\usepackage[utf8]{inputenc} %
\usepackage[T1]{fontenc}    %
\usepackage{hyperref}       %
\usepackage{url}            %
\usepackage{booktabs}       %
\usepackage{amsfonts}       %
\usepackage{nicefrac}       %
\usepackage{microtype}      %
\usepackage{lipsum}
\usepackage{fancyhdr}       %
\usepackage{graphicx}       %
\graphicspath{{media/}}     %
\usepackage{amsmath}

\usepackage[authoryear]{natbib}
\bibpunct{[}{]}{;}{a}{ }{ } 
\let\shortcite\citeyearpar
\let\cite\citep

\usepackage[ruled]{algorithm2e} %

\SetAlFnt{\small}
\SetAlCapFnt{\small}
\SetAlCapNameFnt{\small}
\SetAlCapHSkip{0pt}

\usepackage{graphicx}
\usepackage{subcaption}
\usepackage{multirow}
\usepackage{placeins}
\usepackage{cuted}
\usepackage{caption}

\usepackage{xcolor}
\usepackage[export]{adjustbox}

\title{Joint Orientation and Weight Optimization for Robust Watertight Surface Reconstruction via Dirichlet-Regularized Winding Fields}

\author{
  Jiaze Li \\
  Nanyang Technological University\\
  Singapore\\
   \And
  Daisheng Jin \\
  Nanyang Technological University \\
  Singapore\\
  \And
  Fei Hou\\
  Institute of Software, Chinese Academy of Sciences\\
  University of Chinese Academy of Sciences\\
  China\\
  \And
  Junhui Hou \\
  City University of Hong Kong \\
  Hong Kong\\
  \And
  Zheng Liu \\
  China University of Geosciences (Wuhan) \\
  China\\
    \And
  Shiqing Xin \\
  Shandong University \\
  China\\
  \And
  Wenping Wang \\
  Texas A\&M University \\
  United States of America\\
  \And
  Ying He\thanks{Corresponding author.} \\
  Nanyang Technological University\\
  Singapore
}

\newcommand{\method}{\textsc{DiWR}}

\begin{document}
\maketitle

\begin{abstract}
We propose Dirichlet Winding Reconstruction (\method{}), a robust method for reconstructing watertight surfaces from unoriented point clouds with non-uniform sampling, noise, and outliers. Our method uses the generalized winding number (GWN) field as the target implicit representation and jointly optimizes point orientations, per-point area weights, and confidence coefficients in a single pipeline. The optimization minimizes the Dirichlet energy of the induced winding field together with additional GWN-based constraints, allowing \method{} to compensate for non-uniform sampling, reduce the impact of noise, and downweight outliers during reconstruction, with no reliance on separate preprocessing. We evaluate \method{} on point clouds from 3D Gaussian Splatting, a computer-vision pipeline, and corrupted graphics benchmarks. Experiments show that \method{} produces  plausible watertight surfaces on these challenging inputs and outperforms both traditional multi-stage pipelines and recent joint orientation-reconstruction methods. 
\end{abstract}


\section{Introduction}
\label{sec:intro}

Reconstructing high-quality 3D surfaces from point clouds is an important problem in computer graphics and geometry processing, with applications in digital heritage preservation, robotic perception, and virtual content creation. Modern acquisition devices can capture large point clouds with little effort, but the raw sensor output is typically discrete and disorganized. Real-world scans often exhibit non-uniform sampling, strong noise, outliers, and missing regions, which makes surface reconstruction challenging.

Many classical reconstruction methods assume oriented point clouds. Implicit function methods such as Poisson Surface Reconstruction (PSR)~\cite{kazhdan2006poisson} and Screened PSR (sPSR)~\cite{kazhdan2013screened} are widely used because they produce smooth and watertight surfaces and can tolerate moderate noise when reliable normals are available. However, many acquisition pipelines output \emph{unoriented} point clouds, and enforcing global orientation consistency is difficult, especially for shapes with complex topology, thin structures, or sparse observations.

A standard way to handle unoriented inputs is to decompose the problem into separate stages. A classic example is the work of Huang et al.~\shortcite{huanghui2009}, which introduces a robust point-cloud consolidation pipeline as a preprocessing step for surface reconstruction. Starting from raw scans with noise, outliers, non-uniform sampling, and even closely spaced surface sheets, they first produce a denoised, outlier-free, and more evenly distributed particle set via weighted locally optimal projection, which improves the reliability of local PCA normal estimation. They then propose an iterative normal-estimation framework that combines priority-driven orientation propagation with orientation-aware PCA to obtain globally consistent normals, enabling conventional reconstruction solvers to generate cleaner surfaces. Such multi-stage pipelines are easy to assemble from existing components, but they can be fragile because errors introduced early are difficult to correct later, especially in challenging scenarios. These limitations have motivated unified formulations that couple normal orientation with implicit reconstruction, enabling reconstruction directly from raw, unoriented point clouds.

Despite this progress, robust reconstruction from unoriented point clouds with severe imperfections remains difficult. State-of-the-art unified methods can perform well on reasonably clean inputs, but their performance often degrades when the input exhibits extreme non-uniformity, strong noise, or a large fraction of outliers. A key limitation is that most existing approaches mainly treat \emph{orientation} as the primary unknown and implicitly assume that (i) the discrete sampling adequately represents the underlying surface measure and (ii) all points are equally reliable. Under non-uniform sampling and heavy corruption, these assumptions bias the implicit field and its gradients, leading to unstable orientation updates and degraded reconstructions.

\begin{figure*}[t]
    \centering
    \begin{minipage}[c]{0.48\textwidth}
        \centering
        \includegraphics[width=\linewidth]{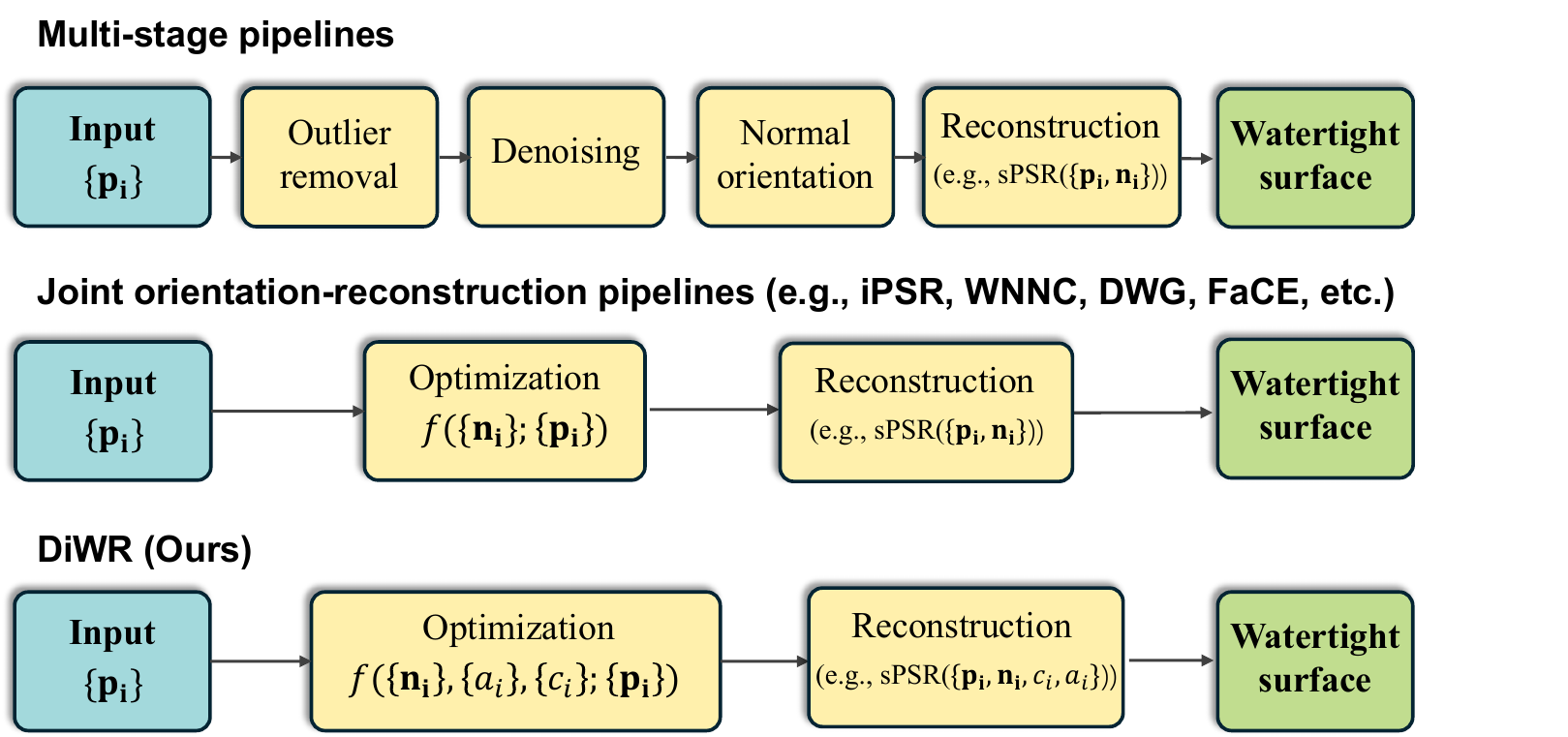}
    \end{minipage}
    \begin{minipage}[c]{0.50\textwidth}
        \centering

        \makebox[0.15in][c]{\centering\rotatebox{90}{\small Input}}
        \includegraphics[width=0.17\linewidth]{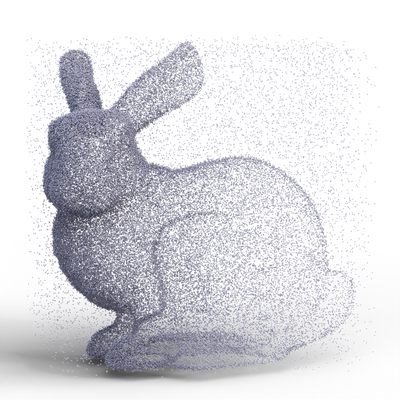}
        \includegraphics[width=0.17\linewidth]{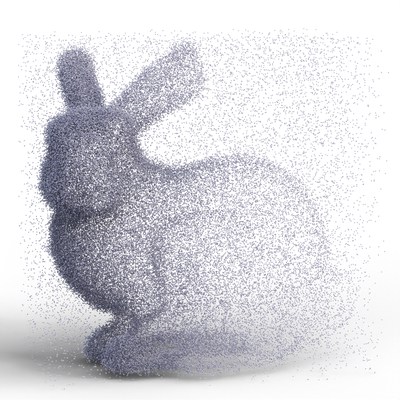}
        \includegraphics[width=0.17\linewidth]{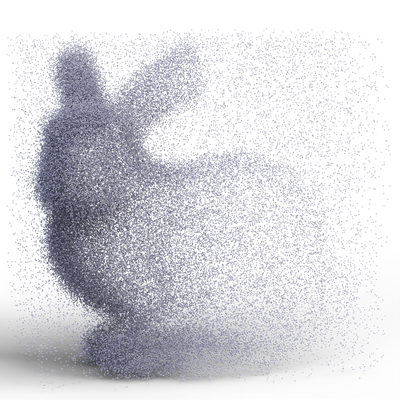}
        \includegraphics[width=0.17\linewidth]{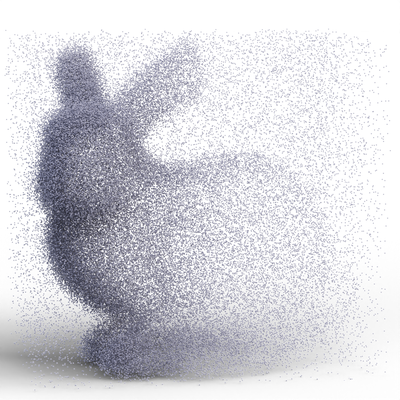}
        \includegraphics[width=0.17\linewidth]{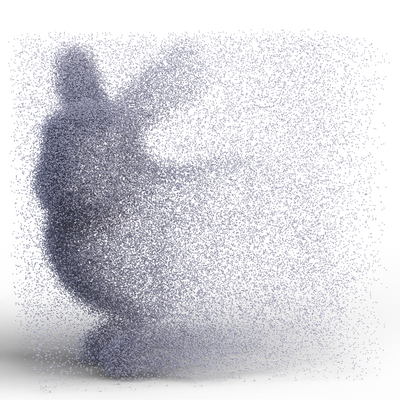}\\

        \makebox[0.15in][c]{\centering\rotatebox{90}{\small Filtered pts}}
        \includegraphics[width=0.17\linewidth]{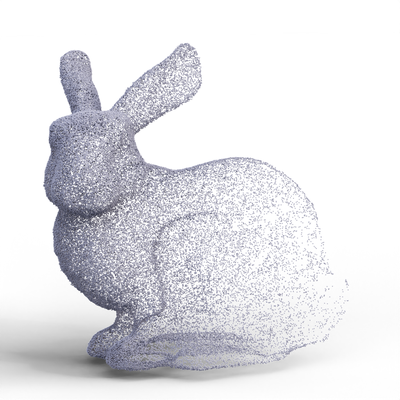}
        \includegraphics[width=0.17\linewidth]{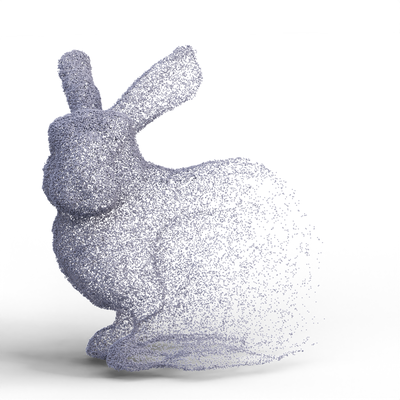}
        \includegraphics[width=0.17\linewidth]{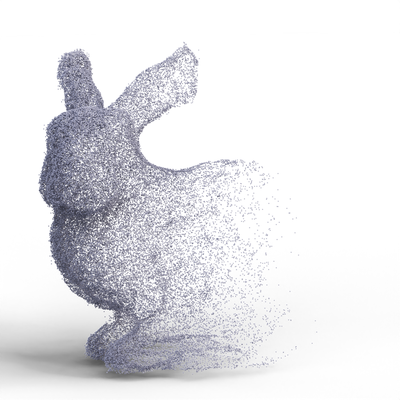}
        \includegraphics[width=0.17\linewidth]{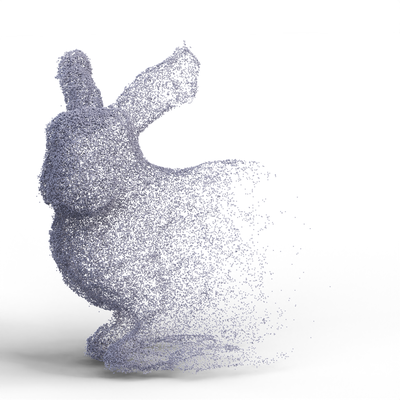}
        \includegraphics[width=0.17\linewidth]{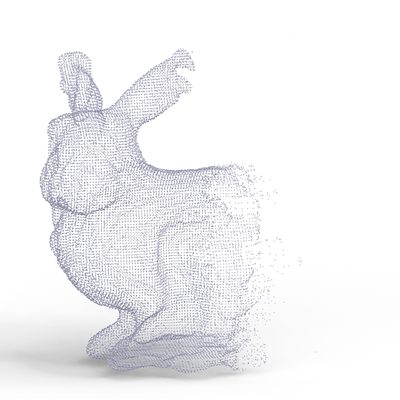}\\

        \makebox[0.15in][c]{\centering\rotatebox{90}{\small WNNC}}
        \includegraphics[width=0.17\linewidth]{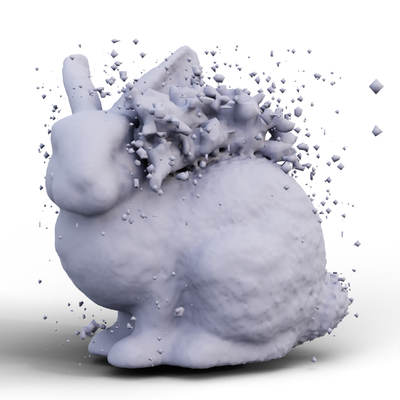}
        \includegraphics[width=0.17\linewidth]{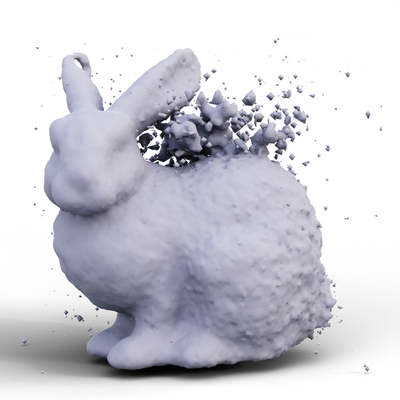}
        \includegraphics[width=0.17\linewidth]{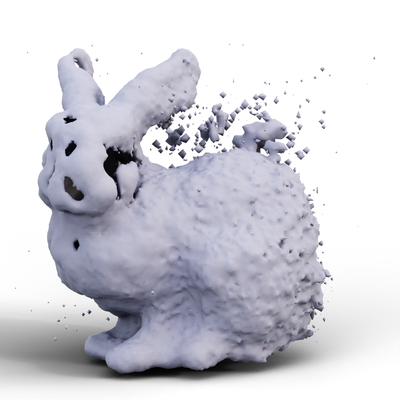}
        \includegraphics[width=0.17\linewidth]{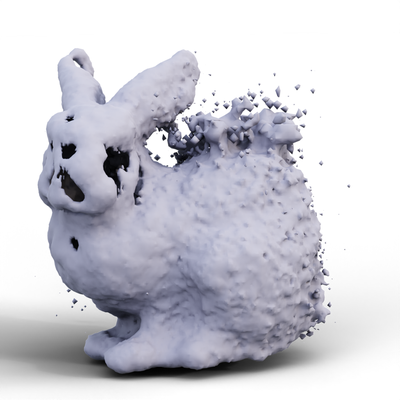}
        \includegraphics[width=0.17\linewidth]{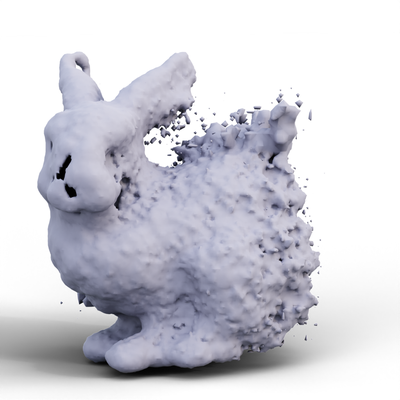}\\

        \makebox[0.15in][c]{\centering\rotatebox{90}{\small WNNC$^\star$}}
        \includegraphics[width=0.17\linewidth]{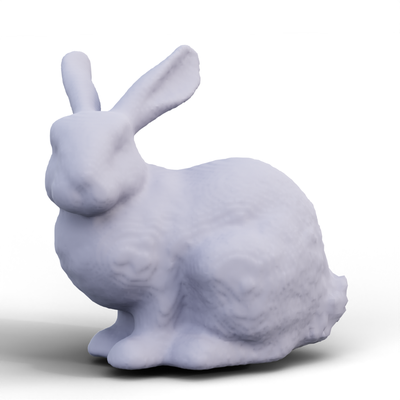}
        \includegraphics[width=0.17\linewidth]{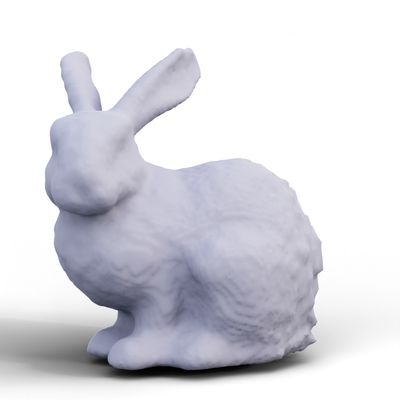}
        \includegraphics[width=0.17\linewidth]{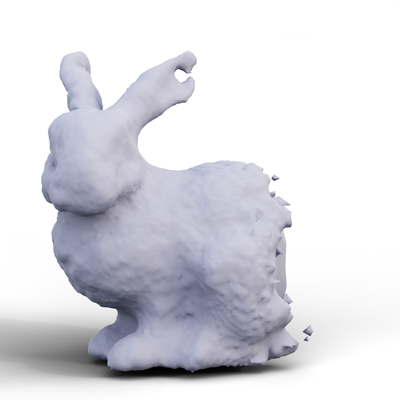}
        \includegraphics[width=0.17\linewidth]{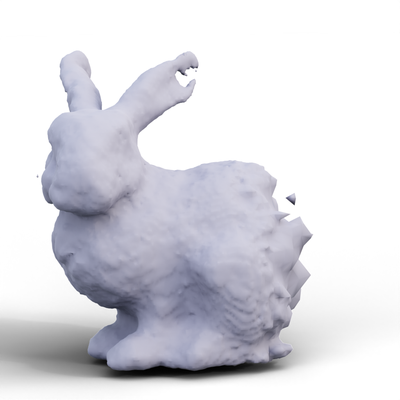}
        \includegraphics[width=0.17\linewidth]{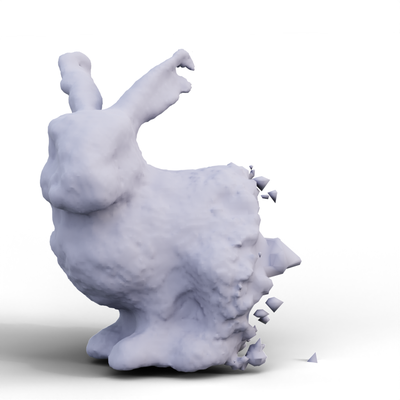}\\

        \makebox[0.15in][c]{\centering\rotatebox{90}{\small Ours}}
        \includegraphics[width=0.17\linewidth]{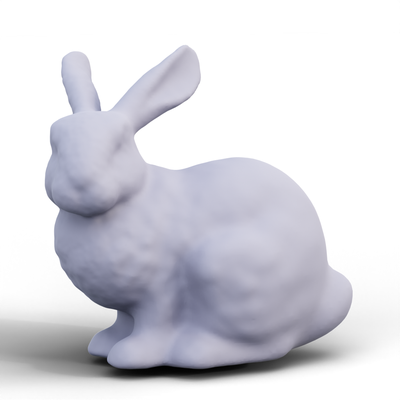}
        \includegraphics[width=0.17\linewidth]{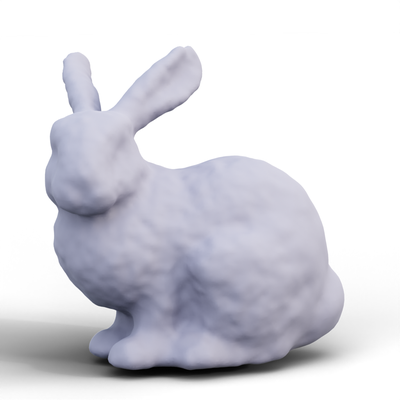}
        \includegraphics[width=0.17\linewidth]{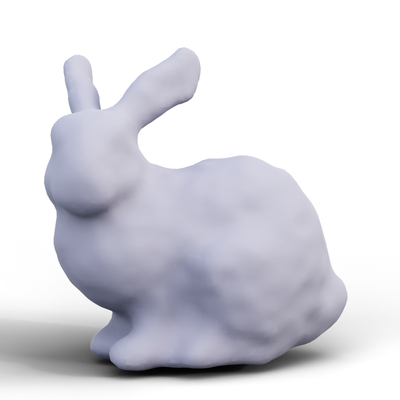}
        \includegraphics[width=0.17\linewidth]{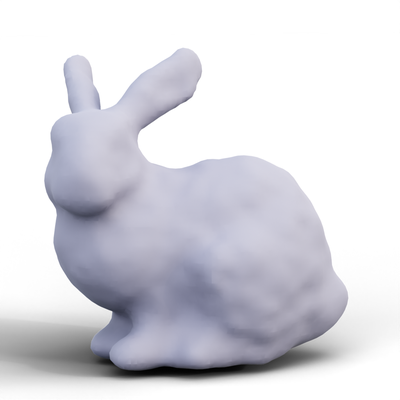}
        \includegraphics[width=0.17\linewidth]{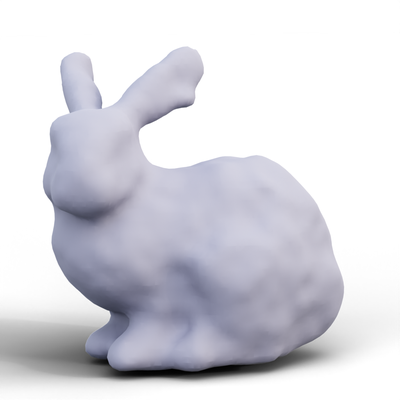}\\

    \end{minipage}
\caption{Conceptual illustration and comparisons. Left: classical multi-stage pipelines typically perform outlier removal and denoising, then estimate and globally orient normals, and finally apply a reconstruction solver (e.g., sPSR~\cite{kazhdan2013screened}) to obtain a watertight surface. Recent unified approaches (e.g., WNNC~\cite{lin2024wnnc}, DWG~\cite{liu2025diffusing}, and FaCE~\cite{faraday}) couple normal orientation with an implicit representation, enabling reconstruction directly from raw, unoriented point clouds. Our method further extends this unified formulation by additionally optimizing per-point area weights and confidence coefficients, improving robustness to non-uniform sampling, noise, and outliers. Right: qualitative results on the Bunny model with increasing corruption (left to right). We show the corrupted inputs, the filtered points after preprocessing, WNNC applied to the raw inputs (WNNC) and to the filtered points (WNNC$^\star$), and our results. As the corruption level increases, WNNC becomes sensitive to outliers and sampling irregularity. In contrast, DiWR operates directly on the raw inputs without preprocessing and maintains coherent watertight reconstructions across a wider range of corruption levels. }
\label{fig:teaser}
\end{figure*}

To address this limitation, we extend the unified formulation by jointly optimizing point orientations together with \emph{per-point area weights} and \emph{confidence coefficients}. Specifically, we adopt the generalized winding number (GWN)~\cite{jacobson2013winding} as the target implicit representation. Its point-based formulation depends explicitly on orientations and per-point weights, and it provides a clear inside-outside indicator that can be evaluated efficiently on point clouds. We regularize the induced GWN field by minimizing its Dirichlet energy, together with additional GWN-based constraints that encourage the desired winding-number values on reliable surface samples and control surface area. Since these variables are coupled nonlinearly, we optimize them in an alternating manner, holding the other variables fixed when updating one group. Our method, Dirichlet Winding Reconstruction (\method{}), iteratively (i) updates orientations to improve global consistency, (ii) refines per-point area weights to compensate for non-uniform sampling, and (iii) estimates confidence coefficients to down-weight outliers and low-quality samples. Unlike existing methods, this joint optimization allows the reconstruction process itself to adapt to uneven sampling and corrupted inputs, reducing reliance on separate preprocessing. As a result, \method{} produces stable, high-quality surfaces on challenging point clouds with significant noise and outliers (see Figure~\ref{fig:teaser}).

\vspace{-0.05in}
\section{Related Work}
\label{sec:related}

Point cloud reconstruction has been studied for more than four decades. Earlier approaches often rely on computational geometry. Representative methods include $\alpha$-Shape~\cite{alphashapes}, Ball Pivoting~\cite{ballpivoting}, Power Crust~\cite{amenta2001power}, and Tight Cocone~\cite{Dey2003}, among others. These methods are computationally efficient and can often provide theoretical guarantees under sampling conditions. However, they lack robustness when handling real-world inputs that often exhibit various types of defects. 

Most modern surface reconstruction pipelines rely on implicit representations, following the seminal work of Hoppe et al.~\shortcite{hoppe1992surface}, because of their robustness and practical performance. Many subsequent methods assume that reliable, consistently oriented normals are given or can be estimated, including moving least squares and Poisson-based approaches~\cite{ohtake2003multi,kazhdan2006poisson,kazhdan2013screened}. In practice, however, obtaining accurate and globally consistent normals from raw scans remains difficult, which limits the applicability of these methods to real-world unoriented point sets. 

To remove the dependency on pre-oriented normals, several methods solve for normal orientation and an implicit surface together. Iterative PSR (iPSR)~\cite{hou2022iterative} alternates between reconstructing an implicit function and updating normals using its gradients. \citet{MA2024102315} incorporate isovalue constraints into the Poisson formulation to solve for globally consistent normal orientations and the implicit function simultaneously via a single sparse linear least-squares system. \citet{faraday} estimate normals by modeling a Faraday-cage electrostatic effect via a Poisson system and using gradients of the resulting field to orient normals, improving robustness in the presence of interior artifacts. 

Parametric Gauss Reconstruction (PGR)~\cite{lin2022pgr} combines normal consistency with gradient-field constraints through parametric Gauss mapping. AGR~\cite{AGR} extends PGR by adding a convection term to the Laplace operator to utilize anisotropic directional information in the point cloud. This approach leads to more effective linear equations and enhances orientation and reconstruction quality, especially for thin structures and small holes.

Recently, techniques based on GWN~\cite{jacobson2013winding} have gained popularity, as GWN enables global inside-outside reasoning and is efficient to evaluate on point clouds~\cite{barill2018fast}. GCNO~\cite{xu2023gcno} and BIM~\cite{liu2024bim} use winding-number-based constraints or energies to recover globally consistent orientation from random initialization. WNNC~\cite{lin2024wnnc} further enforces agreement between normals and the gradients of the induced winding field. DWG~\cite{liu2025diffusing} introduces a diffusion-based framework that supports highly parallel computation and scales well on GPUs. 

In addition to GWN-based formulations, several optimization-driven techniques have been proposed for unoriented point clouds. \citet{Kai_linear} formulate normal orientation via $O(n)$ sparse linear systems derived from Stokes' theorem. \citet{Huang2024Stochastic} propose stochastic normal orientation by optimizing a probabilistic objective that combines global inside-outside cues with local consistency. Beyond orientation, variational methods also provide implicit reconstruction formulations. \citet{huang2019vipss} define the implicit function as the solution to a constrained quadratic optimization problem, enabling exact interpolation of the input points. \citet{fastvipss} accelerate this variational framework by exploiting the locality of natural neighborhoods, substantially improving runtime and scalability.

While the above methods significantly improve reconstruction from unoriented inputs, most of them mainly optimize orientations (and the implicit field) while assuming point positions are reliable. In contrast, our method also optimizes per-point area weights and confidence coefficients, which enhances robustness to uneven sampling, severe noise, and a high outlier ratio.

There is also a large body of work on deep learning methods for 3D reconstruction. These approaches typically learn signed or unsigned distance functions directly from unoriented points~\cite{DBLP:conf/icml/MaHLZ21,10377458,nsh,10.5555/3666122.3668895,Neural-IMLS,ren2023geoudf,10.5555/3666122.3668895,ImplicitFilter,losf,deudf}. While this avoids an explicit normal-orientation step, such methods are often limited to small- to middle-scale inputs due to their high memory consumption. We provide additional discussion and quantitative comparisons with learning-based methods in the Appendix.

\vspace{-0.05in}
\section{Preliminaries}
\label{sec:preliminaries}

The generalized winding number ~\cite{jacobson2013winding} extends the classical winding number from closed curves to closed surfaces, distinguishing the inside and outside of a solid. Originating from potential theory, the winding number can be interpreted as a scalar potential field whose level sets represent the geometric enclosure of space. Let $\Omega \subset \mathbb{R}^3$ be a solid with smooth boundary $\partial\Omega$. For a point $\mathbf{x}\in\mathbb{R}^3$, denote by $\mathbf{n}(\mathbf{y})$ the outward unit normal at $\mathbf{y}\in \partial\Omega$. The continuous winding number field is defined as
\begin{equation}
w(\mathbf{x})=\frac{1}{4\pi}\int_{\partial\Omega}
\frac{\langle \mathbf{n}(\mathbf{y}),\, \mathbf{y}-\mathbf{x}\rangle}{\|\mathbf{y}-\mathbf{x}\|^{3}}\, \mathrm{d}S(\mathbf{y}),
\label{eq:gwn_continuous}
\end{equation}
where $\mathrm{d}S$ is the area element. The generalized winding number measures how many times the surface wraps around the query point $\mathbf{x}$. For a closed orientable surface, $w(\mathbf{x})$ takes values close to $1$ in the interior of $\Omega$, close to $0$ in the exterior, and approximately $0.5$ on the boundary $\partial\Omega$.

Given an oriented point set $\{(\mathbf{p}_i,\mathbf{n}_i)\}_{i=1}^n$ sampled from $\partial\Omega$, Equation~\eqref{eq:gwn_continuous} can be discretized as~\cite{barill2018fast}:
\begin{equation}
w(\mathbf{q}) \approx \sum_{i=1}^{n} a_i
\frac{\langle \mathbf{n}_i, \mathbf{p}_i-\mathbf{q}\rangle}{4\pi\|\mathbf{p}_i-\mathbf{q}\|^{3}},
\label{eq:gwn_discrete}
\end{equation}
where $a_i$ is the local surface element area associated with $\mathbf{p}_i$. 

Ideally, the area weights $\{a_i\}$ should be computed from a geodesic Voronoi diagram on the underlying surface~\cite{intrinsiccvt}, since such cells provide a principled discretization of the surface integral. However, without point connectivity and orientation, constructing accurate geodesic Voronoi cells from a raw point cloud is technically challenging. In practice, one often uses a local planar approximation based on 2D Voronoi diagrams~\cite{barill2018fast,xu2023gcno,liu2024bim}: for each point $\mathbf{p}_i$, estimate a local tangent plane via PCA, project a neighborhood of $\mathbf{p}_i$ onto this plane, and compute the 2D Voronoi cell on the plane. The area of this 2D cell is then used as an approximation of the corresponding geodesic Voronoi area weight $a_i$. This discrete form enables efficient evaluation of GWN directly on point clouds, and has been used in geometry processing tasks such as inside--outside queries, Boolean operations, shape analysis, and global normal orientation.

The GWN field $w$ is harmonic in $\mathbb{R}^3\setminus \partial\Omega$ and satisfies the usual jump conditions across $\partial\Omega$. In particular, $\nabla w$ aligns with the globally consistent outward normals on the surface, indicating that GWN encodes both inside--outside information and orientation continuity. This makes GWN a reliable implicit representation and a useful foundation for formulating globally consistent normal orientation. \citet{TakayamaJKS14} explore orienting polygon soups by minimizing the Dirichlet energy of the induced winding field using a 0-1 integer programming formulation. In practice, this approach is computationally prohibitive due to the large number of binary variables. In contrast, \citet{liu2024bim} relax the problem by treating orientations as continuous variables and solve it via nonlinear optimization on unoriented point clouds.

GWNs have found extensive applications in digital geometry processing, including robust inside-outside segmentation~\cite{jacobson2013winding},  Boolean operations~\cite{10.1145/2897824.2925901}, containment queries for parametric geometry~\cite{10.1145/3730886}, and point cloud orientation~\cite{xu2023gcno,lin2024wnnc},
among others.

\section{Method}

\begin{figure*}[t]
\centering
\includegraphics[height=1.20in]{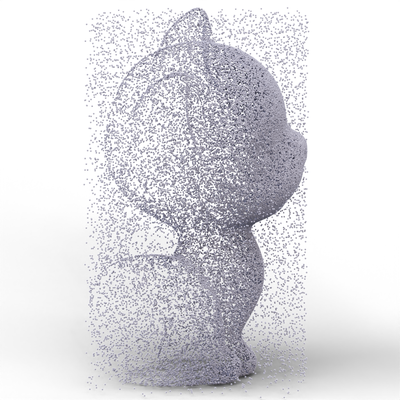}
\includegraphics[height=1.20in]{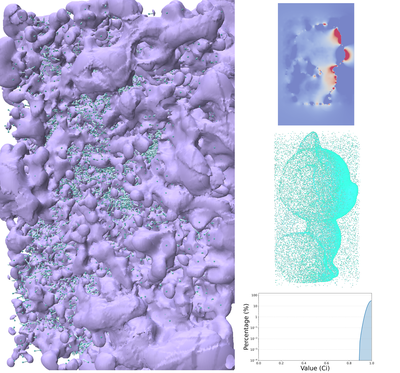}
\includegraphics[height=1.20in]{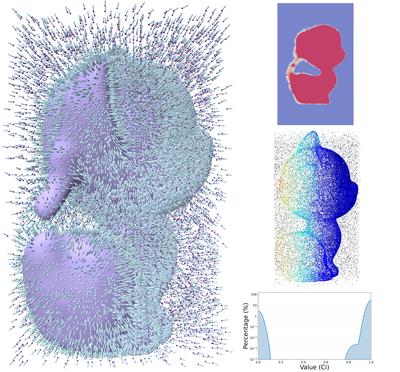}
\includegraphics[height=1.20in]{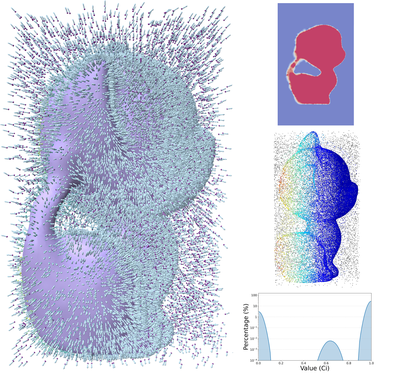}\\
\makebox[1.2in]{Input}
\makebox[1.4in]{$t=0$}
\makebox[1.2in]{$t=1$}
\makebox[1.2in]{$t=2$}\\
\includegraphics[height=1.20in]{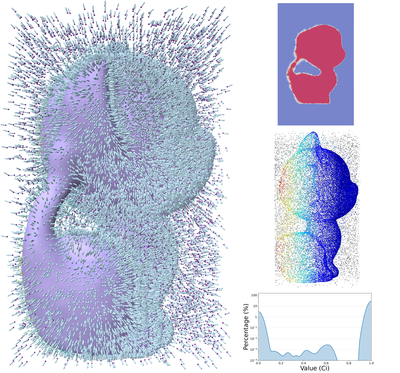}
\includegraphics[height=1.20in]{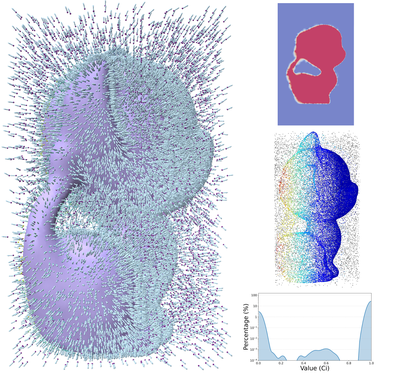}
\includegraphics[height=1.20in]{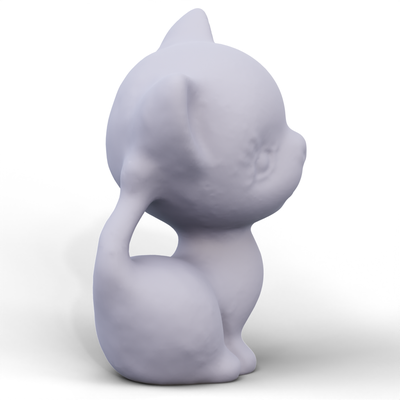}
\includegraphics[height=1.20in]{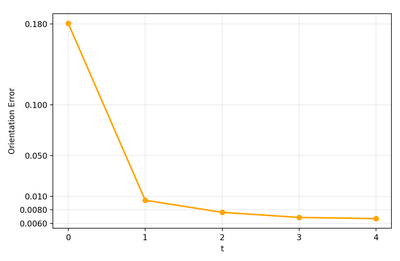}\\
\makebox[1.4in]{$t=3$}
\makebox[1.1in]{$t=4$}
\makebox[1.4in]{$\mathcal{S}$}
\makebox[1.7in]{Error vs.\ iter.}\\
\caption{Overview of \method{} on Kitten ($n=60000,\widehat{u}=0.54,\widehat{o}=0.16,\widehat{\sigma}=6.79\times10^{-5}$).Starting from random orientations, the algorithm alternates updates of point orientations together with per-point area weights $a_i$ and confidence coefficients $c_i$. Each intermediate panel (from $t=1$ to $t=4$) visualizes the reconstructed surface, point orientations, cut views of the induced GWN field, the effective weights $a_ic_i$, and the distribution of confidence coefficients $c_i$ (shown on a log scale to accommodate the large dynamic range in counts). After convergence, we retain high-confidence points and pass their orientations and effective weights to sPSR to obtain a watertight surface $\mathcal{S}$. The rightmost plot reports the mean orientation error versus iteration.}
    \label{fig:pipeline}
\end{figure*}

\label{sec:method}
Let $\mathbf{P}=\{\mathbf{p}_i\}_{i=1}^n$ be an unoriented point set sampled from a watertight manifold surface, possibly corrupted by noise and outliers. We treat the point orientations $\mathbf{N}=\{\mathbf{n}_i\}_{i=1}^n$ and area weights $\mathbf{a}=\{a_i\}_{i=1}^n$ as unknown variables. To reduce the negative effects of noise and outliers, we assign each point $\mathbf{p}_i$ a confidence coefficient $c_i\in[0,1]$, where $c_i=1$ indicates an inlier and $c_i=0$ an outlier. Denote by $\mathbf{c}=\{c_i\}_{i=1}^n$. We collect these unknowns into $\theta:=(\mathbf{N}, \mathbf{a}, \mathbf{c})$, and express the generalized winding number field at a query point $\mathbf{q}$ as
\begin{equation}
w_\theta(\mathbf{q})=\sum_{i=1}^n a_ic_i
\frac{(\mathbf{p}_i-\mathbf{q})\cdot \mathbf{n}_i}{4\pi\|\mathbf{p}_i-\mathbf{q}\|^3}.
\end{equation}

\vspace{-0.05in}
\subsection{Objective Functions}
\label{sec:functions}

\paragraph{Dirichlet Energy}
In the continuous setting, the winding field induced by a globally consistent orientation is harmonic in $\mathbb{R}^3\setminus \partial\Omega$ and therefore minimizes the Dirichlet energy among fields with the same boundary conditions. In our discrete setting, however, the target surface $\partial\Omega$ is unknown. We thus exclude a thin band around the surface by removing a neighborhood of the input points that are deemed reliable. Concretely, let $\mathcal{I}=\{\,i \mid c_i \ge \tau_\text{in}\,\}$ be the set of high-confidence points (with $\tau_\text{in}$ as a threshold), and define the excluded band
$\mathcal{U}_\delta(\theta) \;=\; \textstyle\bigcup_{i\in \mathcal{I}} B(\mathbf{p}_i,\delta)$,
where $\delta$ controls the band thickness. We then define the Dirichlet energy in the remaining region as
\begin{equation}
E_{\text{diri}}(\theta)
=\int_{\mathbb{B}\setminus \mathcal{U}_\delta(\theta)} \left\|\nabla w_\theta(\mathbf{x})\right\|^2\, \mathrm{d} V(\mathbf{x}),
\label{eq:dirichlet_discrete}
\end{equation}
where $\mathrm{d}V$ is the volume element, and $\mathbb{B}$ is a bounding box that surrounds the input points with a sufficient margin. Low-confidence points (small $c_i$) are considered outliers and are not used to define the excluded band.

\paragraph{Discrete Evaluation}
To evaluate $E_{\text{diri}}$ in practice, we approximate the volume integral by uniform sampling in $\mathbb{B}$. Since the integrand is not defined on the (unknown) surface, we ignore samples that fall inside the excluded band and down-weight samples close to it. Specifically, for each high-confidence point $\mathbf{p}_i$ with $c_i>\tau_\text{in}$, we remove energy samples inside the ball $B(\mathbf{p}_i,r_s)$, where $r_s$ controls the band thickness (we set  $\tau_{\text{in}}=0.9$ and  $r_s=0.03$ in all experiments). For samples that lie outside but near this ball, we compute a partial-volume weight based on the intersection volume between the ball and the voxel centered at the sample~\cite{jones2017fast}. The resulting weight is proportional to the remaining (non-intersected) voxel volume and lies in $[0.5,1]$. With a uniform grid of sampling points $\mathbf{q}$ and voxel volume $V_c$, the discrete approximation takes the form
\begin{equation}
\widehat{E}_{\text{diri}}(\theta)
= \sum_{\mathbf{q}\in\mathcal{Q}} \delta_{\mathbf{q}} V_c\left\|\nabla w_\theta(\mathbf{q})\right\|^2,
\label{eq:dirichlet_discrete_sum}
\end{equation}
where $\mathcal{Q}$ is the set of sampled grid points and $\delta_{\mathbf{q}}$ is the partial-volume weight described above.

\paragraph{Surface Points}
For a closed orientable surface, the winding number field takes values close to $1$ inside and close to $0$ outside; ideally, $w=\tfrac{1}{2}$ on the boundary $\partial\Omega$. We therefore introduce a surface term that encourages high-confidence points to lie near the $\tfrac{1}{2}$-level set of the induced field:
\begin{equation}
E_{\text{surf}}(\theta)
=\frac{1}{|\mathcal{I}|}\sum_{i\in\mathcal{I}}
\left(w_\theta(\mathbf{p}_i)-\frac{1}{2}\right)^2.
\end{equation}
When sampling is sparse or highly non-uniform, this term provides extra anchoring constraints at reliable samples, helping prevent drift of the $\tfrac{1}{2}$-level set.

\paragraph{Stable Surface Area}
To avoid degenerate solutions (e.g., driving all weights to zero) and to stabilize the optimization, we constrain the total effective surface area to remain roughly constant within each optimization stage. Specifically, we penalize deviations of the weighted area sum from its value at the start of the current stage:
\begin{equation}
E_{\text{area}}(\mathbf{a},\mathbf{c}) =
\left|\sum_{i=1}^n a_i c_i -\sum_{i=1}^n a_i^{b} c_i^{b}\right|,
\end{equation}
where $a_i^{b}$ and $c_i^{b}$ represent the values of $a_i$ and $c_i$ at the beginning of the current optimization stage. This term encourages the total effective surface area $\sum_i a_i c_i$ to remain stable during the current stage.

\paragraph{Polarized Confidence Coefficients}
To encourage the confidence weights $c_i$ of outliers to decrease toward $0$ and thus reduce their influence on optimization and reconstruction, we introduce a binary term $E_{\text{conf}}$ that encourages $c_i$ to concentrate near $0$ and $1$:
\begin{equation}
E_{\text{conf}}(\mathbf{c}) = \sum_{i=1}^n |c_i(1-c_i)|.
\label{eqn:polarized}
\end{equation}
Without this term, many points may have intermediate confidence values, allowing outliers to continue affecting the winding field and potentially biasing $\nabla w_\theta$ and the related orientation updates. Encouraging near-binary confidences helps create a clearer separation between inliers and outliers and improves the stability of the overall optimization.

\begin{figure}[t]
\centering
\includegraphics[width=0.25\linewidth]{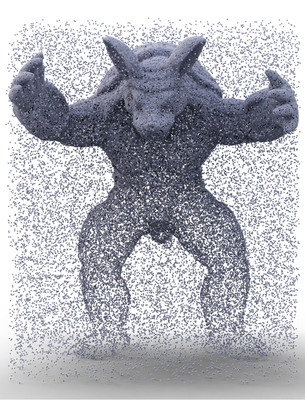}
    \includegraphics[width=0.25\linewidth]{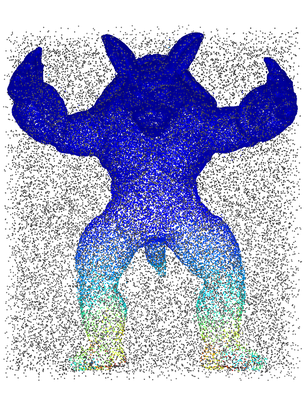}
    \includegraphics[width=0.25\linewidth]{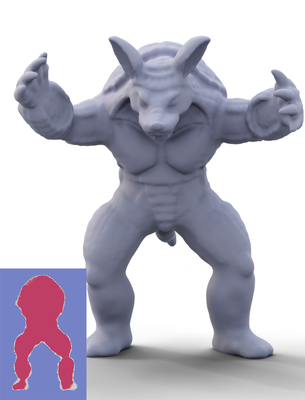}\\

    \makebox[0.25\linewidth]{(a)}
    \makebox[0.25\linewidth]{(b)}
    \makebox[0.25\linewidth]{(c)}\\
\caption{Area-weight optimization compensates for non-uniform sampling on the corrupted Armadillo model ($n=172,500, \widehat{\sigma}=1.6\times10^{-4},\widehat{o}=0.13, \widehat{u}=0.30$). 
(a) Input point cloud exhibiting a dense-to-sparse sampling pattern and outliers; we initialize all area weights $a_i$ uniformly. (b) Optimized effective weights $a_i c_i$ (visualized with a heat colormap, warm = larger) become spatially adaptive: inlier weights increase in sparsely sampled regions and decrease in densely sampled regions, while outliers receive negligible effective weights through $c_i$. (c) The rebalanced weights yield a more balanced discrete winding-number integration, stabilizing the induced field (shown in the inset) and improving downstream watertight reconstruction.}
    \label{fig:area-optimization}
\end{figure}
\subsection{Optimization}
\label{subsec:opt}

\begin{figure}[t]
  \centering
\includegraphics[width=0.15\linewidth]{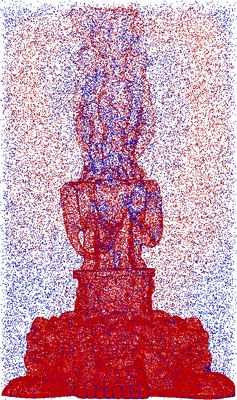}
\includegraphics[width=0.15\linewidth]{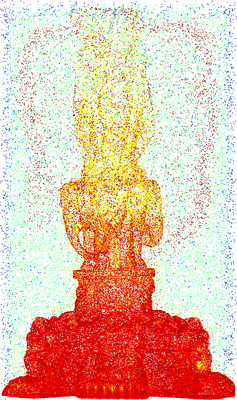}
\includegraphics[width=0.15\linewidth]{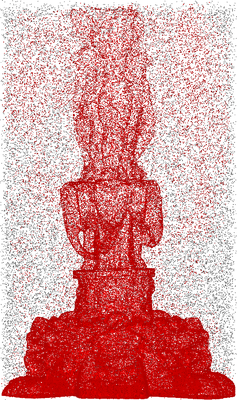}
\includegraphics[width=0.15\linewidth]{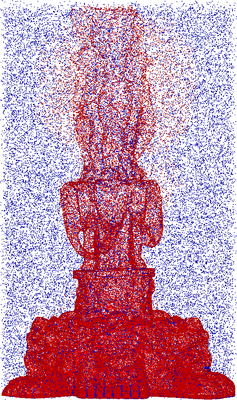}
\includegraphics[width=0.15\linewidth]{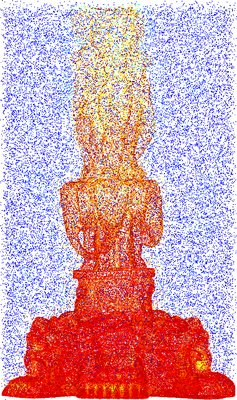}
\includegraphics[width=0.15\linewidth]{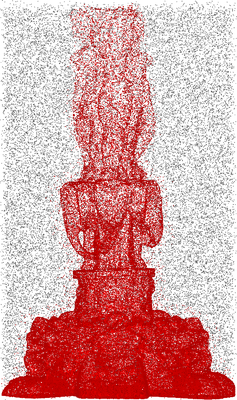}\\
  \makebox[0.15\textwidth]{(a)}
  \makebox[0.15\textwidth]{(b)}
  \makebox[0.15\textwidth]{(c)}
  \makebox[0.15\textwidth]{(d)}
  \makebox[0.15\textwidth]{(e)}
  \makebox[0.15\textwidth]{(f)}\\
\caption{Resetting and optimizing the confidence coefficients in one $c$-optimization stage on Thai Statue ($n=240,000
, \widehat{\sigma}=0.00026
,\widehat{o}=0.16,\widehat{u}=0.45$). (a)-(c): an early iteration where orientations are still inaccurate; (d)-(f): a later iteration close to convergence. (a, d) Coarse initialization via bi-means clustering of the current GWN values: points in the outlier cluster are shown in purple, while the remaining points are treated as inliers in yellow  (b,e) Density-based stratification softens this binary initialization and assigns multi-level confidences in $[0,1]$.  (c,f) After optimization, the confidence distribution becomes strongly bimodal, concentrating near $0$ and $1$. For visualization, we show low-confidence points in gray and color the remaining points by their effective weights $a_ic_i$ using a heat colormap.}
\label{fig:ci_optimization}
\end{figure}

\paragraph{Initialization}
For each input point $\mathbf{p}_i$, we initialize the confidence at $c_i=1$ and set the orientation $\mathbf{n}_i$ to a random unit vector. We compute the initial area weights $\{a_i\}$ using the 2D Voronoi approximation from prior work~\cite{barill2018fast}. For heavily corrupted inputs (e.g., with many outliers), this geometric estimation can become unreliable; in such cases, we optionally initialize all area weights uniformly (e.g., $a_i \leftarrow 1$) and let the subsequent optimization refine them. Finally, we precompute a local density estimate $\rho_i$ by counting the number of neighbors within a ball $B(\mathbf{p}_i, r_\rho)$, where we set the default radius $r_\rho=0.06$ in our implementation. This density is later used in the density-stratified reset of confidence coefficients.

\paragraph{Strategies}
The per-point area weights $a_i$, confidence coefficients $c_i$, and orientations $\mathbf{n}_i$ all influence the induced GWN field. Because these variables are coupled nonlinearly, optimizing them together can create a complex and potentially unstable problem. We therefore use a staged alternating strategy: in each stage, we update one group of variables while keeping the others fixed, which enhances numerical stability and makes the optimization easier to control.

\begin{figure}[t] 
\centering 
\includegraphics[width=0.47\linewidth]{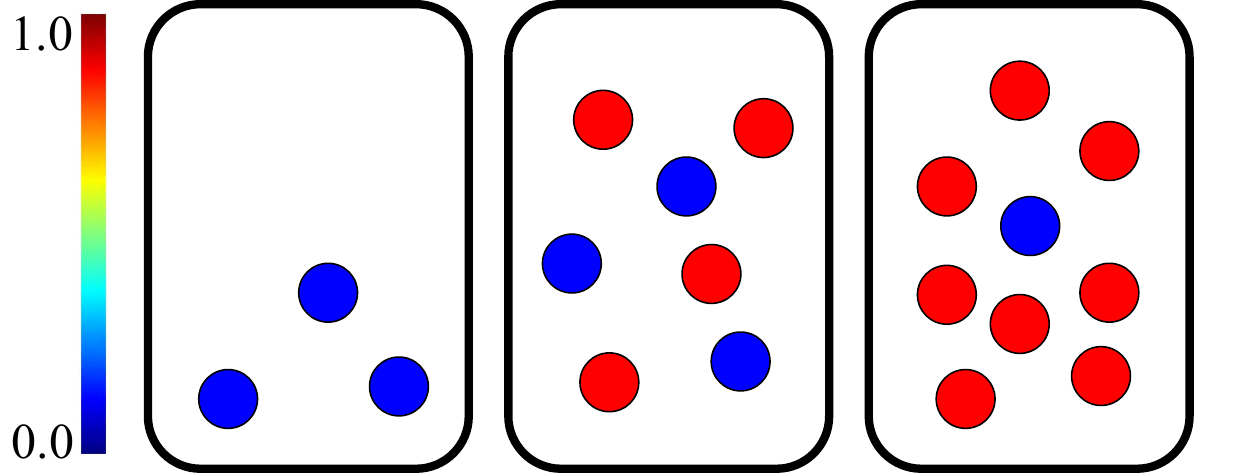} 
\includegraphics[width=0.395\linewidth]{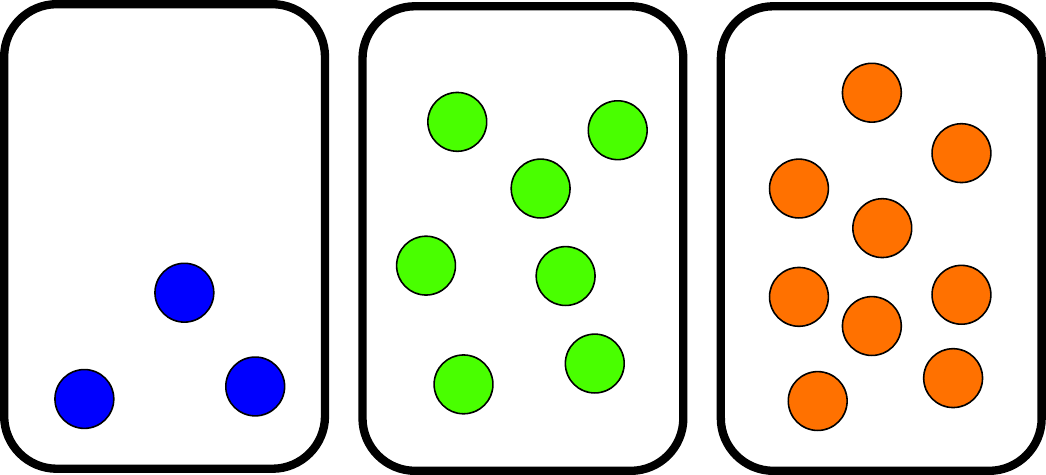}\\ 
\makebox[0.47\textwidth]{(a)} 
\makebox[0.395\textwidth]{(b)} 
\caption{
Density stratification for resetting confidence coefficients. We illustrate the refinement step using a simplified example with three density levels. Each dot is an input point; color encodes confidence (blue: $c_i=0$, red: $c_i=1$, with warmer colors indicating higher confidence). (a) After the coarse bi-means split on winding values, each point receives a binary confidence ($0$ for the outlier cluster and $1$ for the inlier cluster). We then group points into density levels according to their local density $\rho_i$ (each box denotes one level). (b) Within each density level, we replace the binary assignments by the level-wise mean confidence, assigning all points in the level the same averaged value. This produces a smoother initialization with confidence values in $[0,1]$, which is used as the starting point for the subsequent $c$-optimization.
} 
\label{fig:stratification} 
\end{figure}

Since the three variable groups influence each other, the overall procedure is iterative. Each outer iteration consists of three stages: (i) DWG-based orientation update for $\{\mathbf{n}_i\}$, (ii) optimization of area weights $\{a_i\}$, and (iii) optimization of confidence coefficients $\{c_i\}$. Within an outer iteration, we first alternate between stages (i) and (ii) until the area weights stabilize. Specifically, letting $a_i^{b}$ denote the area weights at the beginning of the current $a$-optimization stage, we monitor the average relative change 
$\delta_a=\tfrac{1}{n}\textstyle\sum_{i=1}^{n}\left|(a_i-a_i^{b})/a_i^{b}\right|$. We stop alternating between the orientation update and the area-weight optimization when $\delta_a\leq\epsilon_a$. We then proceed to stage (iii) to update $\{c_i\}$. After finishing the confidence optimization, we advance to the next outer iteration. The details of each stage are described next. See Algorithm~\ref{alg:diwr} for high-level pseudo-code.

\paragraph{Optimizing Area Weights $\{a_i\}$}
With orientations $\{\mathbf{n}_i\}$ and confidence coefficients $\mathbf{c}$ fixed, we optimize the area weights $\mathbf{a}$ by minimizing a weighted sum of the Dirichlet energy, the surface-point term, and the stable-area term:
\begin{equation}
\min_{\mathbf{a}}
\widehat{E}_{\text{diri}}(\mathbf{N},\mathbf{a},\mathbf{c})
+\lambda_1\,E_{\text{surf}}(\mathbf{N},\mathbf{a},\mathbf{c})
+\lambda_2\,E_{\text{area}}(\mathbf{a},\mathbf{c}),
\label{eqn:area_optim}
\end{equation}
where $\lambda_1,\lambda_2>0$ are balancing weights that control the trade-off among the terms.

\paragraph{Optimizing Confidence Coefficients $\{c_i\}$} Since our optimization is staged, the current winding field provides a useful signal for separating inliers from outliers. We therefore reset the confidence coefficients at the beginning of each $c$-optimization stage using the current field. The reset consists of a coarse split followed by a refinement step (see Figure~\ref{fig:ci_optimization}). \textbf{(i) Coarse inlier/outlier split from winding values.} We apply bi-means clustering~\cite{SHEN2009256,WANG201531} to the current winding values $\{w_\theta(\mathbf{p}_i)\}$, obtaining two clusters. Let $\overline{w}:=\frac{1}{n}\sum_{i=1}^n w_\theta(\mathbf{p}_i)$ be the global mean winding value. We treat as outliers the cluster whose mean winding value has the larger absolute deviation from $\overline{w}$. Points in this cluster are assigned $c_i \leftarrow 0$, while points in the other cluster are assigned $c_i \leftarrow 1$. \textbf{(ii) Refinement via density stratification.} The binary assignment above can be overly coarse, especially near regions where the winding values are ambiguous. To obtain a smoother initialization, we stratify the points into $128$ density levels according to their precomputed densities $\rho_i$. Within each level, we replace the bi-means clustering induced binary confidence by average confidence of the points in the same level, yielding multi-level values in $[0,1]$ (see Figure~\ref{fig:stratification}). To avoid disrupting points that are already consistent with current surfaces, we keep $c_i$ unchanged for points whose winding values lie close to the global mean (i.e., $\|w_\theta(\mathbf{p}_i)-\overline{w}| \leq 0.1$ in our implementation). This refinement produces an initialization that reflects both global winding-field consistency and local sampling density.

Starting from the above initialization, we optimize $\mathbf{c}$ while keeping orientations $\mathbf{N}$ and area weights $\mathbf{a}$ fixed. The objective is
\begin{equation}
\min_{\mathbf{c}\in[0,1]^n}
\widehat{E}_{\text{diri}}(\mathbf{N},\mathbf{a},\mathbf{c})
+\lambda_3 E_{\text{surf}}(\mathbf{N},\mathbf{a},\mathbf{c})
+\lambda_4 E_{\text{area}}(\mathbf{a},\mathbf{c})
+\lambda_5 E_{\text{conf}}(\mathbf{c}),
\label{eqn:confidence_optimization}
\end{equation}
where $E_{\text{conf}}$ encourages near-binary confidences and the weights $\lambda_3,\lambda_4,\lambda_5>0$ balance the terms.

\paragraph{Updating Point Orientation $\{\mathbf{n}_i\}$}
To update the orientations $\{\mathbf{n}_i\}$, we leverage DWG~\cite{liu2025diffusing}, a parallel and GPU-friendly algorithm that iteratively constructs a winding field from randomly initialized normals and drives them toward global consistency. In our implementation, we use DWG as a black-box normal update operator: given the current orientations $\{\mathbf{n}_i\}$ and the effective per-point weights $\{a_i c_i\}$, DWG returns an updated set of orientations. %

\begin{algorithm}[t]
\small
\caption{Dirichlet Winding Reconstruction (DiWR)}
\label{alg:diwr}

\KwIn{Unoriented points $\{\mathbf{p}_i\}_{i=1}^n$, area-weight threshold $\epsilon_a$, normal-update threshold $\epsilon_n$, maximum outer iterations $t_{\max}$, and balancing weights $\lambda_i$ ($i=1,\ldots,5$)}
\KwOut{Reconstructed watertight surface $\mathcal{S}$}

\For{each input point $\mathbf{p}_i$}{
$c_i \leftarrow 1$ \\
Initialize $a_i$ via local 2D Voronoi approximation\\
Randomly initialize $\mathbf{n}_i$ as a unit vector\\
}

$t \leftarrow 1$\\
\Repeat{$\Delta_{\mathbf{n}} \leq \epsilon_n$ \textbf{or} $t>t_{\max}$}{
\Repeat(\tcp*[f]{inner loop until area weights stabilize}){$\delta_a \leq \epsilon_a$}{
Update $\{\mathbf{n}_i\}$ using DWG with effective weights $\{a_ic_i\}$\\
Optimize $\{a_i\}$ by minimizing Eq.~\eqref{eqn:area_optim}\\
}
Optimize $\{c_i\}$ by minimizing Eq.~\eqref{eqn:confidence_optimization}\\
Compute normal change $\Delta_{\mathbf{n}}$ on high-confidence points\\
$t \leftarrow t+1$
}

Retain high-confidence points $\mathcal{I}=\{i\mid c_i\geq \tau_{\text{in}}\}$ \\
Run sPSR on $\{(\mathbf{p}_i,\mathbf{n}_i)\}_{i\in\mathcal{I}}$ (optionally with screening weights proportional to $\{a_ic_i\}_{i\in\mathcal{I}}$) to obtain $\mathcal{S}$\\
\Return{$\mathcal{S}$}
\end{algorithm}

\vspace{-0.05in}
\subsection{Reconstruction}
\label{subsec:reconstruction}

After optimization, we obtain per-point orientations $\{\mathbf{n}_i\}$, area weights $\{a_i\}$, and confidence coefficients $\{c_i\}$. We then reconstruct a watertight surface using Screened Poisson Surface Reconstruction (sPSR)~\cite{kazhdan2013screened}. Due to the polarization term (Eq.~\eqref{eqn:polarized}), the optimized confidence coefficients become approximately binary. We therefore treat low-confidence points as outliers and retain only points with $c_i\geq \tau_{\text{in}}$. The retained points, along with their orientations $\mathbf{n}_i$, define the input normal field for sPSR.

A key difference from the standard sPSR pipeline is that we optionally provide sPSR with per-point weights derived from our optimization. In sPSR, the screened term controls how strongly the reconstructed implicit function is encouraged to match prescribed values at the sample locations, and the default setting treats samples uniformly~\cite{kazhdan2013screened}. In our pipeline, when the input exhibits strong non-uniform sampling and/or residual outliers, we weight the screened term using the effective area weights ${c_ia_i}$. This rebalances the contribution of samples across regions with different densities and further reduces the influence of low-confidence points, preventing densely sampled areas from dominating the reconstruction. %

On easier inputs with relatively low sampling non-uniformity $\widehat{u}$ and/or low noise $\widehat{\sigma}$ and outlier rate $\widehat{o}$, we find that directly feeding the oriented points into sPSR without additional weighting already produces high-quality watertight reconstructions. This is because sPSR is robust to mild sampling variation and moderate defections through its global formulation and adaptive discretization. We therefore mainly enable the weighted screened term for challenging cases where non-uniformity and outliers are more severe.

\vspace{-0.05in}
\section{Experimental Results}
\label{sec:results}

\method{} repeatedly invokes DWG~\cite{liu2025diffusing}, a GPU-based algorithm, for point orientation. To reduce the overhead caused by frequent data transfers between the CPU and GPU, we implement the core components of \method{} in CUDA. Optimization is performed using RMSProp, which is chosen for its simplicity and ease of use in parallel processing. All experiments are conducted on a single NVIDIA GeForce RTX 4090 GPU. For an input point cloud with $100$K points, one run of area-weight optimization, confidence optimization, and DWG-based orientation update typically takes 30, 20, and 2 seconds, respectively. CPU-based baselines are evaluated on a high-end
workstation equipped with an Intel Xeon(R) Gold 6430 CPU and 128 GB of RAM.

\subsection{Test Models} 
\label{subsec:testmodels}
We evaluate on 25 models from three categories: (i) point clouds produced by a computer-vision pipeline, specifically the recent vision foundation model VGGT~\cite{Wang_2025_CVPR};
(ii) point clouds extracted from 3D Gaussian Splatting (3DGS) reconstructions of multi-view images~\cite{3DGS}; and
(iii) commonly used graphics benchmarks. 

For the 3DGS category, we use multi-view images from the OmniObject3D dataset~\cite{wu2023omniobject3d}, which provides ground-truth meshes for evaluation. These images are background-free, allowing 3DGS~\cite{3DGS} to produce comparatively clean Gaussian primitives. In this setting, most spurious primitives appear as \emph{interior outliers}, i.e., Gaussians located inside the true surface boundary. Many of these primitives have non-zero opacity and contribute to rendering, but they do not correspond to valid surface geometry. The reconstructed Gaussian centers also exhibit non-uniform sampling, especially in areas with little texture and on planar or low-curvature surfaces. For each Gaussian primitive, we use its center as an input point; no other Gaussian attributes (e.g., opacity, anisotropy, or color) are used.

The computer vision category is built from more challenging multi-view images~\cite{yao2020blendedmvs} with cluttered backgrounds and less controlled capture conditions. Under a sparse-view setting (five to ten views), the point clouds generated by VGGT~\cite{Wang_2025_CVPR} are typically dense but often contain substantially more noise and outliers. Moreover, even when the underlying geometry is a single-layer surface, the reconstructed points frequently form shells around it, that is, several closely spaced layers enclosing the true surface. When background structures lie close to the object of interest, additional spurious surface sheets may also appear around the object. These imperfections make accurate surface reconstruction significantly more difficult than the 3DGS category. 

The graphics-benchmark category includes three widely used models: Armadillo, Dragon and Kitten. For each model, we generate four variants by applying spatially varying non-uniform sampling, perturbing point positions (by up to 2\% times the bounding-box side length), and injecting outliers (up to 20\%). This results in $4$ variants per model and a total of $12$ test cases in this category. Variants of the same base model are indexed by a subscript indicating the corruption level, with larger subscripts corresponding to higher levels of distortion and difficulty. In addition, we perform a controlled stress test using the Bunny model, creating $125$ test cases that span a broad range of noise levels, outlier rates, and sampling non-uniformity. Complete results for this stress test are provided in the Appendix.

To quantify the difficulty of the test models, we report three measures that capture local noise $\widehat{\sigma}$, sampling non-uniformity $\widehat{u}$, and outlier contamination $\widehat{o}$, where lower values indicate higher-quality input point clouds. The definitions of these measures are provided in the Appendix. Figure~\ref{fig:distribution} visualizes the distribution of our test models in this measure space. The models cover a wide range of difficulties, with many in the middle-to-right regions, indicating substantial noise, non-uniform sampling, and outlier rates that create challenging reconstruction scenarios.

\begin{figure*}[t]
    \centering    
    \includegraphics[height=0.13\linewidth, trim=0pt 0pt 0pt 10pt, clip]{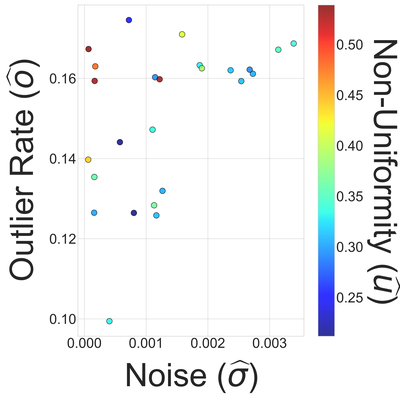}
\includegraphics[height=0.13\linewidth]{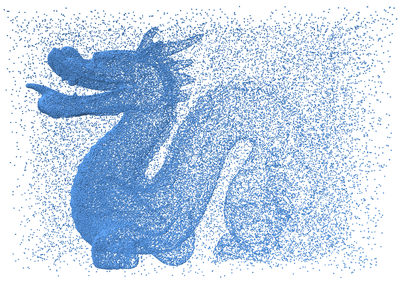}
\includegraphics[height=0.13\linewidth]{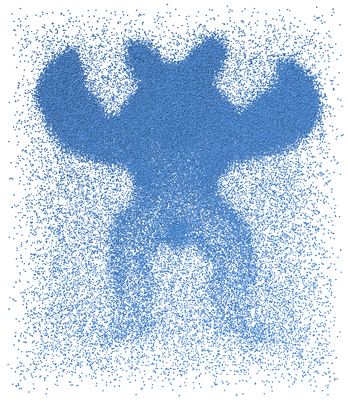}
\includegraphics[height=0.13\linewidth]{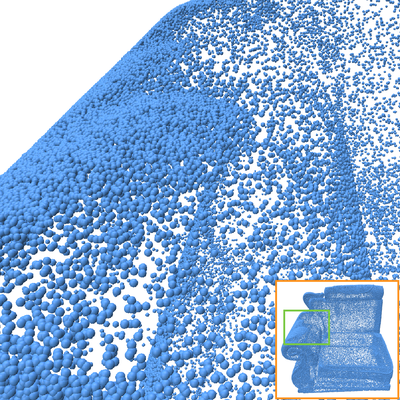}
\includegraphics[height=0.13\linewidth]{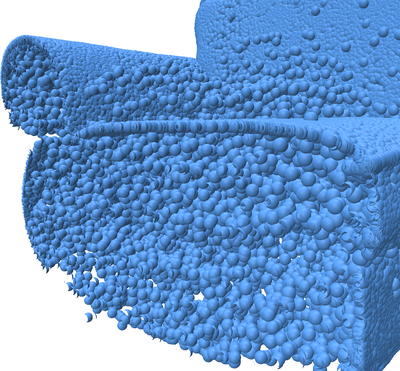}
\includegraphics[height=0.13\linewidth]{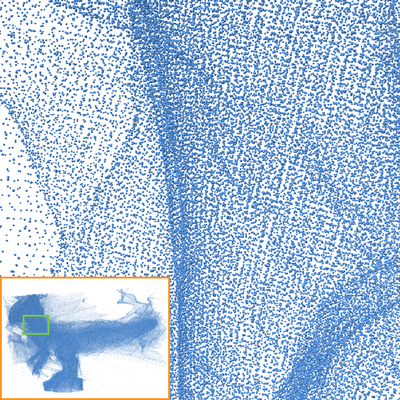}
\includegraphics[height=0.13\linewidth]{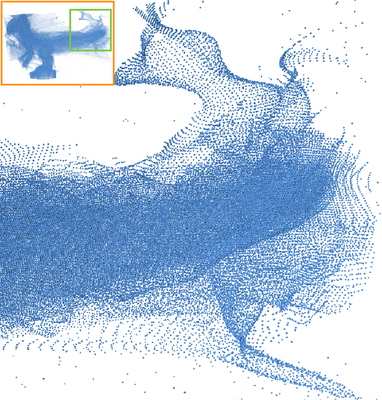}\\
  \caption{Distribution of test models in the quality-measure space $(\widehat{\sigma},\widehat{o},\widehat{u})$. The \emph{x}-axis shows the noise level $\widehat{\sigma}$, the \emph{y}-axis shows the outlier rate $\widehat{o}$, and the color map encodes the sampling non-uniformity $\widehat{u}$. Each marker corresponds to one input point cloud. Insets illustrate typical defects: uneven sampling, noise, outliers (both interior and exterior), near-surface ``thickness'', and dense but detached sheet-like fragments around the object.}
    \label{fig:distribution}
\end{figure*}

\begin{table*}[t]
\centering
\renewcommand{\arraystretch}{1.5}
\setlength{\tabcolsep}{2pt}
\begin{scriptsize}
\resizebox{\textwidth}{!}{
\begin{tabular}{c|c|c|c|c|c|c|c|c|c|c|c|c|c|c|c|c|c|c|c|c|c|c|c|c|c|c|c|c|c|c}
\toprule
\multicolumn{2}{c|}{\multirow{2}{*}{}}
& \multicolumn{5}{c|}{\textbf{Model}} 
& \multicolumn{2}{c|}{\textbf{MSP}} 
& \multicolumn{2}{c|}{\textbf{WNNC}} 
& \multicolumn{2}{c|}{\textbf{WNNC$^\star$}} 
& \multicolumn{2}{c|}{\textbf{FaCE}} 
& \multicolumn{2}{c|}{\textbf{FaCE$^\star$}}  
& \multicolumn{2}{c|}{\textbf{DWG}} 
& \multicolumn{2}{c|}{\textbf{DWG$^\star$}} 
&\multicolumn{2}{c|}{\textbf{NSH}}
&\multicolumn{2}{c|}{\textbf{NSH$^\star$}}
&\multicolumn{2}{c|}{\textbf{LoSF-UDF}}
&\multicolumn{2}{c|}{\textbf{LoSF-UDF$^\star$}}
& \multicolumn{2}{c}{\textbf{Ours}}\\
\cline{3-31} 
\multicolumn{2}{c|}{} & Name & $n$ & $\widehat{u}$ & $\widehat{\sigma}$ & $\widehat{o}$ & CD & NC & CD & NC & CD & NC 
 & CD & NC &CD &NC &CD & NC &CD &NC & CD & NC & CD & NC & CD &NC & CD & NC & CD & NC\\
\hline
\hline

\multirow{9}{*}{\rotatebox{90}{\textbf{Category}}} & \multirow{5}{*}{\rotatebox{90}{\textbf{3DGS}}} 
 & Boat  & 30,384 & 0.29 & 0.0015 & 0.17 & 43.96 & 0.70 & 5.52 & 0.86 & 9.52 & 0.85 & 5.76 & 0.93 & 17.03 & 0.76 & 19.19 & 0.69 & 15.11 & 0.77 & 6.46	& 0.86 & 9.11 & 0.88 & 14.80 & 0.80 & 15.87 & 0.89& \textcolor{red}{4.63} & \textcolor{red}{0.96} \\ \cline{3-31}
 & & Doll  & 59,237 & 0.32 & 0.0024 & 0.16 & 65.40 & 0.64 & 24.50 & 0.75 & 8.64 & 0.93 & \textcolor{red}{4.47} & 0.97 & 8.64 & 0.94 & 13.97 & 0.86 & 8.95 & 0.95 & 21.98 & 0.79 & 14.14 & 0.88 & 40.00 & 0.81 & 50.44 & 0.88 &4.78 & \textcolor{red}{0.98} \\  \cline{3-31}
 && Orna & 133,111 & 0.41 & 0.0021 & 0.17 &  28.02 & 0.83  & 27.31 & 0.72 & 11.60 & 0.91 & 11.81 & \textcolor{red}{0.95} & 15.19 & 0.92 & 24.51 & 0.79 & 19.02 & 0.93 & 17.47 & 0.81 & 21.12 & 0.85 & 24.89 & 0.80 & 33.93 & 0.88 &\textcolor{red}{9.38}	& \textcolor{red}{0.95}\\ \cline{3-31}
 && Sofa & 199,601 & 0.36 & 0.0015 & 0.13 & 24.03 & 0.81 & 25.16 & 0.74 & 11.07 & 0.89 & 13.28 & 0.94 & 17.99 & 0.86 & 18.12 & 0.82 & 24.10 & 0.82 & 14.53 & 0.85 & 18.56 & 0.81 & 16.92 & 0.85 & 29.59 & 0.92 &\textcolor{red}{8.91} & \textcolor{red}{0.95}\\ 
 \cline{3-31}
\noalign{\global\arrayrulewidth=0.7pt}\cline{3-31}
\noalign{\global\arrayrulewidth=0.4pt}

&& Mean (9 models) & 100,268 & 0.33 & 0.0012 & 0.11 & 32.44 & 0.76 & 21.37 & 0.75 & 10.30 & 0.89 & 7.46 & \textcolor{red}{0.93} & 11.56 & 0.88 &  16.25 & 0.82 & 15.52 & 0.84 & 15.68 & 0.81 & 16.48 & 0.84 & 23.72 & 0.78 & 30.83 & 0.87 & \textcolor{red}{6.74} & \textcolor{red}{0.93} \\ \noalign{\global\arrayrulewidth=1.0pt}\cline{2-31}
\noalign{\global\arrayrulewidth=0.4pt}

 & \multirow{4}{*}{\rotatebox{90}{\textbf{Graphics}}} 
  & Armadillo$_4$ & 180,000  &  0.31 & 0.0027& 0.16 & 9.04 & 0.87 &  23.45 & 0.77 &  10.57 &0.83  & 64.24&0.63 & 17.11 &0.81 &  58.13 &0.63  & 11.28 & 0.78 & 15.90 & 0.63 & 7.37 & 0.86 & 14.66 & 0.58 & 26.50 & 0.74 &\textcolor{red}{5.79}	& \textcolor{red}{0.90}\\ \cline{3-31}
 && Dragon$_4$ & 120,000  & 0.29 &0.0027	&0.16 & 10.23 & 0.89 & 12.65	&0.78 & 9.43	&0.85  & 32.29 & 0.73 & 10.51 &0.87 &  22.37 & 0.74  & 16.65 & 0.75 & 19.04& 0.66 & 8.52 & 0.84 & 10.36 & 0.64 & 33.41& 0.52&\textcolor{red}{5.13}	& \textcolor{red}{0.91}\\ \cline{3-31}
 && Kitten$_4$ & 60,000 &  0.34 &0.0034&0.17 & 30.40 & 0.90 & 18.20	&0.80 & 18.81	&0.86  & 28.99 &0.73 & 16.98 &0.92 & 31.80 & 0.68 & 28.35 & 0.88 & 18.66 & 0.74 & 17.05 & 0.84 & 83.37 & 0.53 & 64.55 &0.75&\textcolor{red}{6.10}	& \textcolor{red}{0.97}\\ \cline{3-31}

 \noalign{\global\arrayrulewidth=0.7pt}\cline{3-31}
\noalign{\global\arrayrulewidth=0.4pt}
&& Mean (12 models) & 118,750 & 0.38 & 0.0015 & 0.16 & 14.03 & 0.92 & 25.08 & 0.73& 8.90 & 0.90 & 34.85 & 0.73 & 10.07& 0.93 & 36.26 & 0.67 & 13.34 & 0.89 & 44.19 & 0.56 & 5.96 & 0.67 & 14.07 & 0.53 & 16.30 & 0.56 & \textcolor{red}{4.11} & \textcolor{red}{0.96} \\ 
\bottomrule

\end{tabular}}     
\end{scriptsize}
\caption{Statistics on test models from the 3DGS category and graphics benchmarks, where ground truth meshes are available. Here $n$ is the number of input points, and $\widehat{u}$, $\widehat{\sigma}$, and $\widehat{o}$ are quality measures of sampling non-uniformity, noise level, and outlier rate, respectively. Lower values indicate higher-quality input point clouds. A method name marked with $^\star$ indicates that outlier removal (PointCleanNet) and denoising (PCDNF) are applied as preprocessing. Chamfer distances (CD, $\downarrow$) are multiplied by $10^{3}$, and normal consistency (NC, $\uparrow$) is reported as cosine similarity, where $1$ indicates perfect agreement. The best results are highlighted in \textcolor{red}{red}. Due to space constraints, the table shows results on seven representative models, and reports category-wise average metrics. More results are provided in the Appendix.  }

 \label{tab:statistics}
\end{table*}







\subsection{Baselines} 
\label{subsec:baselines}

\begin{figure*}[t]
  \centering
  \def\mycropparams{trim=20pt 20pt 20pt 20pt, clip}
  \includegraphics[width=0.1500\linewidth]{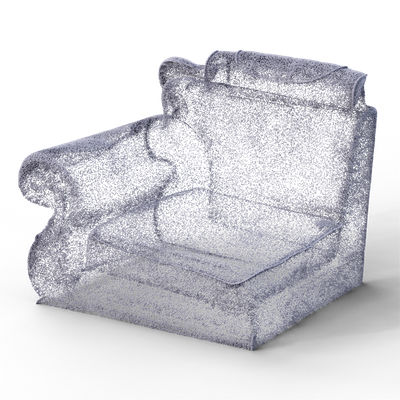}\hfill
  \includegraphics[width=0.1500\linewidth]{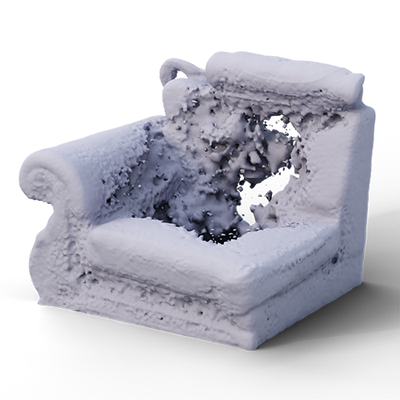}\hfill
  \includegraphics[width=0.1500\linewidth]{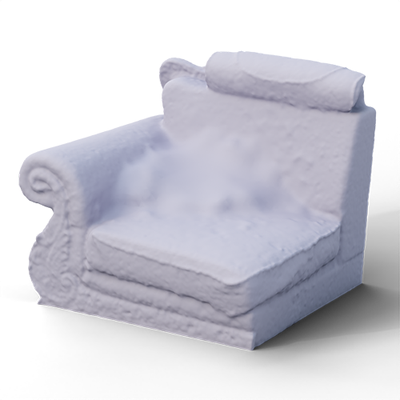}\hfill
  \includegraphics[width=0.1500\linewidth]{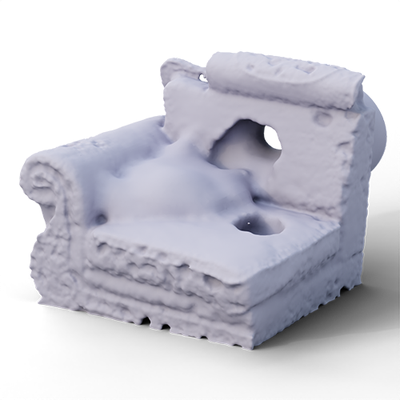}\hfill
  \includegraphics[width=0.1500\linewidth]{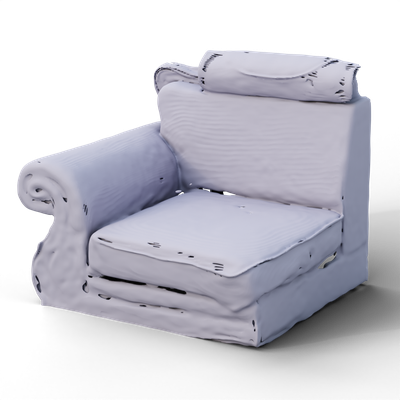}\hfill
  \includegraphics[width=0.1500\linewidth]{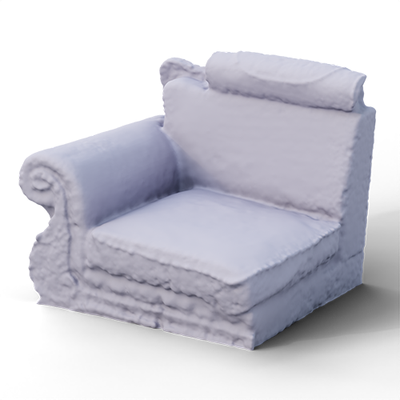}\\
  \includegraphics[width=0.1500\linewidth]{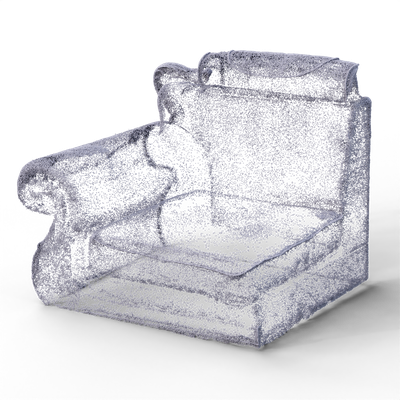}\hfill
  \includegraphics[width=0.1500\linewidth]{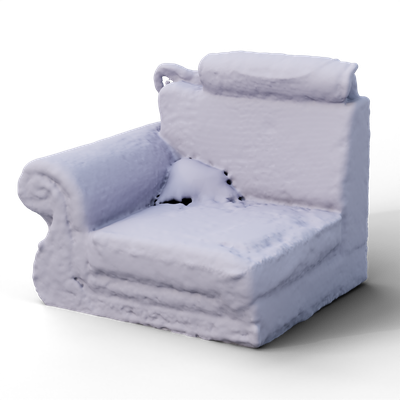}\hfill
  \includegraphics[width=0.1500\linewidth]{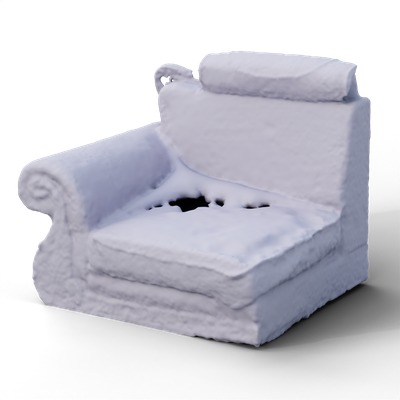}\hfill
  \includegraphics[width=0.1500\linewidth]{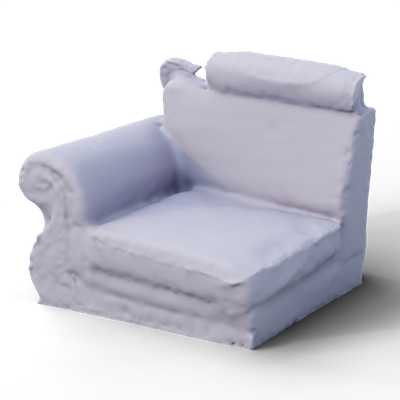}\hfill
  \includegraphics[width=0.1500\linewidth]{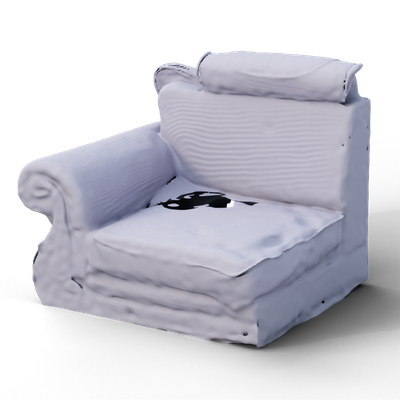}\hfill 
  \includegraphics[width=0.1500\linewidth]{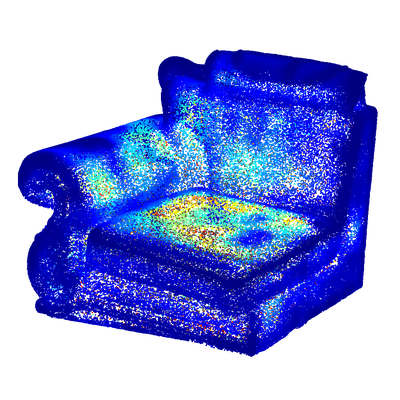}\\

  \includegraphics[width=0.1500\linewidth]{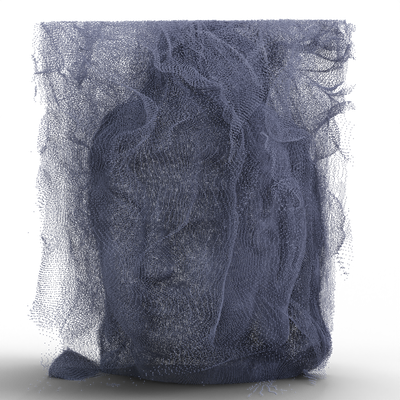}\hfill
    \includegraphics[width=0.1500\linewidth]{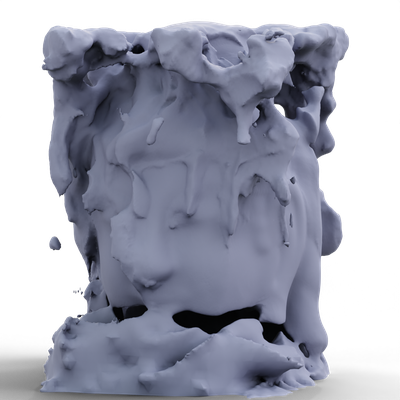}\hfill
\includegraphics[width=0.1500\linewidth]{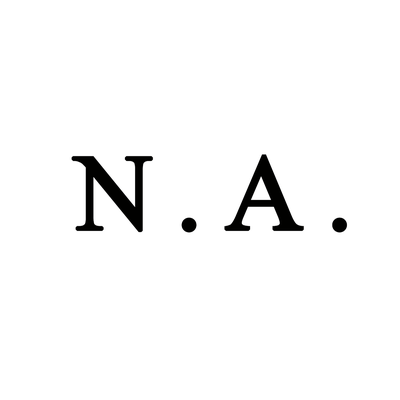}\hfill
  \includegraphics[width=0.1500\linewidth]{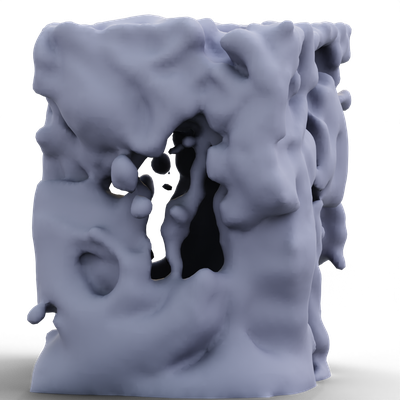}\hfill  
  \includegraphics[width=0.1500\linewidth]{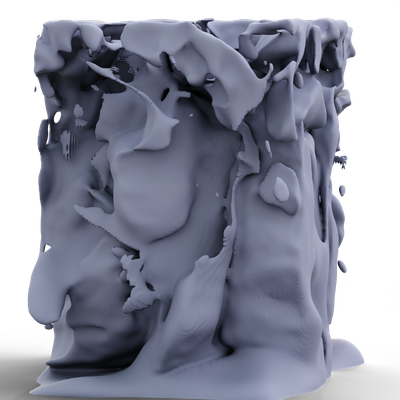}\hfill
  \includegraphics[width=0.1500\linewidth]{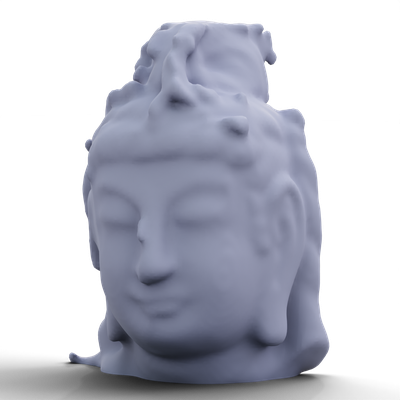}\\
  \includegraphics[width=0.1500\linewidth]{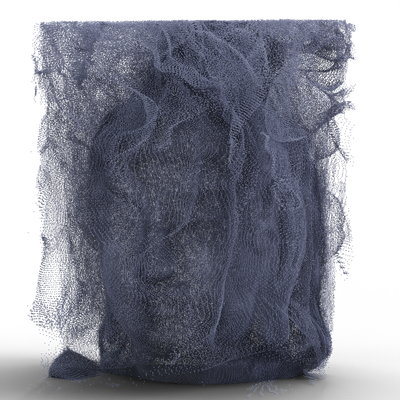}\hfill
  \includegraphics[width=0.1500\linewidth]{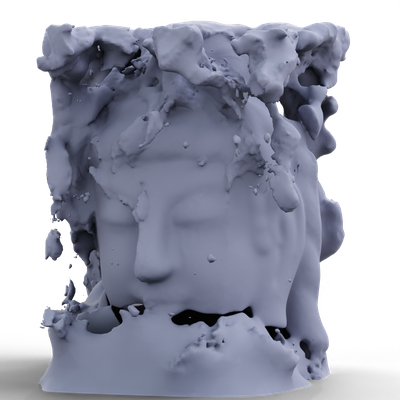}\hfill
\includegraphics[width=0.1500\linewidth]{fig/results_render/na.png}\hfill
  \includegraphics[width=0.1500\linewidth]{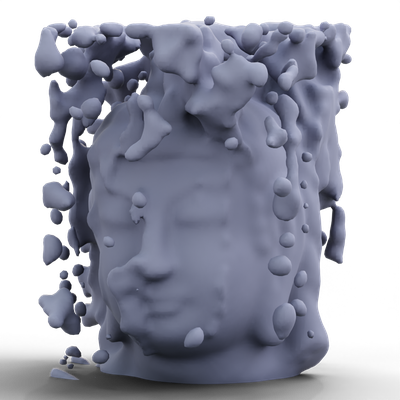}\hfill
  \includegraphics[width=0.1500\linewidth]{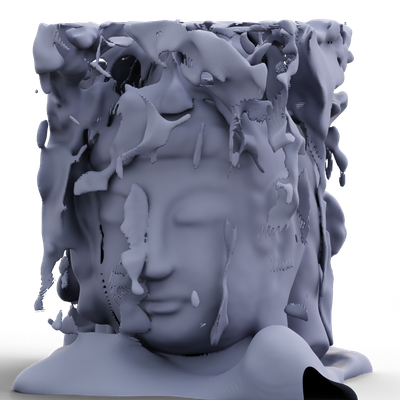}\hfill
  \includegraphics[width=0.1500\linewidth]{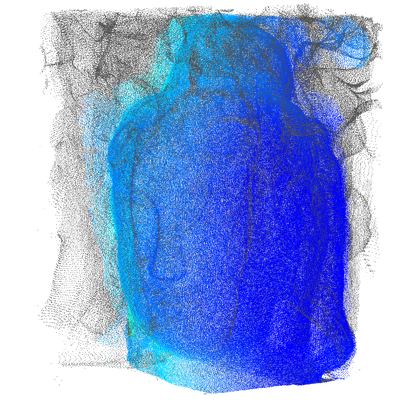}\\
 
\includegraphics[width=0.1500\linewidth, trim=0pt 50pt 0pt 50pt, clip]{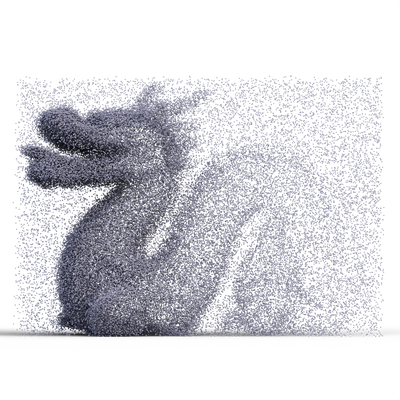}\hfill
    \includegraphics[width=0.1500\linewidth, trim=0pt 50pt 0pt 50pt, clip]{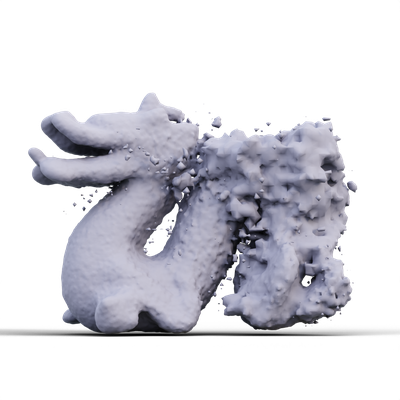}\hfill
  \includegraphics[width=0.1500\linewidth, trim=0pt 50pt 0pt 50pt, clip]{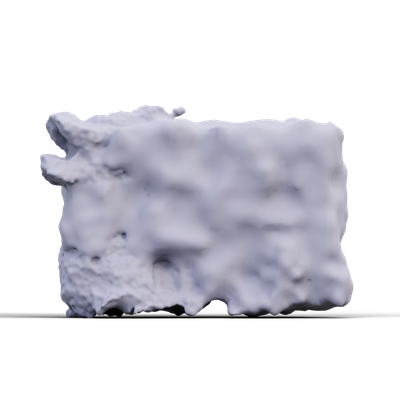}\hfill
  \includegraphics[width=0.1500\linewidth, trim=0pt 50pt 0pt 50pt, clip]{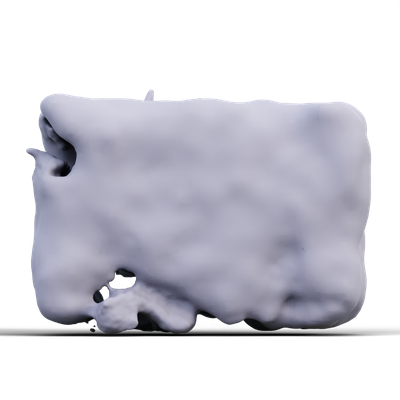}\hfill
  \includegraphics[width=0.1500\linewidth, trim=0pt 50pt 0pt 50pt, clip]{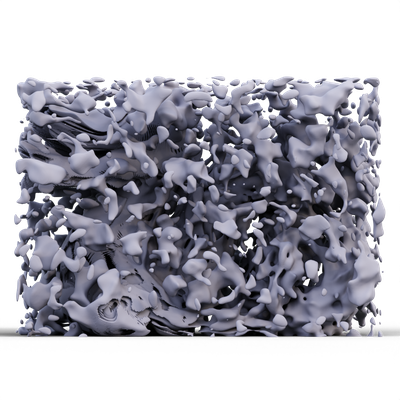}\hfill
  \includegraphics[width=0.1500\linewidth, trim=0pt 50pt 0pt 50pt, clip]{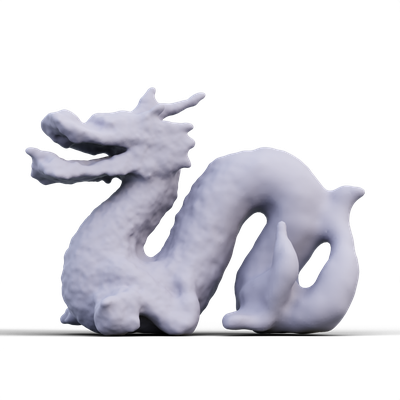}\\
  \includegraphics[width=0.1500\linewidth, trim=0pt 50pt 0pt 50pt, clip]{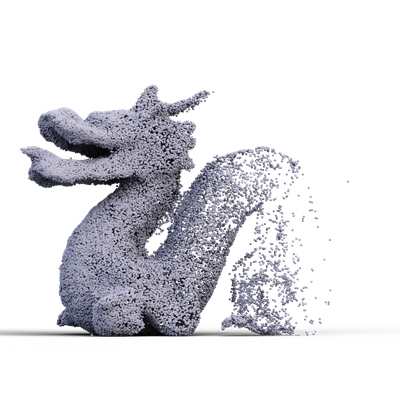}\hfill
  \includegraphics[width=0.1500\linewidth, trim=0pt 50pt 0pt 50pt, clip]{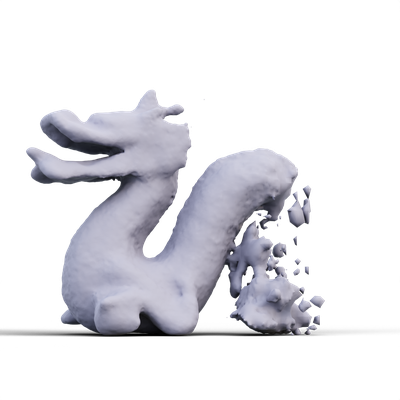}\hfill
\includegraphics[width=0.1500\linewidth, trim=0pt 50pt 0pt 50pt, clip]{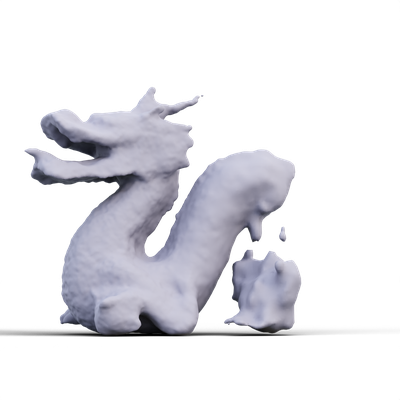}\hfill
  \includegraphics[width=0.1500\linewidth, trim=0pt 50pt 0pt 50pt, clip]{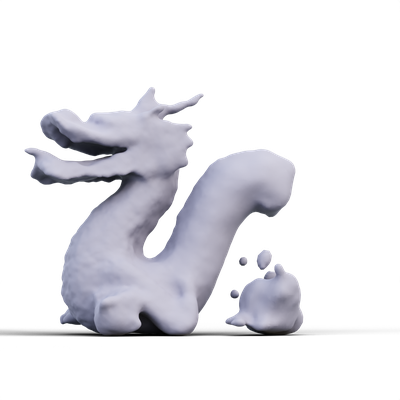}\hfill
  \includegraphics[width=0.1500\linewidth, trim=0pt 50pt 0pt 50pt, clip]{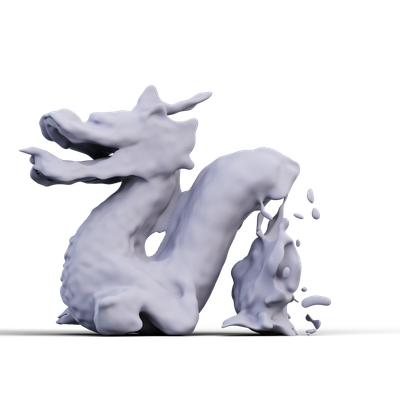}\hfill
\includegraphics[width=0.1500\linewidth, trim=0pt 50pt 0pt 50pt, clip]{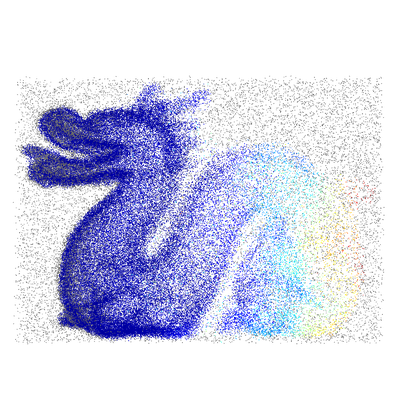}\\
  
   \begin{small}\parbox{0.1500\linewidth}{\centering\small Input/Filtered pts}\hfill
 \parbox{0.1500\linewidth}{\centering\small WNNC}\hfill  
  \parbox{0.1500\linewidth}{\centering\small FaCE}\hfill
  \parbox{0.1500\linewidth}{\centering\small DWG}\hfill 
  \parbox{0.1500\linewidth}
  {\centering\small NSH}\hfill 
  \parbox{0.1500\linewidth}{\centering\small DiWR/Optimized pts}\hfill \\
  \end{small}
  \caption{Representative qualitative results on point clouds extracted from (top) 3DGS primitives, (middle) a computer-vision pipeline (VGGT), and (bottom) a degraded graphics benchmark. The leftmost column shows the raw input and the preprocessed point cloud after outlier removal and denoising. For each baseline, we report results on both the corrupted input (top row) and the preprocessed input (bottom row). FaCE is unable to produce reliable results on the VGGT point cloud due to the large number of input points.
  For \method{}, we also visualize the effective weights $a_i c_i$ as heatmaps; in sparsely sampled regions, inliers are assigned higher optimized area weights, compensating for low sampling density. All visualizations are rendered at high resolution to support close-up inspection; see the Appendix for more results. }
  \label{fig:results} 
\end{figure*}

\begin{figure}[t]
\centering
\includegraphics[width=1.600in]{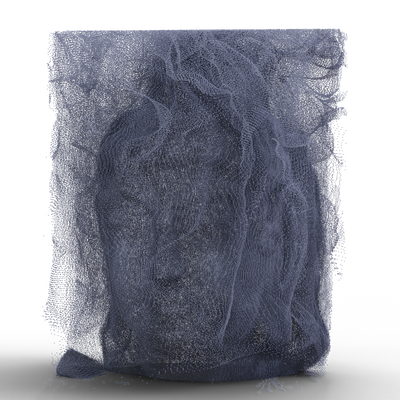}
\includegraphics[width=1.600in]{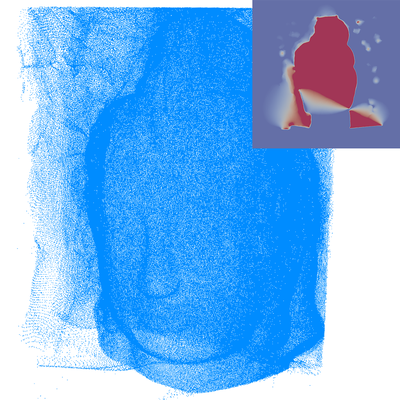}
\includegraphics[width=1.600in]{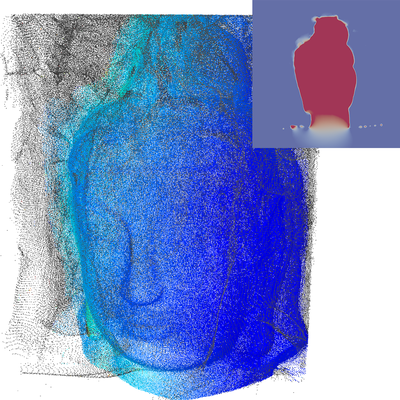}
\includegraphics[width=1.600in]{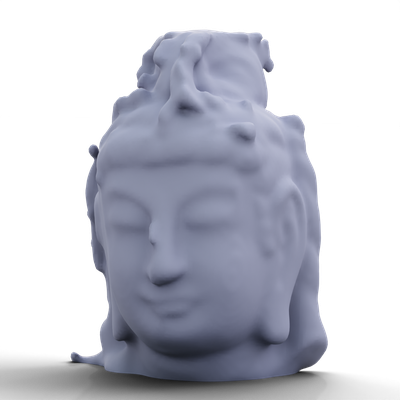}\\
\makebox[1.60in]{(a)}
\makebox[1.60in]{(b)}
\makebox[1.60in]{(c)}
\makebox[1.60in]{(d)}\\
\caption{Suppressing dense spurious sheets in a VGGT point cloud ($n=962,492
, \widehat{\sigma}=7.2\times10^{-4}, \widehat{o}=0.17, \widehat{u}=0.25$).
(a) The input contains dense near-surface layers and detached sheet-like fragments that are locally coherent and can be mistaken as true surface evidence by baseline methods. Because these sheets do not form closed boundary surfaces, they are inconsistent with a smooth winding field and can be suppressed by \method{}.
(b) Initializing all points with uniform confidence coefficients and weights, DWG produces a winding field with sharp local variations around these structures, resulting in a high discrete Dirichlet energy $\widehat{E}_{\text{diri}}= 3.35$. (c) After \method{} optimization, points on isolated sheets are assigned low confidence and small effective weights $a_ic_i$, and the winding field becomes smoother with a reduced energy $\widehat{E}_{\text{diri}} = 0.22$. (d) As a result, the final reconstructed surface contains substantially fewer artifacts caused by these spurious sheets.}
    \label{fig:sheets}
\end{figure}

We compare \method{} with (i) a traditional multi-stage pipeline (MSP) that performs denoising/filtering, normal orientation, and reconstruction from oriented points, (ii) recent methods that jointly solve normal orientation and surface reconstruction from unoriented inputs, and (iii) deep learning methods, including both supervised and self-supervised approaches.

For the multi-stage pipeline, we use a state-of-the-art method at each stage. Specifically, for outlier removal we use PointCleanNet~\cite{PointCleanNet}, a learning-based approach that predicts an outlier label per point and removes the detected outliers. For denoising, we use PCDNF~\cite{PCDNF}, an end-to-end network that denoises point clouds while jointly filtering and estimating normals to better preserve geometric features. For normal orientation, we use Dipole~\cite{dipole}, which first predicts locally coherent normal directions within patches and then propagates orientations globally via dipole propagation to obtain globally consistent normals. Finally, given the filtered and oriented points, we reconstruct a watertight surface using sPSR~\cite{kazhdan2013screened}.

We also include recent joint orientation-reconstruction baselines, including WNNC~\cite{lin2024wnnc}, DWG~\cite{liu2025diffusing}, and FaCE~\cite{faraday}, to evaluate the benefit of jointly optimizing orientation, area weights, and confidence within a unified framework. Since these methods are not primarily designed for severe noise and high outlier rates, we additionally report results with PointCleanNet and PCDNF as preprocessing to reduce their sensitivity to corrupted inputs. For completeness, we also report results without preprocessing.

Deep learning methods are often limited to small- to middle-scale point clouds due to their high GPU memory demands during training and inference. 
We compare against two representative methods. NSH~\cite{nsh} is a self-supervised approach that fits a neural signed distance field directly from unoriented point clouds; beyond the standard first-order Eikonal regularization, it enforces a singular-Hessian constraint to stabilize training and suppress spurious geometry. LoSF-UDF~\cite{losf} is a supervised framework that learns an unsigned distance field (UDF) from local shape functions. It trains a lightweight attention-based network on synthetic, smooth local patches with known ground-truth distances, enabling the predictor to infer UDF values from a fixed-radius neighborhood around each query point. This local, patch-based formulation reduces reliance on global topology and encourages the network to focus on locally consistent geometric cues, which makes LoSF-UDF comparatively resilient to moderate noise and sparse outliers, provided that each neighborhood still contains sufficient inlier support.



\subsection{Comparison}
\label{subsec:comparison}

Table~\ref{tab:statistics} reports quantitative results for the 3DGS and graphics benchmark categories, where ground-truth meshes are available. This allows us to evaluate reconstruction quality using standard accuracy metrics such as Chamfer Distance (CD) and Normal Consistency (NC). For models from the computer vision pipeline category, where ground truth is unavailable, we instead provide qualitative results in Figure~\ref{fig:results}, with additional visualizations included in the Appendix. Overall, \method{} achieves consistently strong performance across all three model categories, whereas no baseline attains comparable results across all categories. We show representative results in Figures~\ref{fig:teaser} and~\ref{fig:results}, and we provide a detailed analysis below. More results are included in the Appendix.

For the 3DGS category, our results show that interior outliers and non-uniform sampling pose major challenges to GWN-based methods such as WNNC and DWG. Interior outliers corrupt the discrete winding-number integration, which ideally aggregates contributions only from boundary samples, thereby biasing the induced field and its gradients. In addition, both WNNC and DWG are sensitive to non-uniform sampling and tend to  degrade in sparsely sampled regions. Although \method{} is also based on GWN, it explicitly optimizes per-point area weights and confidence coefficients. The confidence coefficients suppress interior outliers by down-weighting their contributions to the winding field, while the area weights compensate for non-uniform sampling by rebalancing the discrete integration. As a result, \method{} improves robustness to both interior outliers and sampling irregularity. FaCE addresses the issue of interior outliers through a different mechanism: it simulates a Faraday-cage effect to create a conductive enclosure that shields the interior from external fields. This makes FaCE inherently robust to interior contamination, enabling it to reconstruct complete geometry. Quantitatively, averaged over nine 3DGS models, FaCE achieves performance comparable to \method{} in terms of CD and NC. Qualitatively, however, \method{} often produces cleaner surfaces with fewer artifacts (see Figure~\ref{fig:results} (top) and the Appendix).

Point clouds produced by VGGT often contain dense near-surface layers and detached sheet-like fragments. Although these structures are locally coherent and can appear plausible, they often fail to represent a closed boundary and therefore do not support a globally consistent inside-outside indicator. When incorporated into the discrete winding-number construction, such fragments introduce  sharp spatial variations in the winding field and increase its (discrete) Dirichlet energy. By explicitly regularizing the winding field via Dirichlet energy and jointly optimizing the confidence coefficients, \method{} reduces the influence of these fragments by driving their confidences toward $0$, which shrinks their effective weights$a_i c_i$. This allows the reconstruction to focus on the subset of samples that collectively supports a smooth, globally consistent winding field, resulting in substantially fewer sheet-induced artifacts in the final watertight surface (Figure~\ref{fig:sheets}). In our experiments, baseline methods are particularly sensitive to these near-surface layers and detached sheets; even with outlier removal and denoising as preprocessing, residual fragments often remain and continue to bias the reconstruction.


For the graphics benchmarks, the injected non-uniform sampling, measurement noise, and high outlier rates make surface reconstruction particularly challenging. When applied directly to these corrupted inputs, all baseline methods fail to recover coherent surfaces. Pre-filtering the points substantially improves their outputs; however, as discussed above, any errors introduced during preprocessing (e.g., imperfect outlier removal or over-aggressive denoising) are effectively irreversible and cannot be corrected in subsequent orientation and/or reconstruction stages. Consequently, the baselines still often produce reconstructions with missing regions or severe geometric distortions. In contrast, our joint optimization removes the need for a separate preprocessing step and yields visually plausible surfaces  under these challenging conditions.

\section{Conclusion}
\label{sec:conclusion}

We presented \method{}, a robust method for reconstructing 3D watertight surfaces from unoriented point clouds with imperfections such as uneven sampling, noise, and outliers. In contrast to existing pipelines that treat preprocessing, normal orientation and surface reconstruction as separate steps, \method{} solves these subproblems jointly: it optimizes point orientations together with per-point confidence coefficients and adaptive area weights, coupled through the induced GWN field in a unified formulation. Extensive experiments demonstrate that this joint optimization improves robustness on challenging inputs and produces visually plausible
watertight surfaces across a wide range of conditions.

\bibliographystyle{unsrtnat}
\bibliography{references}  

\appendix
\section*{Appendix}
In the Appendix, we introduce quantitative model-quality measures (Section~\ref{sec:measures}), recommend typical parameter settings based on input corruption levels (Section~\ref{sec:parameter-settings}), and present extensive stress tests to evaluate the robustness of \method{} (Section~\ref{sec:stress}). We also discuss several design choices in \method{} (Section~\ref{sec:discussions}), report runtime and memory consumption (Section~\ref{sec:runtime}), provide full per-model statistics, and include additional qualitative comparisons (Section~\ref{sec:additionalresults}). Finally, we discuss the limitations of our method (Section~\ref{sec:limitations}).

\section{Model-Quality Measures}
\label{sec:measures}

To quantitatively characterize the imperfections of our test inputs and to stratify their difficulty, we define three model-quality measures that capture (i) noise/roughness, (ii) sampling non-uniformity, and (iii) outlier contamination. 

All measures are computed from simple statistics of the input point set. We uniformly scale each model into a unit cube. For each point $\mathbf{p}_i$, let $\mathcal{N}_{i,k}$ denote its $k$-nearest neighbors, and let $\Pi_i$ be the least-squares best-fit plane to $\mathcal{N}_{i,k}$. We use $k\in[10,40]$ in practice.

We first estimate a global sampling scale from the average $k$NN distance:
\[
s_i := \frac{1}{k}\sum_{\mathbf{q}\in \mathcal{N}_{i,k}} \|\mathbf{p}_i-\mathbf{q}\|,
\qquad \widehat{s} := \operatorname{median}_i\, s_i.
\]
Here, $s_i$ measures the local inter-point spacing around $\mathbf{p}_i$, and $\widehat{s}$ provides a robust estimate of the typical spacing of the input.

We then measure local noise (or roughness) using the plane-fit residual:
\[
\sigma_i := \sqrt{\frac{1}{k}\sum_{\mathbf{q}\in \mathcal{N}_{i,k}}
\!\operatorname{dist}\!\big(\mathbf{q},\Pi_i\big)^2},
\qquad \widehat{\sigma} := \operatorname{median}_i\, \sigma_i.
\] 
The median $\widehat{\sigma}$ summarizes the overall noise/roughness level.

To quantify global sampling irregularity, we measure the relative dispersion of the local scales $\{s_i\}$. Standard dispersion metrics can be dominated by extreme density peaks (e.g., overlapping points) or by distant outliers, so we use a trimmed coefficient of variation. 
Let $s_{(1)} \le s_{(2)} \le \dots \le s_{(n)}$ be the sorted values of $\{s_i\}$, and let $\mathcal{S}_{\tau}$ be the subset obtained by discarding the bottom and top $\tau\%$ values (we use $\tau=10$). We then define 
\[
\widehat{u} := \frac{\sigma(\mathcal{S}_{\tau})}{\mu(\mathcal{S}_{\tau})},
\]
where $\mu(\cdot)$ and $\sigma(\cdot)$ are the sample mean and standard deviation. Smaller $\widehat{u}$ indicates more uniform sampling, while larger values reflect stronger density variation.

Finally, we estimate outlier contamination from the global distribution of $\{s_i\}$, following the statistical outlier removal idea~\cite{rusu2008towards}. We compute the global mean $\mu_s$ and standard deviation $\sigma_s$ of $\{s_i\}$, and classify points with unusually sparse neighborhoods (large $s_i$) as outliers:
\[
\widehat{o} := \frac{\#\{\, i \mid s_i > \mu_s + 2 \sigma_s \,\}}{n},
\]
where $n$ is the number of input points. This measure serves as a simple proxy for the fraction of points that are isolated or lie in low-density regions relative to the bulk of the data.

\section{Parameter Settings}
\label{sec:parameter-settings}

Our algorithm has several user-specified parameters: the area-weight convergence threshold $\epsilon_a$, the normal-update convergence threshold $\epsilon_n$, the maximum number of iterations $t_{\max}$, and the weights $\lambda_i$ ($i=1,\ldots,5$) used in the area-weight optimization (Eq. (9)) and confidence optimization (Eq. (10)). 

We use the following \emph{model-independent} defaults in all experiments: $\epsilon_a=0.15$, $\epsilon_n=0.02$, and $t_{\max}=10$. The remaining parameters are set based on the three model-quality measures $\widehat{\sigma}$, $\widehat{o}$, and $\widehat{u}$, of the input point cloud, which provide a global summary of noise/roughness, sampling non-uniformity, and outlier contamination for each model.

We fix the weight of the discrete Dirichlet-energy term to $1.0$. The remaining weights  $\lambda_i$ in Eq.~(9) and Eq.~(10) are scheduled during optimization to improve  stability. Specifically, we start with relatively small weights $\lambda_i (i=1,\cdots,5)$, because in early iterations the orientations, area weights, and confidence coefficients  are still inaccurate and the winding field benefits from  stronger smoothness regularization. As the solution stabilizes, we gradually increase these weights $\lambda_i$ to strength the influence of constraint terms. For easy to moderate models with $\widehat{\sigma}\leq 0.002$, $\widehat{u}\leq 0.3$ and $\widehat{o}\leq 0.08$, we use the initial weights  $\lambda_1=5.0$, $\lambda_2=1.0$, $\lambda_3=1.0$, $\lambda_4=0.5$, and $\lambda_5=5\times10^{-3}$. For more severely corrupted inputs, we recommend smaller initial weights (e.g., halving the above values) so that the Dirichlet-energy regularization remains dominant for longer.  

\section{Stress Tests}
\label{sec:stress}

\begin{figure}[!htbp]
    \centering
\includegraphics[width=1.0in]{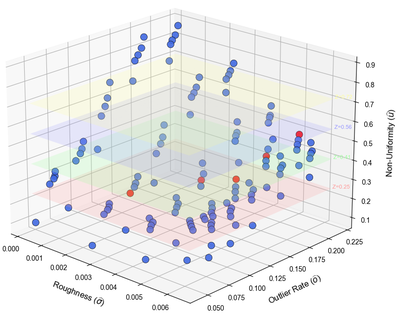}
    \includegraphics[width=1.0in]{fig/teaser/1/points.png}
    \includegraphics[width=1.0in]{fig/teaser/2/points.png}
    \includegraphics[width=1.0in]{fig/teaser/3/points.png}
    \includegraphics[width=1.0in]{fig/teaser/4/points.png}
    \includegraphics[width=1.0in]{fig/teaser/5/points.png}\\
     \makebox[1.0in]{}
    \makebox[1.0in]{$\widehat{o}=0.089$}
    \makebox[1.0in]{$\widehat{o}=0.16$}
    \makebox[1.0in]{$\widehat{o}=0.14$}
    \makebox[1.0in]{$\widehat{o}=0.17$}
    \makebox[1.0in]{$\widehat{o}=0.21$}\\
     \makebox[1.0in]{}
    \makebox[1.0in]{$\widehat{\sigma}=0.0025$}
    \makebox[1.0in]{$\widehat{\sigma}=0.0043$}
    \makebox[1.0in]{$\widehat{\sigma}=0.0052$}
    \makebox[1.0in]{$\widehat{\sigma}=0.0053$}
    \makebox[1.0in]{$\widehat{\sigma}=0.0053$}\\
     \makebox[1.0in]{}
    \makebox[1.0in]{$\widehat{u}=0.28$}
    \makebox[1.0in]{$\widehat{u}=0.53$}
    \makebox[1.0in]{$\widehat{u}=0.35$}
    \makebox[1.0in]{$\widehat{u}=0.46$}
    \makebox[1.0in]{$\widehat{u}=0.50$}\\         
\caption{Stress tests on the Bunny model with varying noise level $\widehat{\sigma}$, outlier rate $\widehat{o}$, and sampling non-uniformity $\widehat{u}$. We visualize the distribution of the $125$ test cases in the 3D measure space $(\widehat{\sigma},\widehat{o},\widehat{u})$: cases near the origin correspond to low noise, low outlier rates, and nearly uniform sampling, while cases toward the far corner are the most challenging. We also highlight five representative test cases (in red), approximately sampled along the main diagonal of the space, to illustrate increasing levels of corruption from easy to hard.}
\label{fig:stressdistribution}
\end{figure}

\paragraph{Setup}
To thoroughly evaluate the robustness of \method{} under increasing corruption, we conduct controlled stress tests on the Bunny model. Starting from a clean point cloud, we progressively increase the difficulty by varying sampling non-uniformity $\widehat{u}$, noise level $\widehat{\sigma}$, and outlier rate $\widehat{o}$. We sample each factor at five levels, yielding $5^3=125$ test cases in total. These resulting models cover the ranges $\widehat{\sigma}\in[0.00018, 0.0064]$, $\widehat{o}\in[0.042,0.22]$, and $\widehat{u}\in[0.10,0.87]$ in a roughly uniform manner.  Figure~\ref{fig:stressdistribution} shows the distribution of these test cases in the quality-measure space $(\widehat{\sigma},\widehat{o},\widehat{u})$. 

\paragraph{Easy Cases} When sampling non-uniformity is mild ($\widehat{u} \leq 0.3$), the location disturbance is small ($\widehat{\sigma} \leq 0.002$), and the outlier rate is low ($\widehat{o} \leq 0.08$), \method{} preserves fine-scale details and produces clean watertight reconstructions. In this regime, most points receive high confidence, and the optimized solution stays close to the input geometry.

\paragraph{Moderate Cases}
As sampling becomes more uneven ($0.3<\widehat{u} \leq 0.7$) and both noise and outlier levels increase ($0.002<\widehat{\sigma} \leq 0.005$ and $0.08<\widehat{o}\leq 0.17$), the estimated confidence coefficients become increasingly bimodal: reliable samples concentrate near high confidence, whereas outliers (and inconsistent measurements) are pushed toward low confidence. Consequently, points that are incompatible with the recovered surface are down-weighed, while consistent inliers continue to guide orientation and reconstruction. At this corruption level, baseline methods often exhibit typical failure modes such as locally flipped regions, spurious surface components, or missing regions and over-smoothing caused by aggressive preprocessing, whereas \method{} still produces coherent watertight surfaces.

\paragraph{Difficult Cases} When  sampling becomes highly uneven ($\widehat{u}\geq 0.7$) or corruption is severe ($\widehat{\sigma} \geq 0.005$ or $\widehat{o} \geq 0.17$), the input may no longer contain sufficient reliable geometric evidence to support accurate reconstruction. In such cases, \method{} may produce incomplete or overly simplified surfaces, or fail to recover the correct structure. 

\begin{figure*}
    \centering
    \includegraphics[width=0.60in]{fig/teaser/1/points.png}
    \includegraphics[width=0.60in]{fig/teaser/2/points.png}
    \includegraphics[width=0.60in]{fig/teaser/3/points.png}
    \includegraphics[width=0.60in]{fig/teaser/4/points.png}
    \includegraphics[width=0.60in]{fig/teaser/5/points.png}
    \includegraphics[width=0.60in]{fig/teaser/1/points_msp.png}
    \includegraphics[width=0.60in]{fig/teaser/2/points_msp.png}
    \includegraphics[width=0.60in]{fig/teaser/3/points_msp.png}
    \includegraphics[width=0.60in]{fig/teaser/4/points_msp.png}
    \includegraphics[width=0.60in]{fig/teaser/5/points_msp.png}\\
\makebox[0.475\linewidth]{Input points}\makebox[0.475\linewidth]{Filtered points}\\
    \includegraphics[width=0.60in]{fig/teaser/1/wnnc.png}
    \includegraphics[width=0.60in]{fig/teaser/2/wnnc.png}
    \includegraphics[width=0.60in]{fig/teaser/3/wnnc.png}
    \includegraphics[width=0.60in]{fig/teaser/4/wnnc.png}
    \includegraphics[width=0.60in]{fig/teaser/5/wnnc.png}
    \includegraphics[width=0.60in]{fig/teaser/1/wnnc_msp.png}
    \includegraphics[width=0.60in]{fig/teaser/2/wnnc_msp.png}
    \includegraphics[width=0.60in]{fig/teaser/3/wnnc_msp.png}
    \includegraphics[width=0.60in]{fig/teaser/4/wnnc_msp.png}
    \includegraphics[width=0.60in]{fig/teaser/5/wnnc_msp.png}\\
\makebox[0.475\linewidth]{WNNC}
\makebox[0.475\linewidth]{WNNC$^\star$}
\includegraphics[width=0.60in]{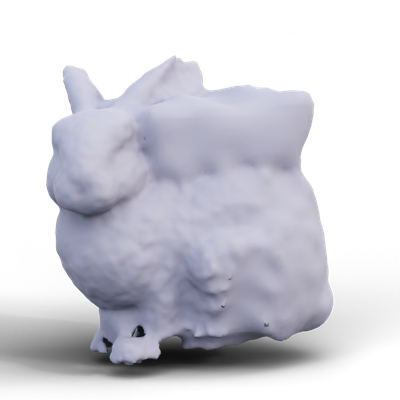}
\includegraphics[width=0.60in]{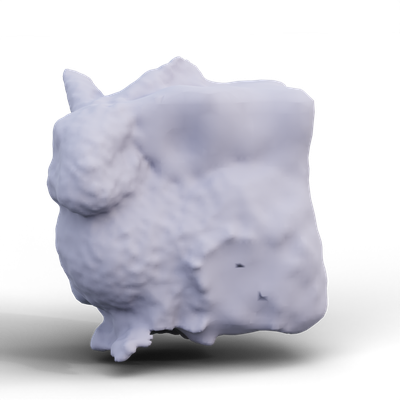}
\includegraphics[width=0.60in]{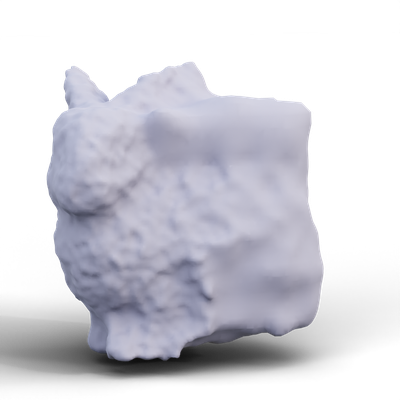}
\includegraphics[width=0.60in]{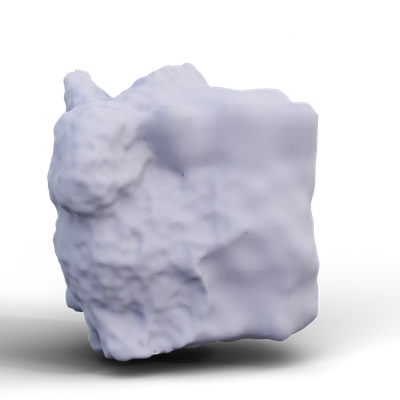}
\includegraphics[width=0.60in]{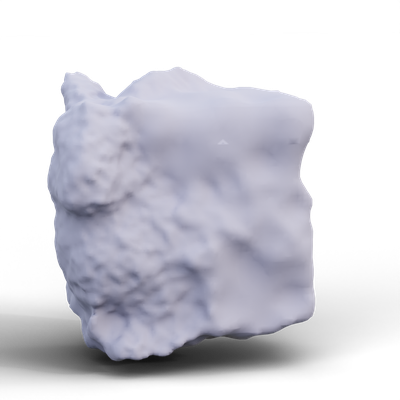}
\includegraphics[width=0.60in]{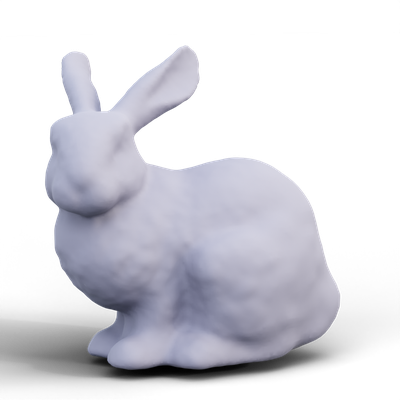}
\includegraphics[width=0.60in]{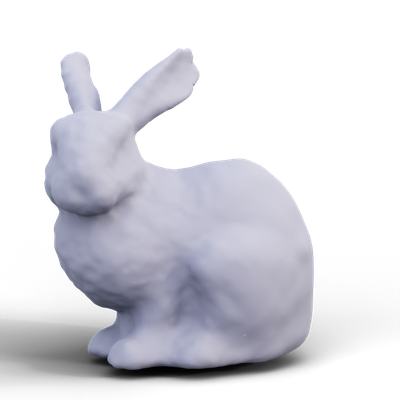}
\includegraphics[width=0.60in]{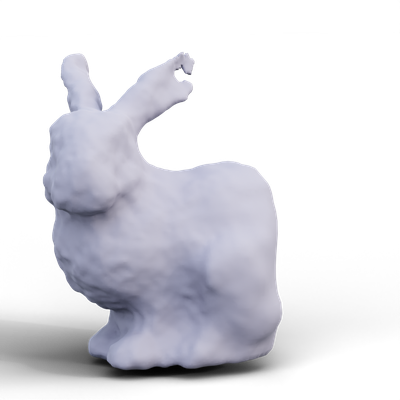}
\includegraphics[width=0.60in]{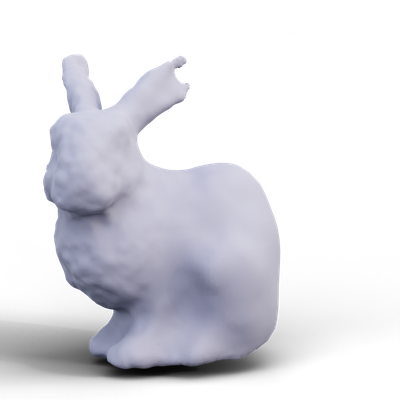}
\includegraphics[width=0.60in]{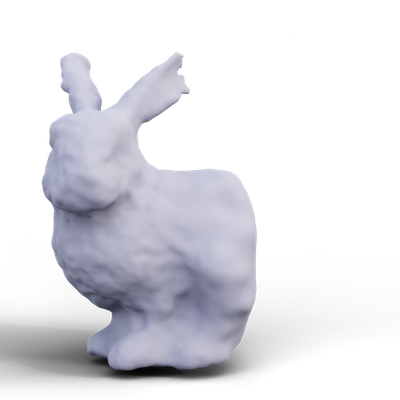}\\
\makebox[0.475\linewidth]{FaCE}
\makebox[0.475\linewidth]{FaCE$^\star$}

    \includegraphics[width=0.60in]{fig/teaser/1/ours.png}
    \includegraphics[width=0.60in]{fig/teaser/2/ours.png}
    \includegraphics[width=0.60in]{fig/teaser/3/ours.png}
    \includegraphics[width=0.60in]{fig/teaser/4/ours.png}
    \includegraphics[width=0.60in]{fig/teaser/5/ours.png}
\includegraphics[width=0.60in]{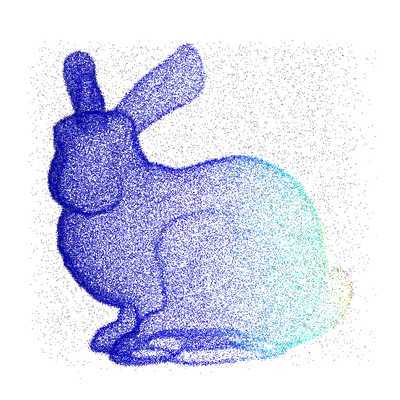}
\includegraphics[width=0.60in]{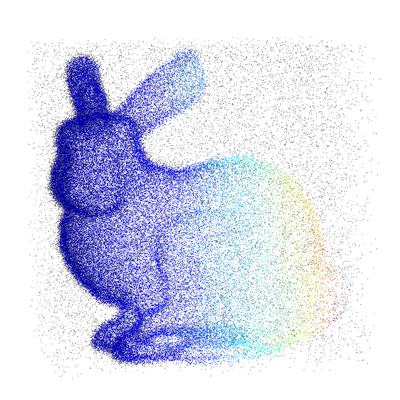}
\includegraphics[width=0.60in]{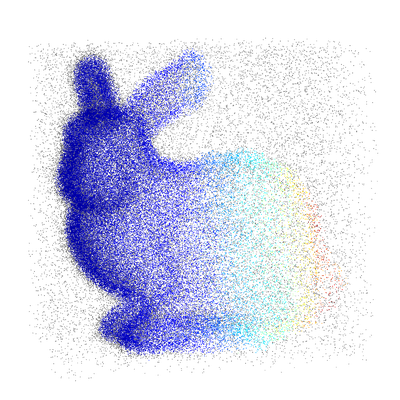}
\includegraphics[width=0.60in]{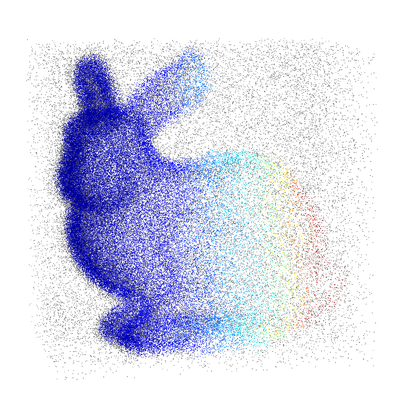}
\includegraphics[width=0.60in]{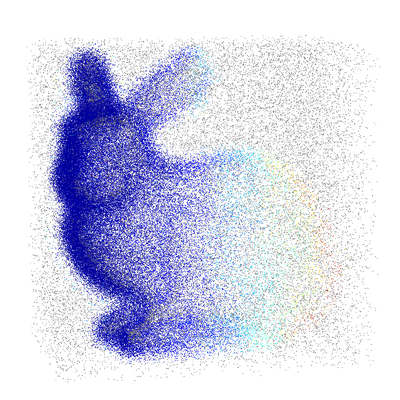}\\
\makebox[0.475\linewidth]{Ours}
\makebox[0.475\linewidth]{Visualization of $a_ic_i$}\\
\caption{Stress-test results on five representative input cases with increasing difficulty. See also the accompanying video.}    \label{fig:stressresults}
\end{figure*}

\paragraph{Summary} For each test case, we not only measure the quantitative accuracy metrics via Chamfer distance and normal consistency, we also visually inspect the reconstructed surfaces. We call a construction successful, if the overall shape is reconstructed completely, i.e., no major geometric component is missing and there is no significant distortion. Overall, \method{} successfully reconstructs 88\% of the stress-test cases, spanning a wide range of noise, outliers, and non-uniform sampling, while baselines break down at substantially lower corruption levels. These results highlight the benefit of jointly optimizing orientation and confidence: as input quality deteriorates, \method{} increasingly down-weights unreliable samples and maintains stable behavior until the available inlier evidence becomes insufficient for meaningful surface recovery. For comparison, the baseline methods' successful rates range between 20\% and 40\%  (see Fig.~\ref{fig:stressresults} for visualization results).

\section{Runtime and Memory}
\label{sec:runtime}

\method{} alternates three core components per iteration: area-weight optimization, confidence-coefficient optimization, and a DWG-based orientation update. Both optimization subproblems are solved with RMSProp and run on the GPU. The orientation update is performed with  DWG~\cite{liu2025diffusing}, which is highly parallel and also runs on the GPU. After convergence, we invoke sPSR on the CPU to extract a watertight surface from the oriented points using the effective weights $\{c_ia_i\}$.

Because the iterative stages are fully GPU-parallel, the overall runtime is dominated by these three GPU components, followed by a single CPU call to sPSR. For an input point cloud with $n=100$K points, one run of area-weight optimization, confidence optimization, and DWG-based orientation update takes approximately 30, 20, and 2 seconds, respectively.

On inputs with moderate imperfections, \method{} typically converges in less than 5 iterations. For more challenging cases, additional iterations may be required. Empirically, the orientations (and the corresponding reconstructed surfaces) change only marginally beyond 10 iterations; we therefore set  $t_{\max}=10$ as a hard stopping criterion.

The GPU memory footprint of \method{} is dominated by per-point arrays (e.g., $\mathbf{n}_i$, $a_i$, $c_i$, and their gradients), and thus scales linearly with the number of points. RMSProp introduces additional per-variable state (e.g., running averages of squared gradients) for $\mathbf{a}$ and $\mathbf{c}$, which also scales as $O(n)$. DWG has linear space complexity in practice, storing per-point quantities and its internal acceleration structures on the GPU. To evaluate the Dirichlet energy, we additionally allocate grid samples $\mathcal{Q}$ within the bounding box $\mathbb{B}$ together with their weights ${\delta_{\mathbf{q}}}$. This adds an $O(|\mathcal{Q}|)$ term. In our implementation, $|\mathcal{Q}|$ is fixed by a constant-resolution grid (typically $64^3$), so the overal memory complexity is $O(n+|\mathcal{Q}|)=O(n)$. %

\section{Discussions}
\label{sec:discussions}

\paragraph{Why are Per-point Area Weights and Confidence Coefficients Separate Variables?}
In our formulation, the generalized winding number uses the product $a_i c_i$ as the contribution of point $\mathbf{p}_i$. Although one could merge them into a single per-point weight, we keep $a_i$ (area weight) and $c_i$ (confidence) as separate variables because they encode different phenomena and should obey different priors. The area weight $a_i$ approximates the local surface element represented by $\mathbf{p}_i$ and is mainly used to compensate for non-uniform sampling; thus, $a_i$ is expected to vary smoothly and remain consistent with local spacing and a global area (or mass) budget. In contrast, the confidence coefficient $c_i\in[0,1]$ models sample reliability and is used to suppress outliers; in practice, $c_i$ often becomes sparse and close to binary. Collapsing $a_i$ and $c_i$ into a single variable would mix these roles, making it difficult to impose appropriate regularization and increasing ambiguity in the optimization. Separating them improves robustness, interpretability, and numerical stability.

\paragraph{Why Not Optimize Point Locations?}
Although optimizing point locations could further correct geometric perturbations, \method{} does not treat $\{\mathbf{p}_i\}$ as optimization variables. First, the derivative of the winding field with respect to point positions is substantially more complex than the derivatives with respect to orientations and weights, and evaluating it efficiently and stably at scale would add significant algorithmic and computational overhead. Second, \method{} builds on DWG as the reconstruction backbone, which already tolerates moderate positional noise when the per-point area weights $a_i$ are reasonably estimated. Empirically, jointly optimizing orientations, area weights, and confidence coefficients captures most of the robustness gains in our target scenarios, while keeping the optimization simple, stable, and scalable.

\section{Additional Results}
\label{sec:additionalresults}

\begin{table*}[h]
\centering
\renewcommand{\arraystretch}{1.5}
\setlength{\tabcolsep}{2pt}
\begin{scriptsize}
\resizebox{\textwidth}{!}{
\begin{tabular}{c|c|c|c|c|c|c|c|c|c|c|c|c|c|c|c|c|c|c|c|c|c|c|c|c|c|c|c|c|c|c}
\toprule
\multicolumn{2}{c|}{\multirow{2}{*}{}}
& \multicolumn{5}{c|}{\textbf{Model}} 
& \multicolumn{2}{c|}{\textbf{MSP}} 
& \multicolumn{2}{c|}{\textbf{WNNC}} 
& \multicolumn{2}{c|}{\textbf{WNNC$^\star$}} 
& \multicolumn{2}{c|}{\textbf{FaCE}} 
& \multicolumn{2}{c|}{\textbf{FaCE$^\star$}}  
& \multicolumn{2}{c|}{\textbf{DWG}} 
& \multicolumn{2}{c|}{\textbf{DWG$^\star$}} 
&\multicolumn{2}{c|}{\textbf{NSH}}
&\multicolumn{2}{c|}{\textbf{NSH$^\star$}}
&\multicolumn{2}{c|}{\textbf{LoSF-UDF}}
&\multicolumn{2}{c|}{\textbf{LoSF-UDF$^\star$}}
& \multicolumn{2}{c}{\textbf{Ours}}\\
\cline{3-31} 
\multicolumn{2}{c|}{} & Name & $n$ & $\widehat{u}$ & $\widehat{\sigma}$ & $\widehat{o}$ & CD & NC & CD & NC & CD & NC 
 & CD & NC &CD &NC &CD & NC &CD &NC & CD & NC & CD & NC & CD &NC & CD & NC & CD & NC\\
\hline
\hline

\multirow{12}{*}{\rotatebox{90}{\textbf{Category}}} & \multirow{10}{*}{\rotatebox{90}{\textbf{3DGS}}} 
 & Boat  & 30,384 & 0.29 & 0.0015 & 0.17 & 43.96 & 0.70 & 5.52 & 0.86 & 9.52 & 0.85 & 5.76 & 0.93 & 17.03 & 0.76 & 19.19 & 0.69 & 15.11 & 0.77 & 6.46	& 0.86 & 9.11 & 0.88 & 14.80 & 0.80 & 15.87 & 0.89& \textcolor{red}{4.63} & \textcolor{red}{0.96} \\ \cline{3-31}
 && Car & 100,199 & 0.30 & 0.0013 & 0.13 & 29.24 & 0.80 & 17.34 & 0.77 & 11.56 & 0.88  &  5.79 & 0.93 & 11.45 & 0.89 &  5.34 & 0.91  & 21.53 & 0.84 & 12.40 & 0.80  & 11.65 & 0.86&  11.60 & 0.77  & 15.10 & 0.86 &\textcolor{red}{4.76}	& \textcolor{red}{0.92}\\ \cline{3-31}
 & & Doll  & 59,237 & 0.32 & 0.0024 & 0.16 & 65.40 & 0.64 & 24.50 & 0.75 & 8.64 & 0.93 & \textcolor{red}{4.47} & 0.97 & 8.64 & 0.94 & 13.97 & 0.86 & 8.95 & 0.95 & 21.98 & 0.79 & 14.14 & 0.88 & 40.00 & 0.81 & 50.44 & 0.88 &4.78 & \textcolor{red}{0.98} \\  \cline{3-31}
 && Light031 & 89,462 & 0.34 & 0.0011 & 0.15 & 20.46 & 0.82 & 5.76 & 0.85 & 5.58 & 0.90  &  \textcolor{red}{4.72} & \textcolor{red}{0.94} & 5.16 & 0.91 & 6.34 & 0.89  & 9.79 & 0.78 & 7.41 & 0.83 & 5.35 & 0.92 &  6.44 & 0.83  & 9.74 & 0.89 & 4.97	& 0.92\\ \cline{3-31}
 && Light032 & 82,640 & 0.53 & 0.0012 & 0.16 & 24.68 & 0.82 & 32.91 & 0.72 &  7.55 & 0.94  &  5.07 & \textcolor{red}{0.96} & 7.56 & 0.95 &  29.10 & 0.82  & 13.41 & 0.94 & 31.77 & 0.78 & 30.58 & 0.81 &  54.08 & 0.82  & 64.66 & 0.86 &\textcolor{red}{5.03}	& \textcolor{red}{0.96}\\ \cline{3-31}
 && Orna & 133,111 & 0.41 & 0.0021 & 0.17 &  28.02 & 0.83  & 27.31 & 0.72 & 11.60 & 0.91 & 11.81 & \textcolor{red}{0.95} & 15.19 & 0.92 & 24.51 & 0.79 & 19.02 & 0.93 & 17.47 & 0.81 & 21.12 & 0.85 & 24.89 & 0.80 & 33.93 & 0.88 &\textcolor{red}{9.38}	& \textcolor{red}{0.95}\\ \cline{3-31}
 && Sofa & 199,601 & 0.36 & 0.0015 & 0.13 & 24.03 & 0.81 & 25.16 & 0.74 & 11.07 & 0.89 & 13.28 & 0.94 & 17.99 & 0.86 & 18.12 & 0.82 & 24.10 & 0.82 & 14.53 & 0.85 & 18.56 & 0.81 & 16.92 & 0.85 & 29.59 & 0.92 &\textcolor{red}{8.91} & \textcolor{red}{0.95}\\ 
 \cline{3-31}
 && Suit & 108,774 & 0.38 & 0.0025 & 0.16 & 28.58 & 0.73 & 39.44 & 0.63 & 19.21 & 0.69 & 9.88 & 0.85 & 10.20 & 0.87 & 18.39 & 0.79 & 12.21 & 0.89 & 18.76 & 0.75 & 26.56 & 0.75 & 18.17 & 0.67 & 35.25 & 0.83 &\textcolor{red}{8.26} & \textcolor{red}{0.89}\\ \cline{3-31}
 && Train & 99,008 & 0.31 & 0.0011 & 0.13 & 27.53 & 0.71 & 14.44 & 0.75 &  12.16 & 0.85  & \textcolor{red}{6.34} & \textcolor{red}{0.93} & 10.78 & 0.85 & 11.27 & 0.80  & 15.58 & 0.67 & 10.34 & 0.81  & 11.22 & 0.82 &  24.64 & 0.73  & 22.86 & 0.83 & 9.94	& 0.88\\ 
\noalign{\global\arrayrulewidth=0.7pt}\cline{3-31}
\noalign{\global\arrayrulewidth=0.4pt}

&& Mean (9 models) & 100,268 & 0.33 & 0.0012 & 0.11 & 32.44 & 0.76 & 21.37 & 0.75 & 10.30 & 0.89 & 7.46 & \textcolor{red}{0.93} & 11.56 & 0.88 &  16.25 & 0.82 & 15.52 & 0.84 & 15.68 & 0.81 & 16.48 & 0.84 & 23.72 & 0.78 & 30.83 & 0.87 & \textcolor{red}{6.74} & \textcolor{red}{0.93} \\ \noalign{\global\arrayrulewidth=1.0pt}\cline{2-31}
\noalign{\global\arrayrulewidth=0.4pt}

 & \multirow{17}{*}{\rotatebox{90}{\textbf{Graphics}}} 
   & Armadillo$_1$ & 180,000 & 0.31 & 0.0025 & 0.16 & 7.37 & 0.90 & 23.09&0.77 & 6.66 &0.89 & 60.35 &0.65 & 5.05	&0.91 & 38.29 &0.66 & 9.55 & 0.78 & 13.86 & 0.64 & 5.96 & 0.88& 22.57 & 0.59 &22.60 &0.73&\textcolor{red}{4.97} & \textcolor{red}{0.92} \\ \cline{3-31}
 && Armadillo$_2$ & 180,000 & 0.53 & 0.00017 & 0.16 & 3.28 & 0.97 & 60.69 &0.60 & 2.31 &0.98  & 43.97 & 0.69 & 2.47& \textcolor{red}{0.98} & 47.68 & 0.64 & 3.37 & 0.93 & 36.06 & 0.76& \textcolor{red}{2.27} & \textcolor{red}{0.98} & 4.58 & 0.95 & 6.53& 0.96& 2.42	& \textcolor{red}{0.98}\\ \cline{3-31}
 && Armadillo$_3$ & 172,500  & 0.30 & 0.00016 & 0.13 & 3.29 & 0.97 & 56.00 & 0.60 &  2.32	&0.97  & 45.65 & 0.71 & 2.46	& \textcolor{red}{0.98} &  50.36 & 0.60  & 3.37 & 0.94 & 29.85 & 0.81 & \textcolor{red}{2.27} & 0.97 & 4.65 & 0.95& 6.54& 0.96& 2.44	& \textcolor{red}{0.98}\\ \cline{3-31}
 && Armadillo$_4$ & 180,000  &  0.31 & 0.0027& 0.16 & 9.04 & 0.87 &  23.45 & 0.77 &  10.57 &0.83  & 64.24&0.63 & 17.11 &0.81 &  58.13 &0.63  & 11.28 & 0.78 & 15.90 & 0.63 & 7.37 & 0.86 & 14.66 & 0.58 & 26.50 & 0.74 &\textcolor{red}{5.79}	& \textcolor{red}{0.90}\\ \cline{3-31}
 
 &&  Dragon$_1$ & 120,000 & 0.31	&0.0024	&0.16 & 16.80 & 0.86 & 14.24 & 0.77 & 11.60 & 0.85 & 31.22 &0.74 & 13.11 & 0.88 & 38.97 & 0.67 & 15.90 & 0.88 & 23.88 & 0.68 & 10.47 & 0.86 & 14.29& 0.67 & 42.43&0.53&\textcolor{red}{4.67} & \textcolor{red}{0.93} \\ \cline{3-31}
 && Dragon$_2$ & 115,000 &  0.36	&0.00016	&0.13 & 17.50 & 0.88 & 20.47	&0.67 & 6.20	&0.91  & 24.68 &0.78 & 7.39 &0.96 & 26.98 &0.66 & 7.04 &0.95 & 28.47 & 0.72& 5.15 & 0.97 & 21.47& 0.76 & 30.45 & 0.55&\textcolor{red}{2.86}	& \textcolor{red}{0.98}\\ \cline{3-31}
 && Dragon$_3$ & 120,000 & 0.48 &0.00018 & 0.16 & 25.16 & 0.94 & 23.46	&0.67 & 5.02	&0.92  & 26.49 & 0.77 & 6.82 & 0.96 &  25.33 & 0.70 & 6.91 & 0.94 & 33.31 & 0.69 & 4.61 & 0.97 & 20.67 & 0.75 & 34.28 & 0.52&\textcolor{red}{3.45}	& \textcolor{red}{0.97}\\ \cline{3-31}
 && Dragon$_4$ & 120,000  & 0.29 &0.0027	&0.16 & 10.23 & 0.89 & 12.65	&0.78 & 9.43	&0.85  & 32.29 & 0.73 & 10.51 &0.87 &  22.37 & 0.74  & 16.65 & 0.75 & 19.04& 0.66 & 8.52 & 0.84 & 10.36 & 0.64 & 33.41& 0.52&\textcolor{red}{5.13}	& \textcolor{red}{0.91}\\ \cline{3-31}
 && Kitten$_1$  & 60,000 & 0.35 &0.0031 &0.17 & 30.45 & 0.91 & 15.75	&0.81 & 15.98	&0.89 & 25.43 &0.75 & 15.03 &0.92 & 32.78 & 0.66 & 28.64 & 0.87& 16.66 & 0.76 & 26.40 & 0.81 & 29.20 & 0.68 & 59.49 & 0.76 &\textcolor{red}{5.26} & \textcolor{red}{0.98} \\ \cline{3-31}
 && Kitten$_2$ & 60,000 & 0.54 & 0.000068 &0.17 & 7.10 & 0.98 & 17.93	&0.78 & 8.94	&0.94  & 18.33 &0.82 & 11.69 &0.95 & 31.93 & 0.72 & 13.43 & 0.96 & 25.16 & 0.76 & 16.13 & 0.87 & 18.21 & 0.94 & 41.78 & 0.84&\textcolor{red}{3.18}	& \textcolor{red}{0.99}\\ \cline{3-31}
 && Kitten$_3$ & 57,500  & 0.44 &0.000062 & 0.14 & 7.67 & 0.98 & 15.01	&0.80 & 8.90	&0.94  & 16.54 &0.84 & 12.22 & 0.96 & 30.53 & 0.69 & 15.65 & 0.92 & 23.08 & 0.80 & 16.29 & 0.87 & 23.41 & 0.92 & 41.40 & 0.85&\textcolor{red}{2.97}	& \textcolor{red}{0.99}\\ \cline{3-31}
 && Kitten$_4$ & 60,000 &  0.34 &0.0034&0.17 & 30.40 & 0.90 & 18.20	&0.80 & 18.81	&0.86  & 28.99 &0.73 & 16.98 &0.92 & 31.80 & 0.68 & 28.35 & 0.88 & 18.66 & 0.74 & 17.05 & 0.84 & 83.37 & 0.53 & 64.55 &0.75&\textcolor{red}{6.10}	& \textcolor{red}{0.97}\\ \cline{3-31}

\noalign{\global\arrayrulewidth=0.7pt}\cline{3-31}
\noalign{\global\arrayrulewidth=0.4pt}

&& Mean (12 models) & 118,750 & 0.38 & 0.0015 & 0.16 & 14.03 & 0.92 & 25.08 & 0.73& 8.90 & 0.90 & 34.85 & 0.73 & 10.07& 0.93 & 36.26 & 0.67 & 13.34 & 0.89 & 44.19 & 0.56 & 5.96 & 0.67 & 14.07 & 0.53 & 16.30 & 0.56 & \textcolor{red}{4.11} & \textcolor{red}{0.96} \\ 

\bottomrule

\end{tabular}
}    
\end{scriptsize}
\caption{Statistics on all test models from the 3DGS category and graphics benchmarks, where ground truth meshes are available. The best results are highlighted in \textcolor{red}{red}.}

 \label{tab:statistics_all}
\end{table*}

\begin{figure*}[htbp]
    \centering

    \newcommand{\imgwidth}{0.105\linewidth} 

    \setlength{\tabcolsep}{1pt}
    \setlength{\arrayrulewidth}{0pt}

    \newcommand{\rotlabel}[1]{%
        \rotatebox{90}{\small\textbf{#1}}%
    }

\begin{small}
    \begin{tabular}{ccccccccc} 
        \rotlabel{Images} &
        \includegraphics[width=\imgwidth, valign=m]{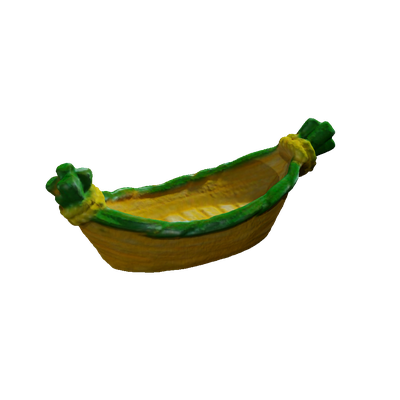} &
        \includegraphics[width=\imgwidth, valign=m]{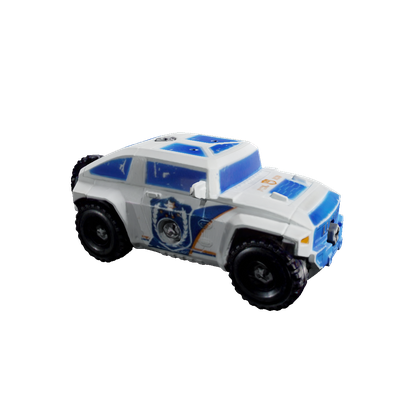} &
        \includegraphics[width=\imgwidth, valign=m]{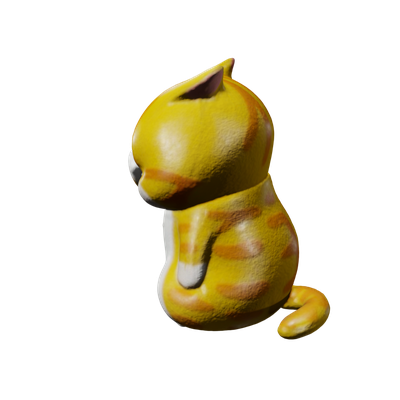} &
        \includegraphics[width=\imgwidth, valign=m]{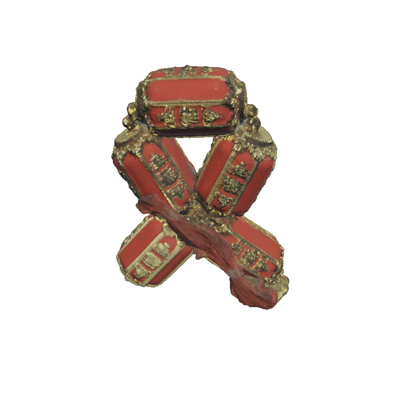} &
        \includegraphics[width=\imgwidth, valign=m]{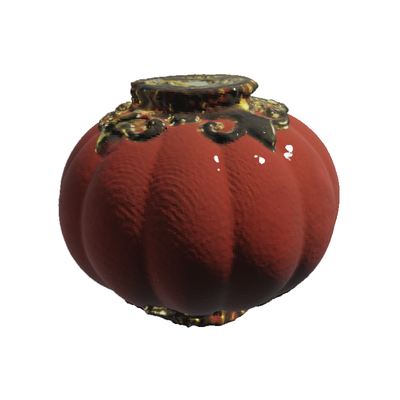} &
        \includegraphics[width=\imgwidth, valign=m]{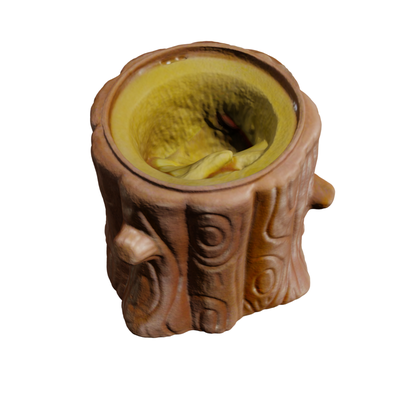} &
        \includegraphics[width=\imgwidth, valign=m]{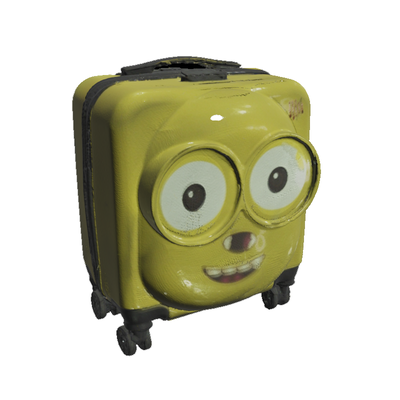} &
        \includegraphics[width=\imgwidth, valign=m]{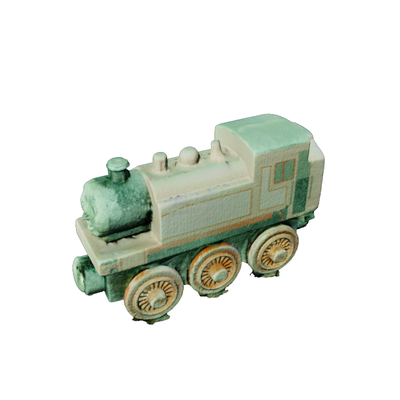}
        \\

        \rotlabel{Raw pts} & 
        \includegraphics[width=\imgwidth, valign=m]{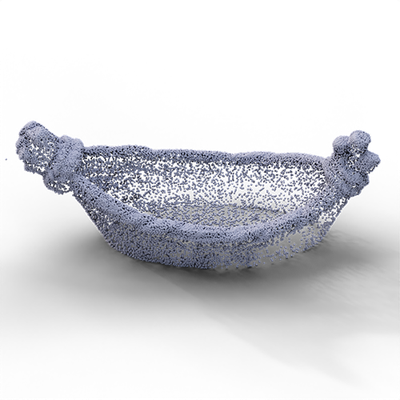} &
        \includegraphics[width=\imgwidth, valign=m]{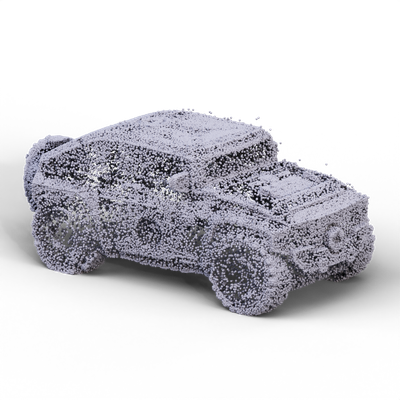} &
        \includegraphics[width=\imgwidth, valign=m]{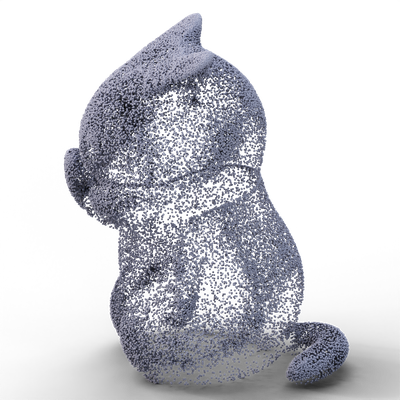} &
        \includegraphics[width=\imgwidth, valign=m]{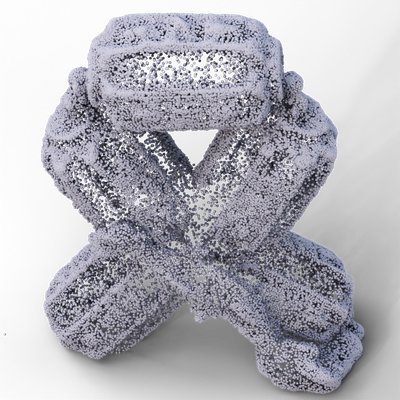} &
        \includegraphics[width=\imgwidth, valign=m]{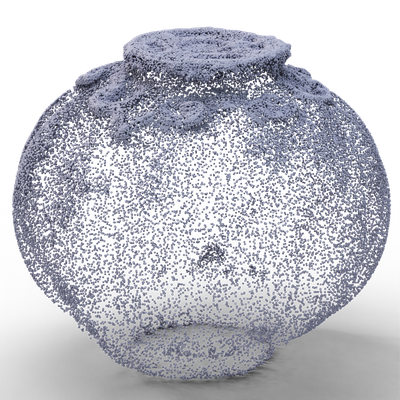} &
        \includegraphics[width=\imgwidth, valign=m]{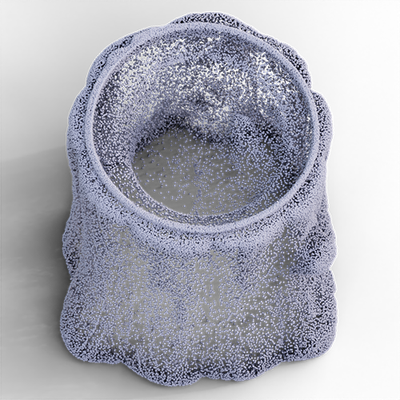} &
        \includegraphics[width=\imgwidth, valign=m]{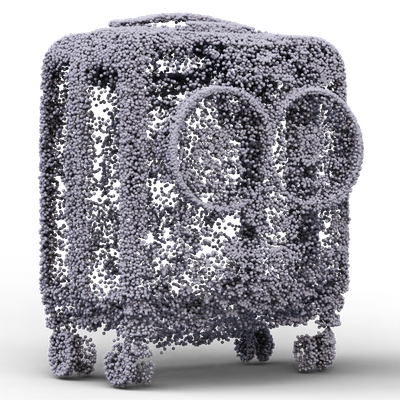} &
        \includegraphics[width=\imgwidth, valign=m]{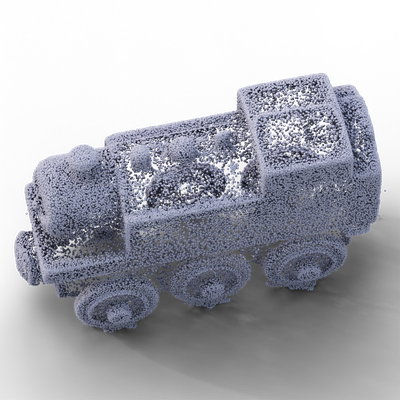}
        \\
        \noalign{\vskip 1mm} 
        & $n=30,384$ & $n=100,199$ & $n=59,237$ & $n=89,462$ & $n=82,640$ & $n=133,111$ & $n=108,774$ & $n=99,008$\\
        & $\widehat{\sigma}=0.0015$ & $\widehat{\sigma}=0.0013$ &
        $\widehat{\sigma}=0.0024$ & $\widehat{\sigma}=0.0011$ &
        $\widehat{\sigma}=0.0012$ & $\widehat{\sigma}=0.0021$ &
        $\widehat{\sigma}=0.0025$ & 
        $\widehat{\sigma}=0.0011$ \\
        & $\widehat{o}=0.17$ & 
        $\widehat{o}=0.13$ & $\widehat{o}=0.16$ &
        $\widehat{o}=0.15$ & $\widehat{o}=0.16$ &
        $\widehat{o}=0.17$ & $\widehat{o}=0.16$ & $\widehat{o}=0.13$ \\
        & $\widehat{u}=0.29$ & 
        $\widehat{u}=0.30$ & $\widehat{u}=0.32$ &
        $\widehat{u}=0.34$ & $\widehat{u}=0.53$ &
        $\widehat{u}=0.41$ & $\widehat{u}=0.38$ & $\widehat{u}=0.31$\\

        \rotlabel{Filtered pts} & 
        \includegraphics[width=\imgwidth, valign=m]{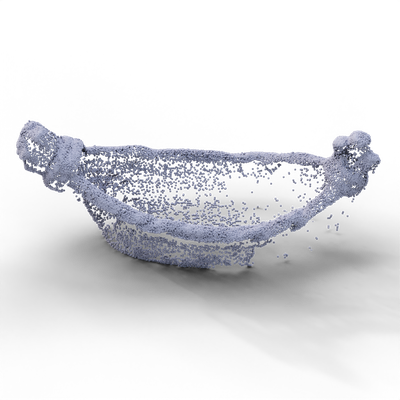} &
        \includegraphics[width=\imgwidth, valign=m]{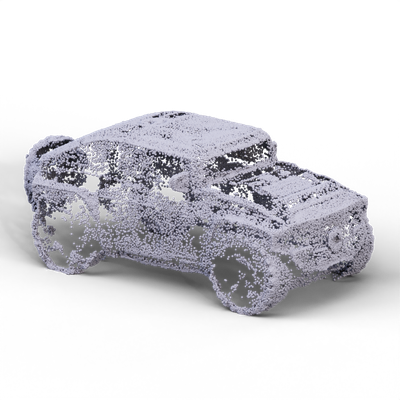} &
        \includegraphics[width=\imgwidth, valign=m]{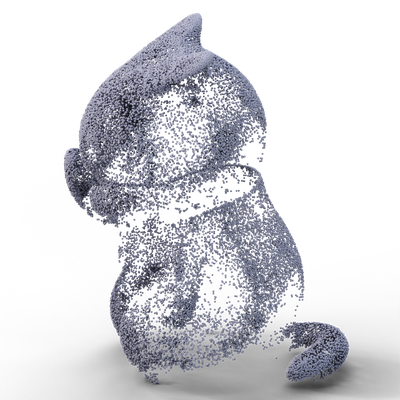} &
        \includegraphics[width=\imgwidth, valign=m]{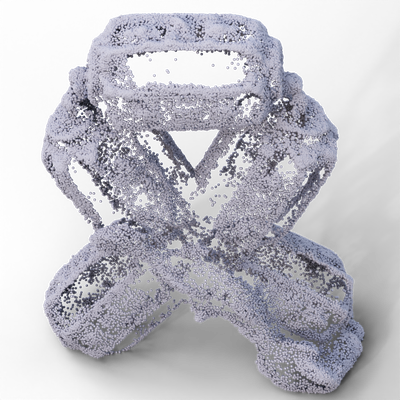} &
        \includegraphics[width=\imgwidth, valign=m]{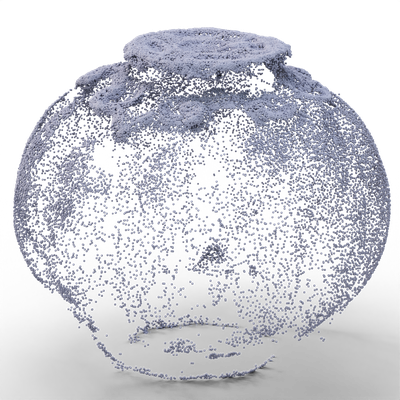} &
        \includegraphics[width=\imgwidth, valign=m]{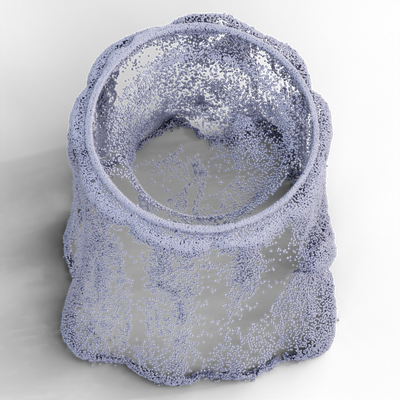} &
        \includegraphics[width=\imgwidth, valign=m]{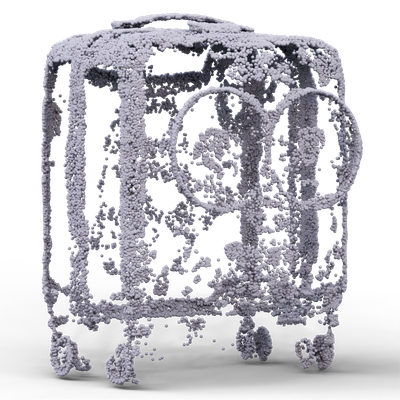} &
        \includegraphics[width=\imgwidth, valign=m]{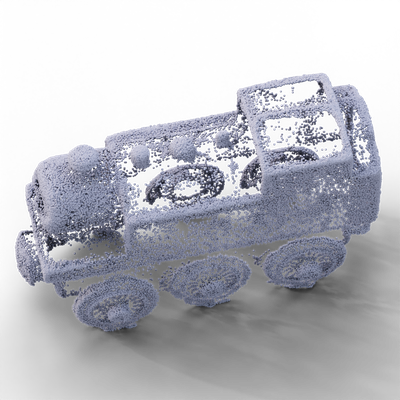}
        \\
        \noalign{\vskip 1mm} 

        \rotlabel{MSP} & 
        \includegraphics[width=\imgwidth, valign=m]{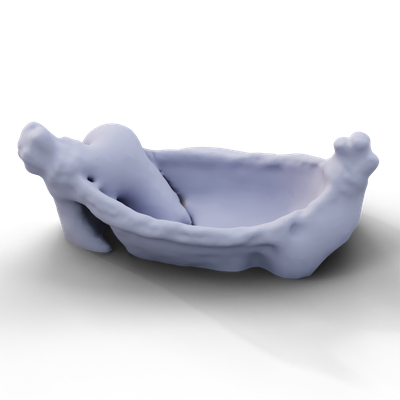} &
        \includegraphics[width=\imgwidth, valign=m]{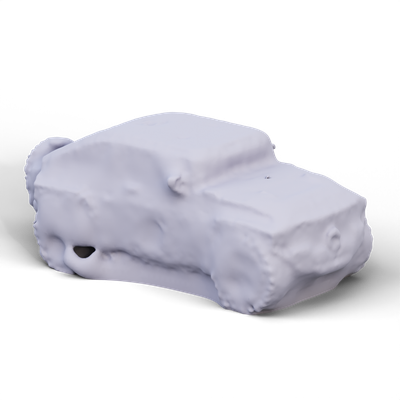} &
        \includegraphics[width=\imgwidth, valign=m]{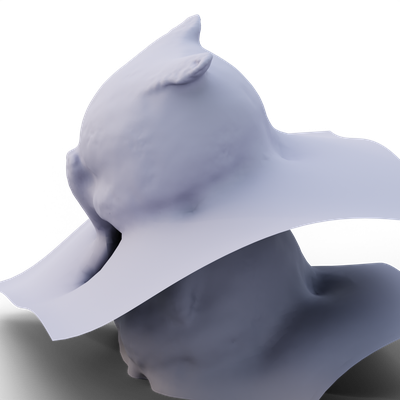} &
        \includegraphics[width=\imgwidth, valign=m]{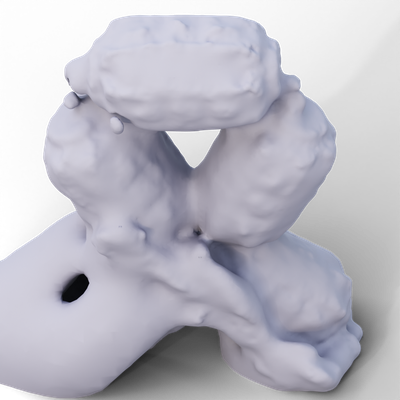} &
        \includegraphics[width=\imgwidth, valign=m]{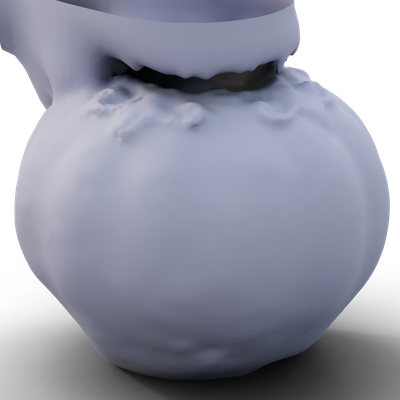} &
        \includegraphics[width=\imgwidth, valign=m]{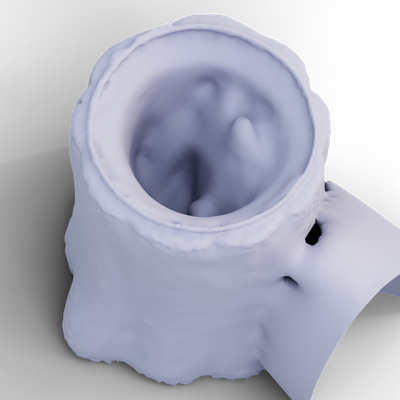} &
        \includegraphics[width=\imgwidth, valign=m]{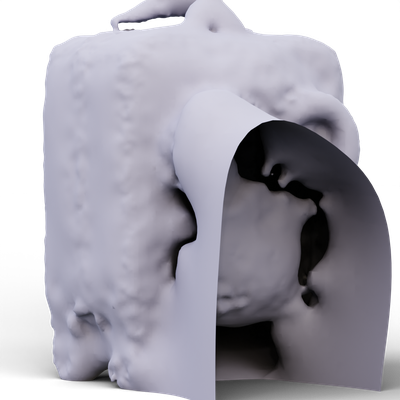} &
        \includegraphics[width=\imgwidth, valign=m]{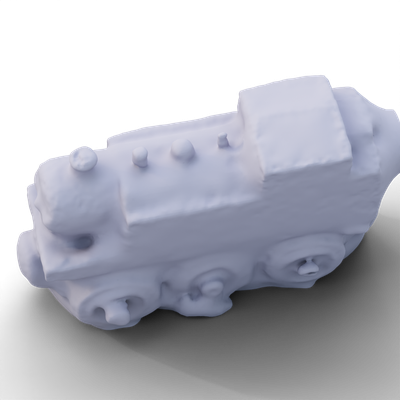}
        \\
        \noalign{\vskip 1mm} 

        \rotlabel{WNNC} & 
        \includegraphics[width=\imgwidth, valign=m]{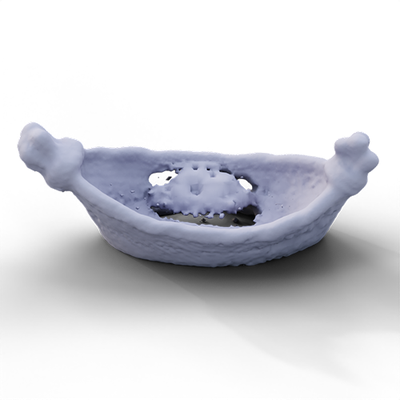} &
        \includegraphics[width=\imgwidth, valign=m]{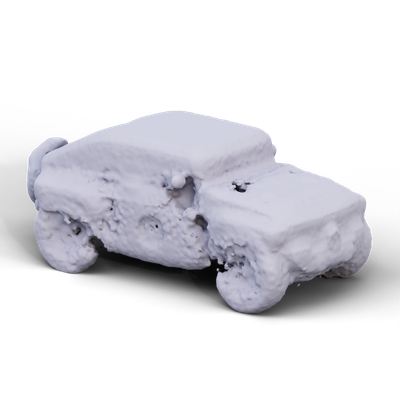} &
        \includegraphics[width=\imgwidth, valign=m]{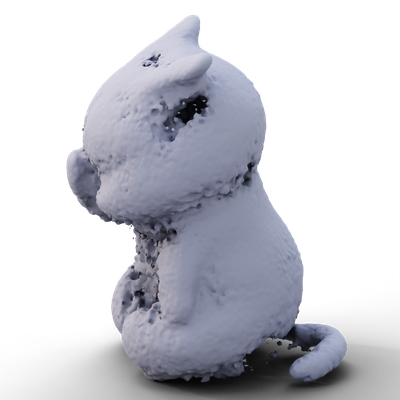} &
        \includegraphics[width=\imgwidth, valign=m]{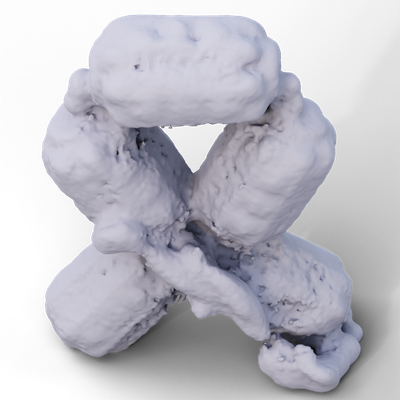} &
        \includegraphics[width=\imgwidth, valign=m]{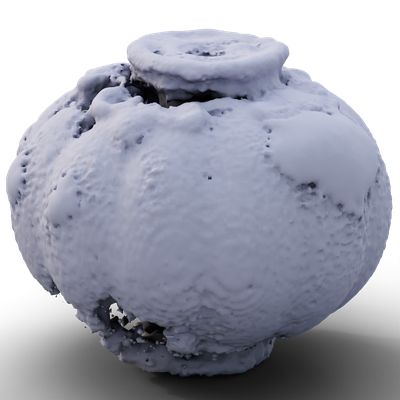} &
        \includegraphics[width=\imgwidth, valign=m]{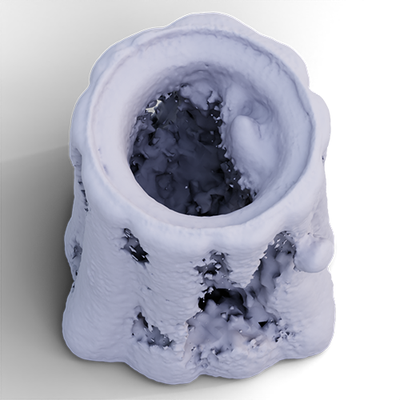} &
        \includegraphics[width=\imgwidth, valign=m]{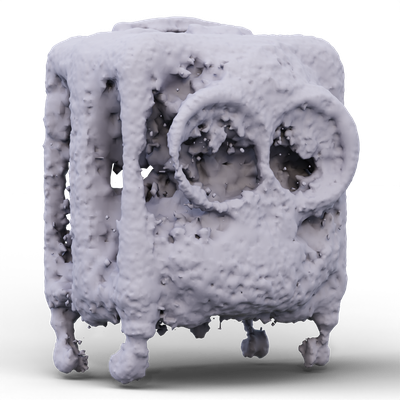} &
        \includegraphics[width=\imgwidth, valign=m]{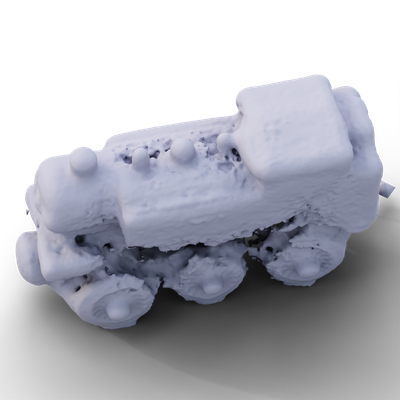}
        \\
        \noalign{\vskip 1mm} 

        \rotlabel{WNNC*} & 
        \includegraphics[width=\imgwidth, valign=m]{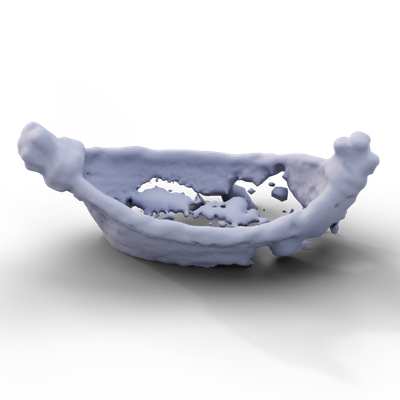} &
        \includegraphics[width=\imgwidth, valign=m]{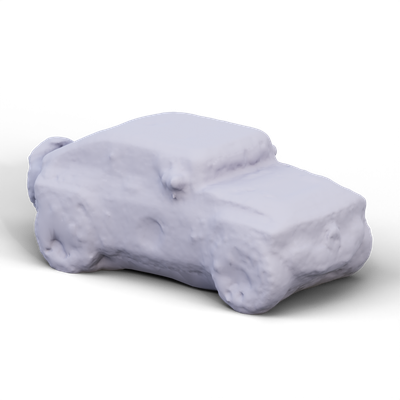} &
        \includegraphics[width=\imgwidth, valign=m]{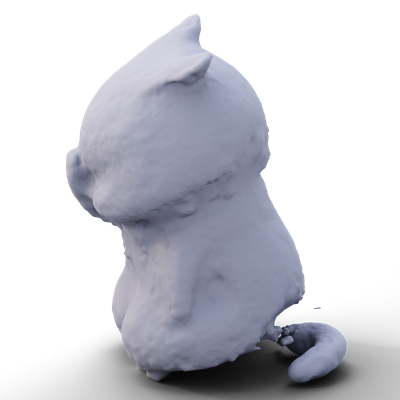} &
        \includegraphics[width=\imgwidth, valign=m]{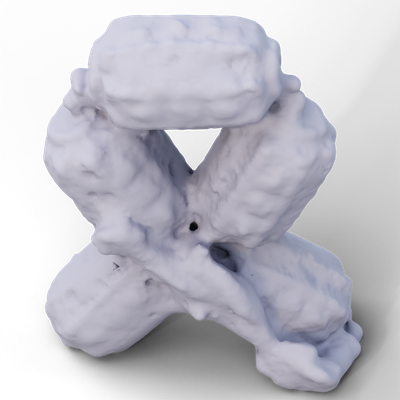} &
        \includegraphics[width=\imgwidth, valign=m]{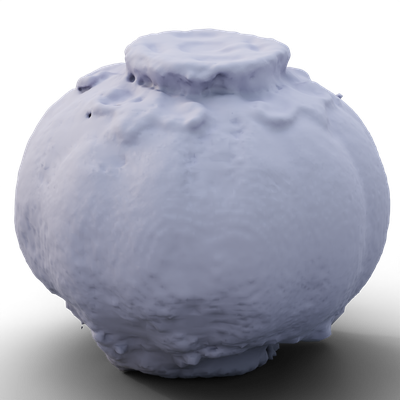} &
        \includegraphics[width=\imgwidth, valign=m]{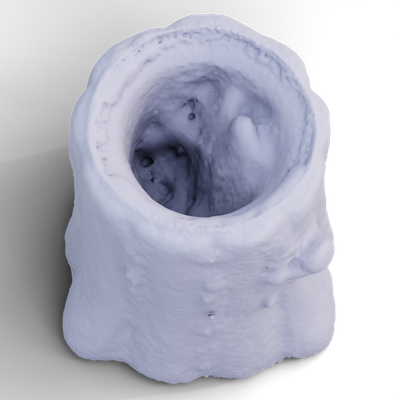} &
        \includegraphics[width=\imgwidth, valign=m]{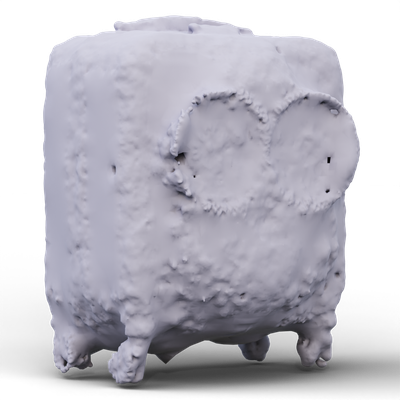} &
        \includegraphics[width=\imgwidth, valign=m]{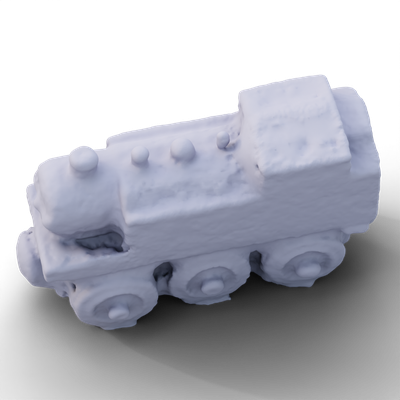}
        \\
        \noalign{\vskip 1mm} 

        \rotlabel{FaCE} & 
        \includegraphics[width=\imgwidth, valign=m]{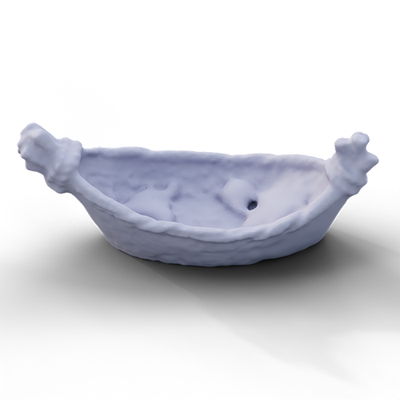} &
        \includegraphics[width=\imgwidth, valign=m]{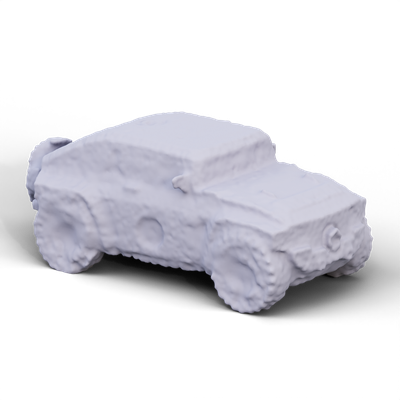} &
        \includegraphics[width=\imgwidth, valign=m]{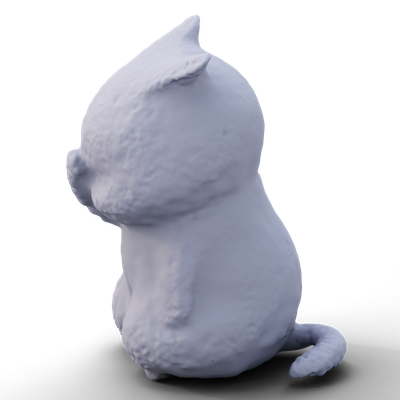} &
        \includegraphics[width=\imgwidth, valign=m]{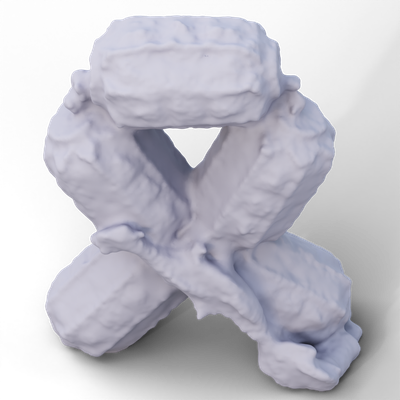} &
        \includegraphics[width=\imgwidth, valign=m]{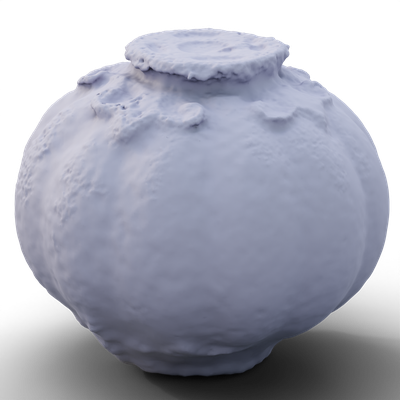} &
        \includegraphics[width=\imgwidth, valign=m]{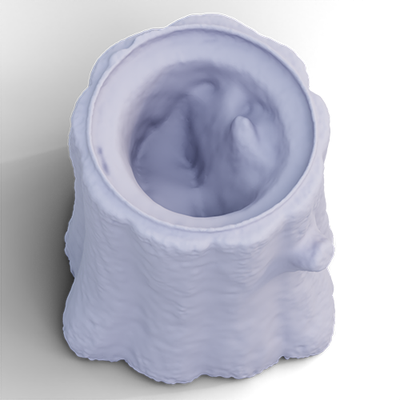} &
        \includegraphics[width=\imgwidth, valign=m]{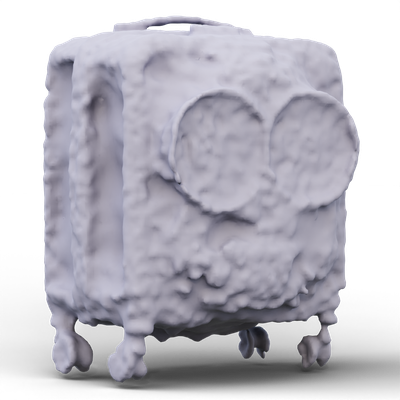} &
        \includegraphics[width=\imgwidth, valign=m]{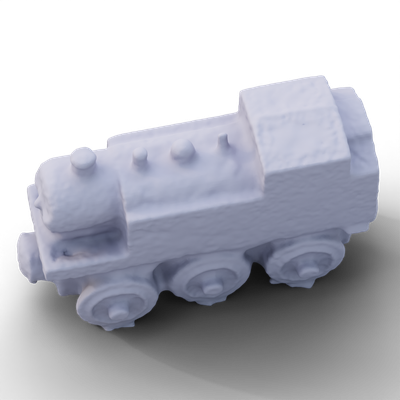}
        \\
        \noalign{\vskip 1mm} 

        \rotlabel{FaCE*} & 
        \includegraphics[width=\imgwidth, valign=m]{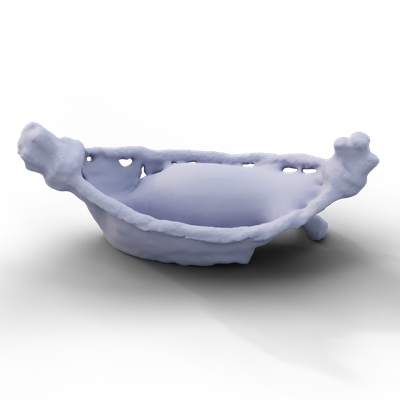} &
        \includegraphics[width=\imgwidth, valign=m]{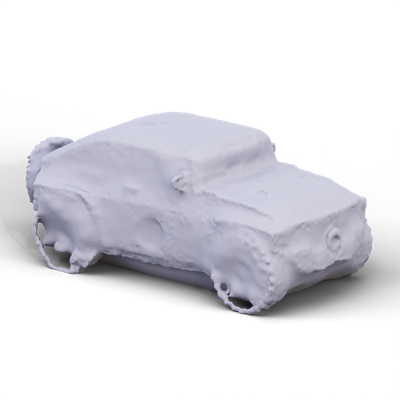} &
        \includegraphics[width=\imgwidth, valign=m]{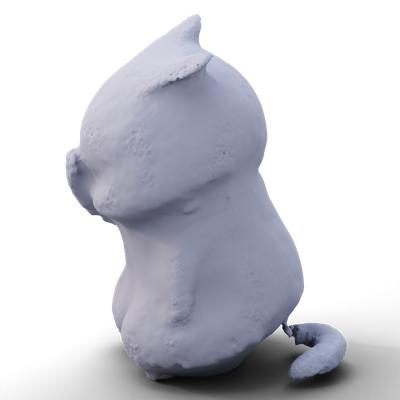} &
        \includegraphics[width=\imgwidth, valign=m]{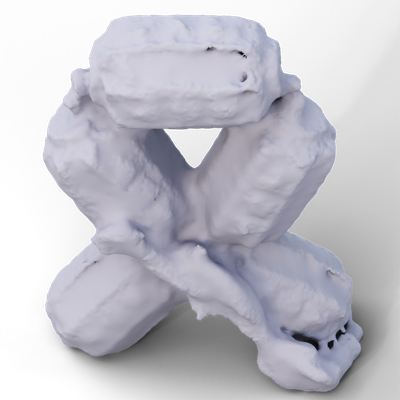} &
        \includegraphics[width=\imgwidth, valign=m]{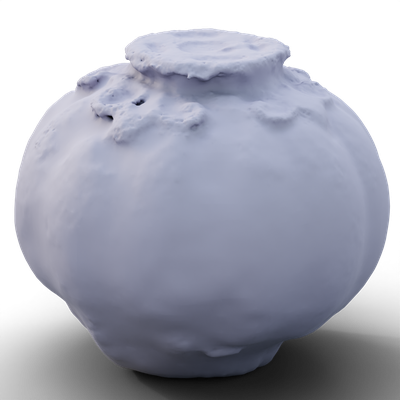} &
        \includegraphics[width=\imgwidth, valign=m]{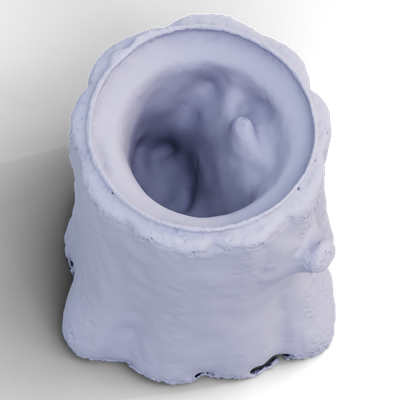} &
        \includegraphics[width=\imgwidth, valign=m]{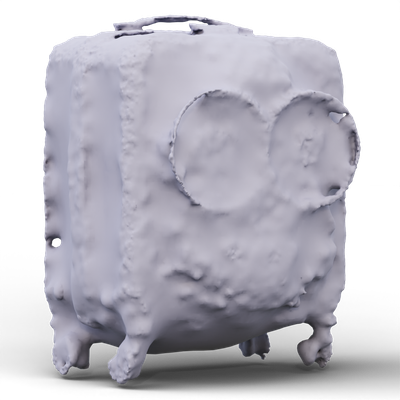} &
        \includegraphics[width=\imgwidth, valign=m]{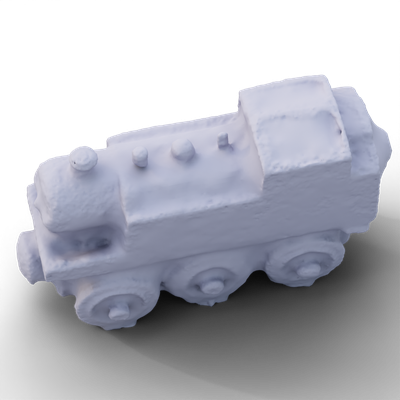}
        \\
        \noalign{\vskip 1mm} 

        \rotlabel{NSH} & 
        \includegraphics[width=\imgwidth, valign=m]{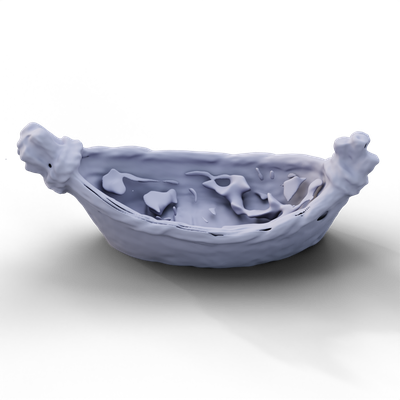} &
        \includegraphics[width=\imgwidth, valign=m]{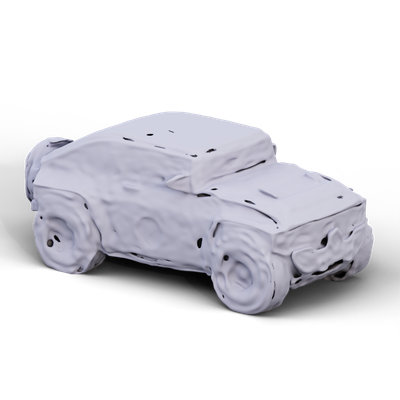} &
        \includegraphics[width=\imgwidth, valign=m]{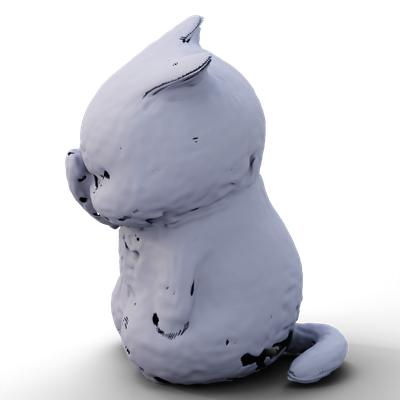} &
        \includegraphics[width=\imgwidth, valign=m]{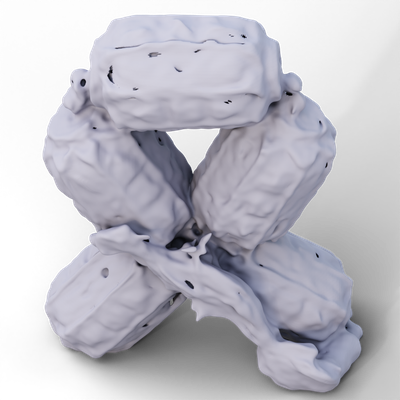} &
        \includegraphics[width=\imgwidth, valign=m]{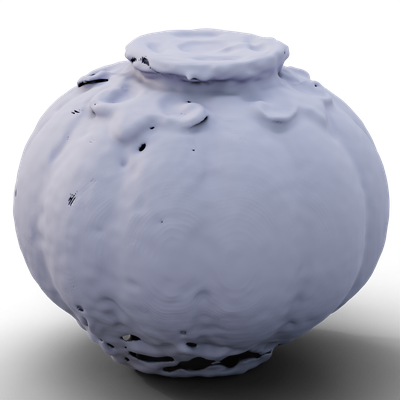} &
        \includegraphics[width=\imgwidth, valign=m]{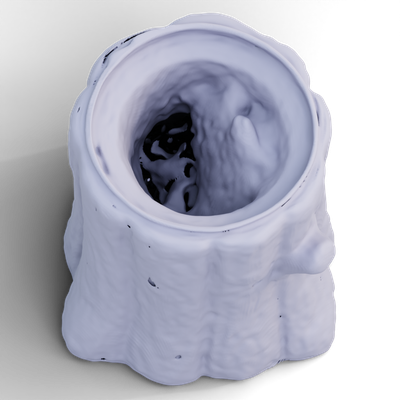} &
        \includegraphics[width=\imgwidth, valign=m]{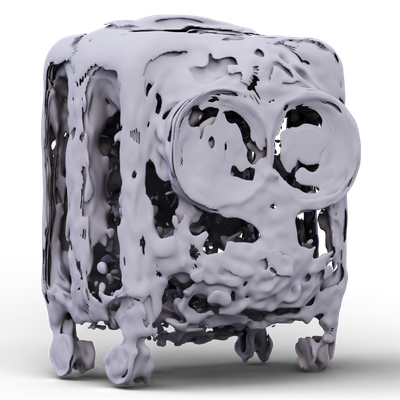} &
        \includegraphics[width=\imgwidth, valign=m]{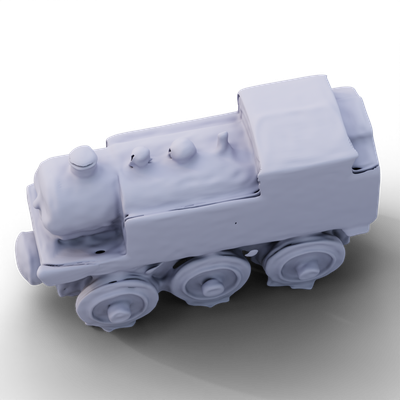}
        \\
        \noalign{\vskip 1mm} 

        \rotlabel{NSH*} & 
        \includegraphics[width=\imgwidth, valign=m]{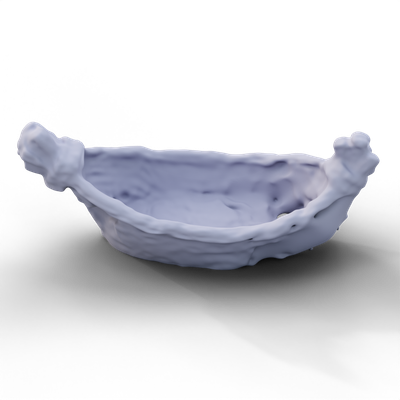} &
        \includegraphics[width=\imgwidth, valign=m]{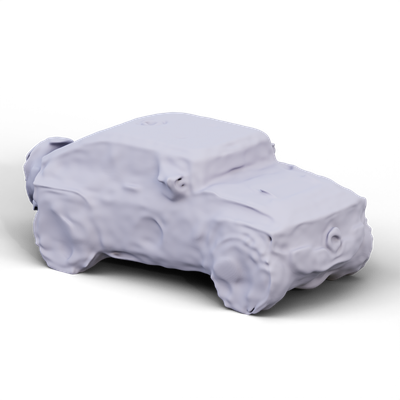} &
        \includegraphics[width=\imgwidth, valign=m]{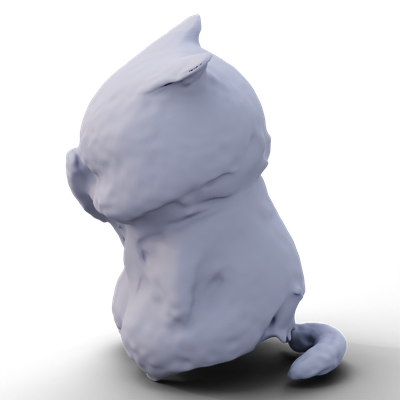} &
        \includegraphics[width=\imgwidth, valign=m]{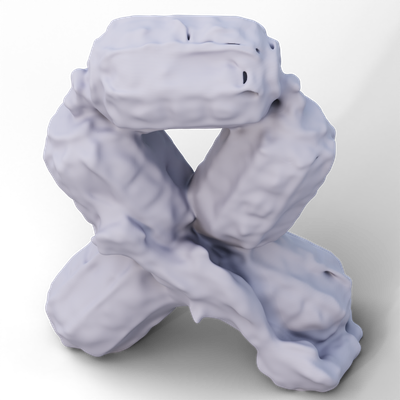} &
        \includegraphics[width=\imgwidth, valign=m]{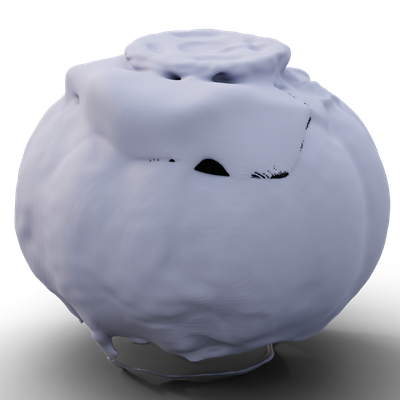} &
        \includegraphics[width=\imgwidth, valign=m]{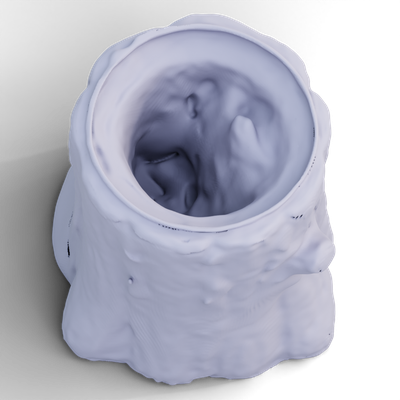} &
        \includegraphics[width=\imgwidth, valign=m]{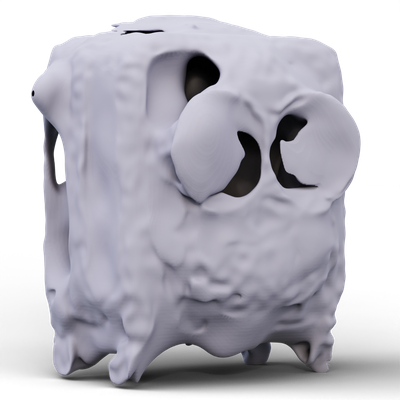} &
        \includegraphics[width=\imgwidth, valign=m]{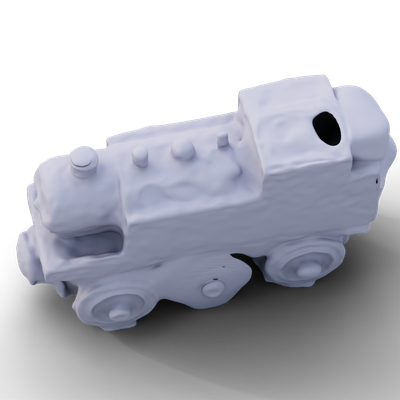}
        \\
        \noalign{\vskip 1mm} 

        \rotlabel{Ours} & 
        \includegraphics[width=\imgwidth, valign=m]{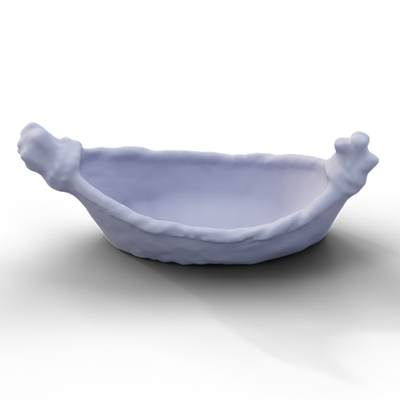} &
        \includegraphics[width=\imgwidth, valign=m]{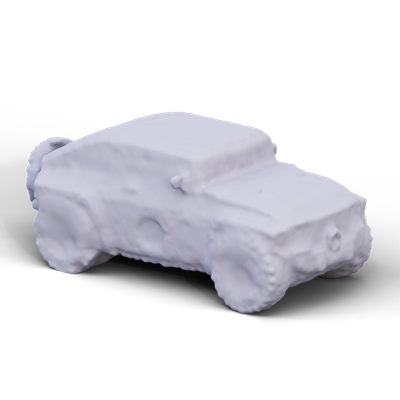} &
        \includegraphics[width=\imgwidth, valign=m]{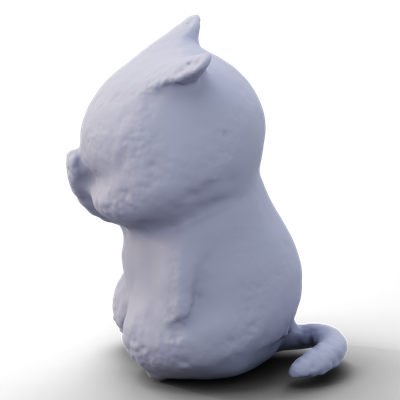} &
        \includegraphics[width=\imgwidth, valign=m]{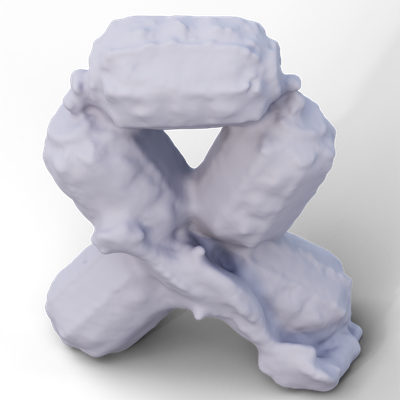} &
        \includegraphics[width=\imgwidth, valign=m]{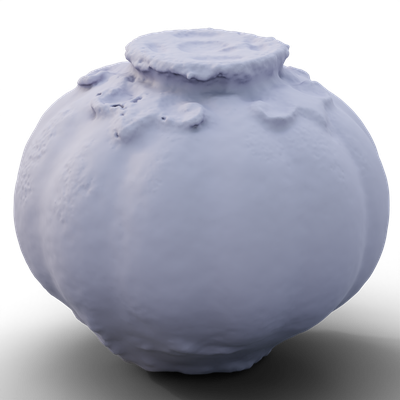} &
        \includegraphics[width=\imgwidth, valign=m]{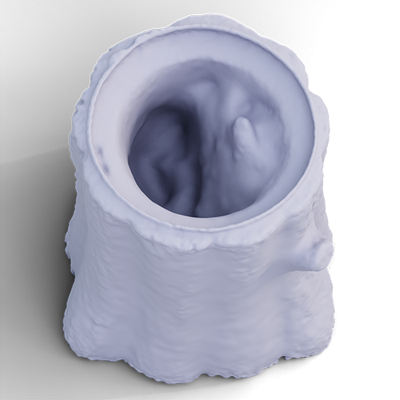} &
        \includegraphics[width=\imgwidth, valign=m]{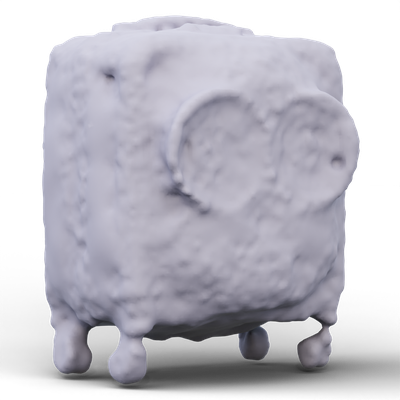} &
        \includegraphics[width=\imgwidth, valign=m]{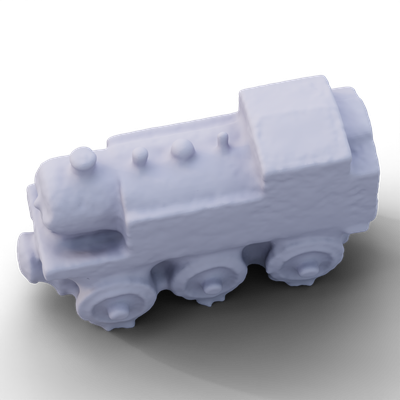}
        \\
        \noalign{\vskip 1mm} 

    \end{tabular}
    \end{small}
    \caption{Results on the 3DGS category. }
    \label{fig:3dgs}
\end{figure*}

\begin{figure*}
    \centering

    \newcommand{\imgwidth}{0.13\linewidth}

    \newcommand{\rotlabel}[1]{%
        \rotatebox{90}{\small\textbf{#1}}%
    }

    \begin{tabular}{cccccccc}

        \rotlabel{Raw pts} & 
        
        \includegraphics[width=\imgwidth, valign=m]{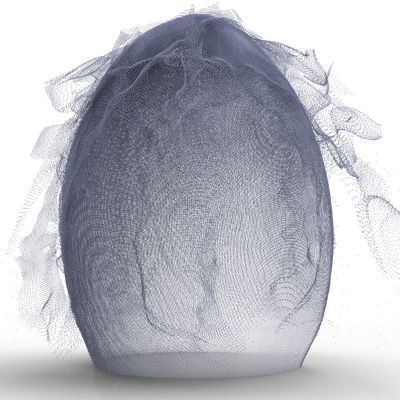} &
        \includegraphics[width=\imgwidth, valign=m]{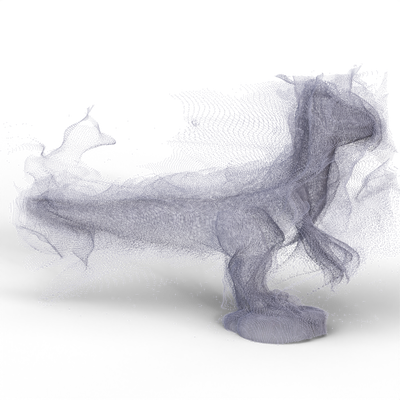} &
        \includegraphics[width=\imgwidth, valign=m]{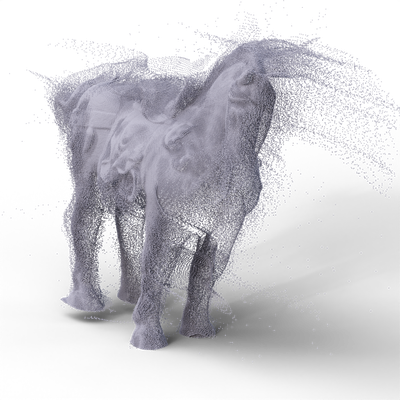} &

        \rotlabel{Filtered pts} & 
        \includegraphics[width=\imgwidth, valign=m]{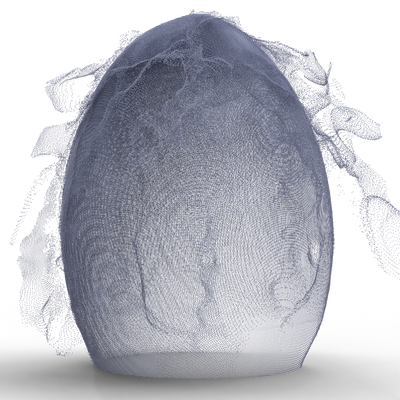} &
        \includegraphics[width=\imgwidth, valign=m]{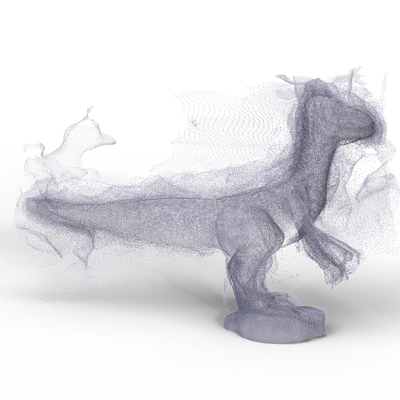} &
        \includegraphics[width=\imgwidth, valign=m]{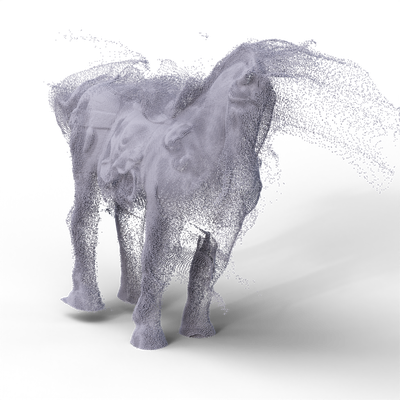} 

        \\

                \noalign{\vskip 1mm} 
        & $n=636,311
$ & $n=298,836
$ & $n=341,870
$  &&&& \\
        & $\widehat{\sigma}=0.00058$ & $\widehat{\sigma}=0.00080$ &
        $\widehat{\sigma}=0.00041$ &&&& \\
        & $\widehat{o}=0.14$ & 
        $\widehat{o}=0.13$ & $\widehat{o}=0.10$  &&&&\\
        & $\widehat{u}=0.21$ & 
        $\widehat{u}=0.21$ & $\widehat{u}=0.33$ &&&&\\

        \rotlabel{Images} &
        \includegraphics[width=\imgwidth, valign=m]{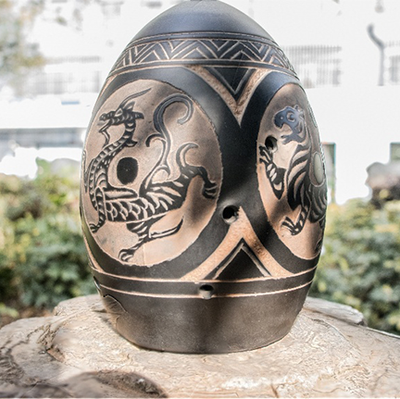} &
        \includegraphics[width=\imgwidth, valign=m]{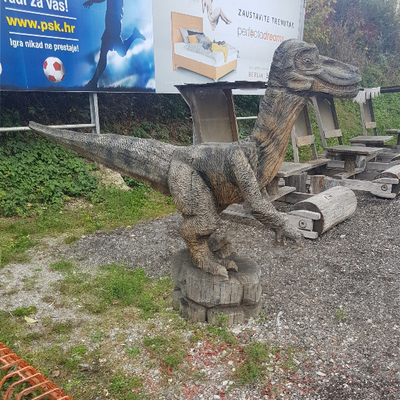} &
        \includegraphics[width=\imgwidth, valign=m]{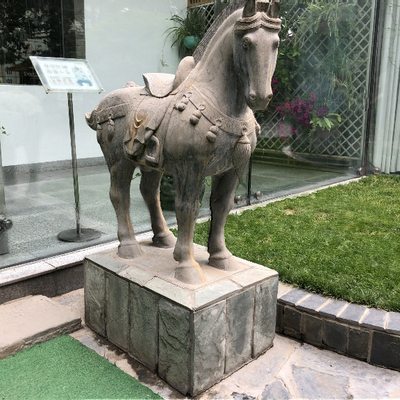} &

        \rotlabel{Ours} & 
        \includegraphics[width=\imgwidth, valign=m]{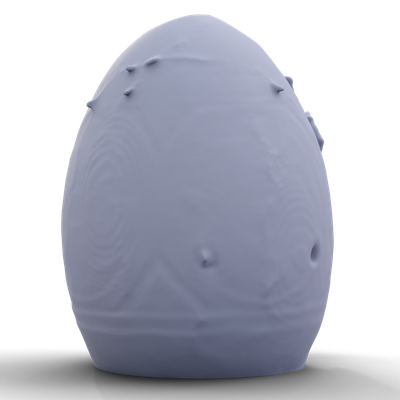} &
        \includegraphics[width=\imgwidth, valign=m]{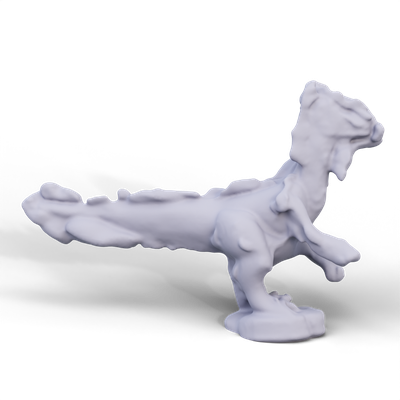} &
        \includegraphics[width=\imgwidth, valign=m]{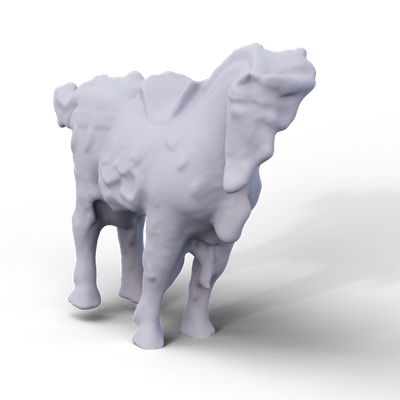} 

        \\

        \rotlabel{WNNC} & 
        \includegraphics[width=\imgwidth, valign=m]{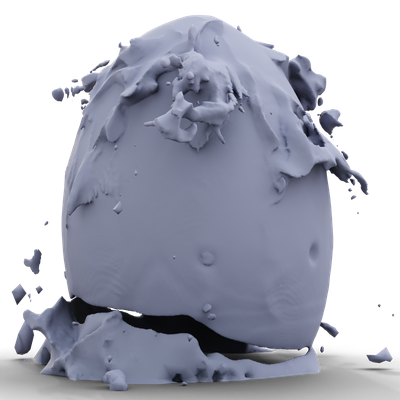} &
        \includegraphics[width=\imgwidth, valign=m]{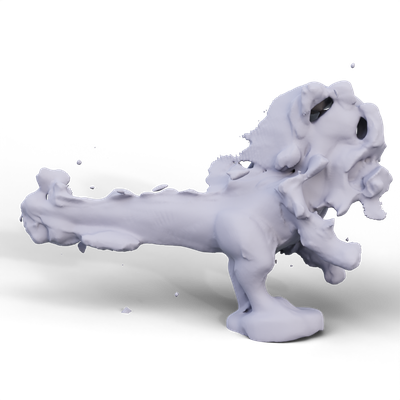} &
        \includegraphics[width=\imgwidth, valign=m]{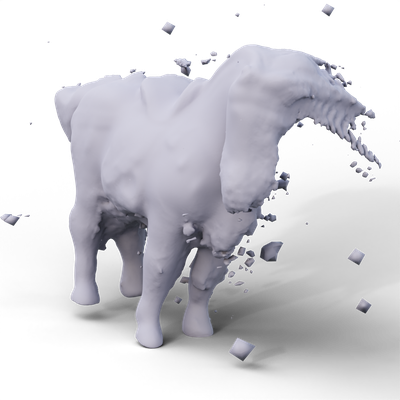} &

        \rotlabel{WNNC*} & 
        \includegraphics[width=\imgwidth, valign=m]{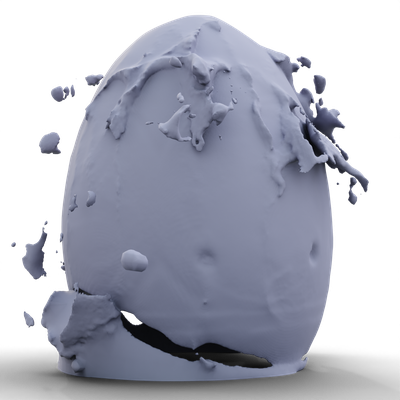} &
        \includegraphics[width=\imgwidth, valign=m]{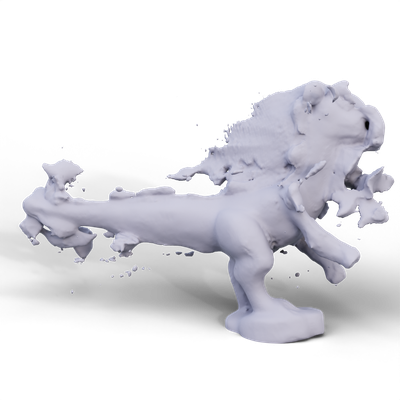} &
        \includegraphics[width=\imgwidth, valign=m]{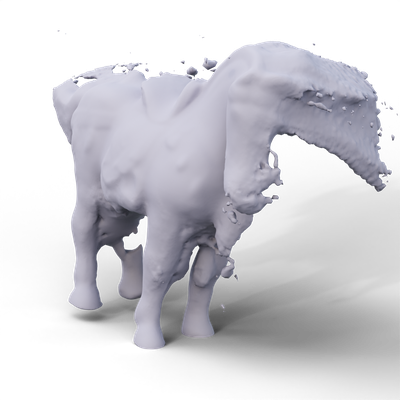} 

        \\

        \rotlabel{FaCE} & 
        \includegraphics[width=\imgwidth, valign=m]{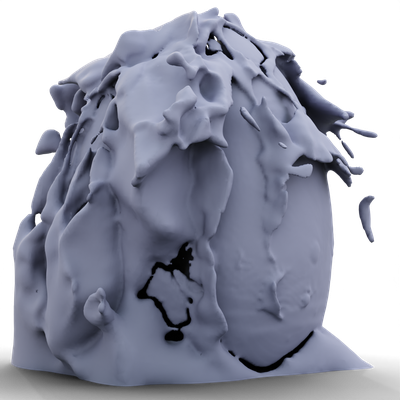} &
        \includegraphics[width=\imgwidth, valign=m]{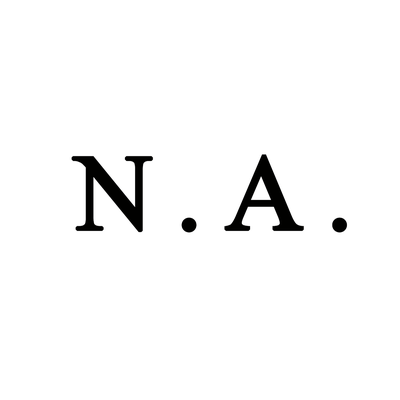} &
        \includegraphics[width=\imgwidth, valign=m]{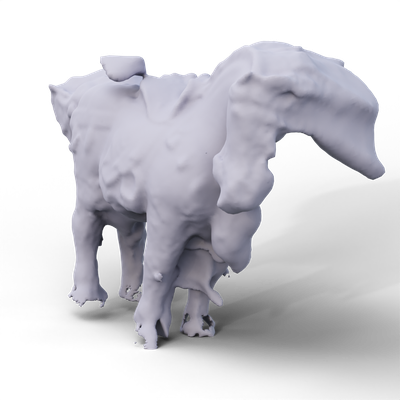} &

        \rotlabel{FaCE*} & 
        \includegraphics[width=\imgwidth, valign=m]{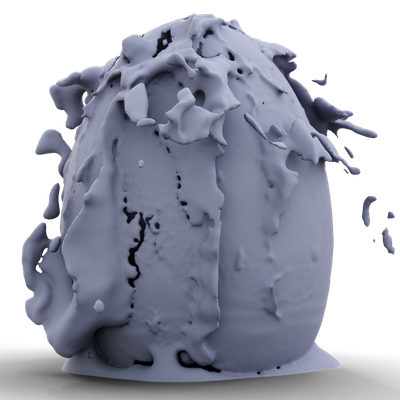} &
        \includegraphics[width=\imgwidth, valign=m]{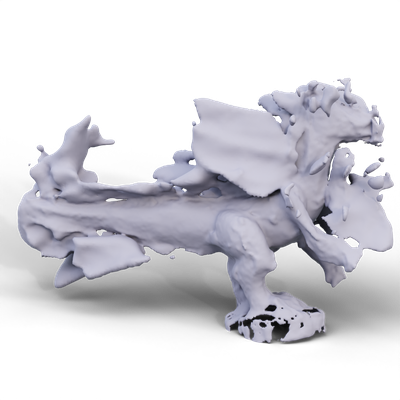} &
        \includegraphics[width=\imgwidth, valign=m]{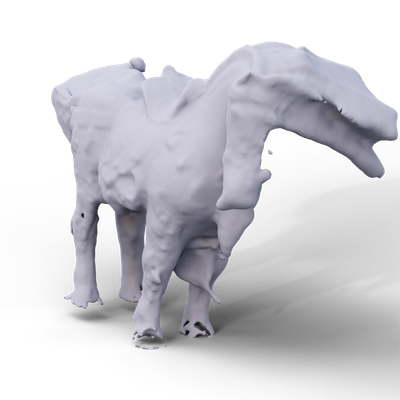}

        \\
        \noalign{\vskip 1mm} 

        \rotlabel{NSH} & 
        \includegraphics[width=\imgwidth, valign=m]{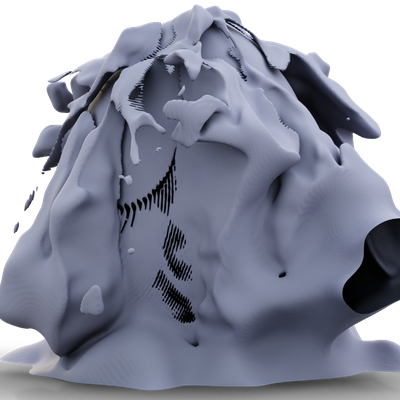} &
        \includegraphics[width=\imgwidth, valign=m]{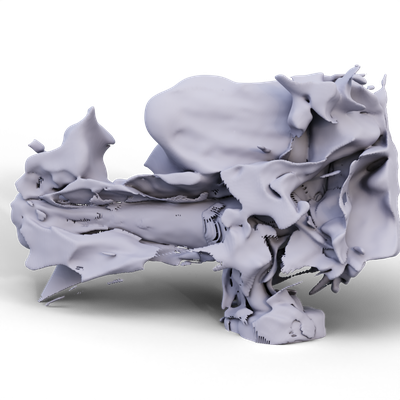} &
        \includegraphics[width=\imgwidth, valign=m]{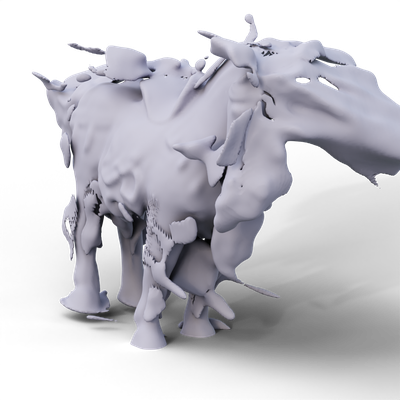} &

        \rotlabel{NSH*} & 
        \includegraphics[width=\imgwidth, valign=m]{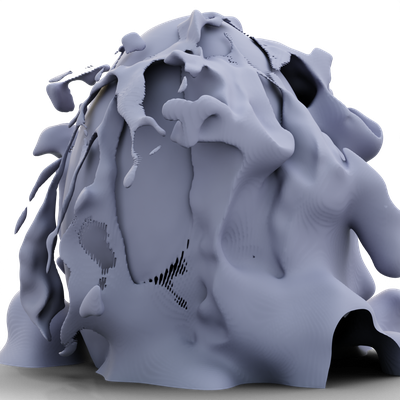} &
        \includegraphics[width=\imgwidth, valign=m]{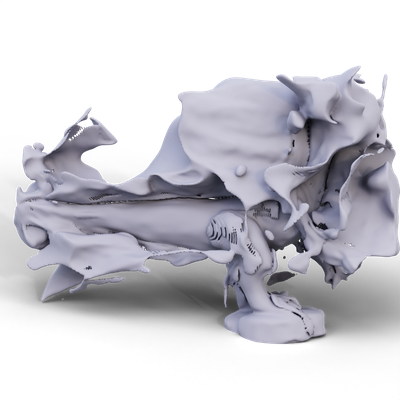} &
        \includegraphics[width=\imgwidth, valign=m]{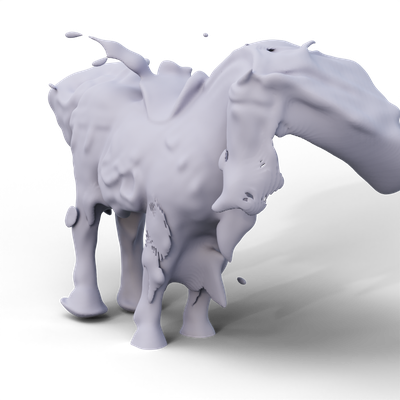} 
        \\
        \noalign{\vskip 1mm}

        \\
        \noalign{\vskip 1mm} 

    \end{tabular}
    
    \caption{Results on the VGGT category. }
    \label{fig:vggt}
\end{figure*}

\begin{figure*}[htbp]
\centering
\setlength{\tabcolsep}{1pt} %

\begin{tabular}{ccccccccccc}

\includegraphics[width=0.09\textwidth]{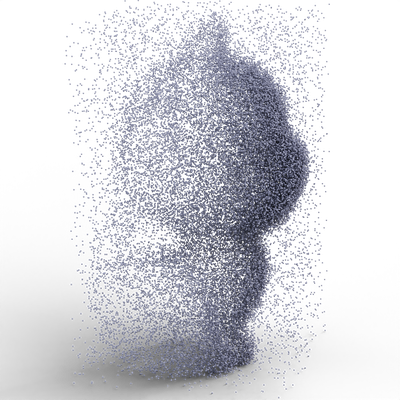} &
\includegraphics[width=0.09\textwidth]{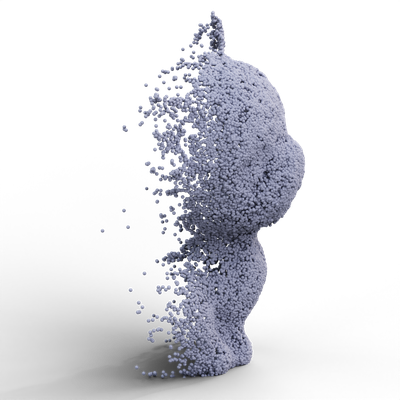} &
\includegraphics[width=0.09\textwidth]{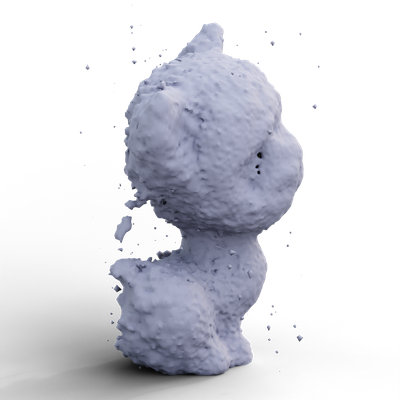} &
\includegraphics[width=0.09\textwidth]{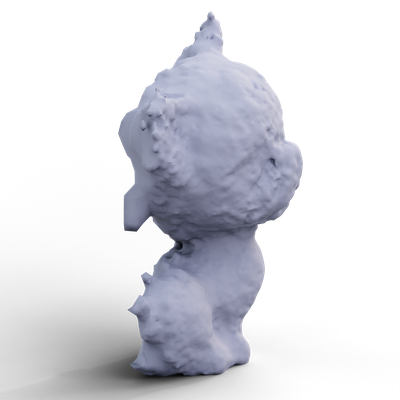} &
\includegraphics[width=0.09\textwidth]{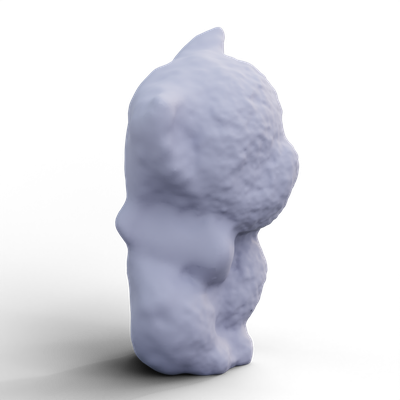} &
\includegraphics[width=0.09\textwidth]{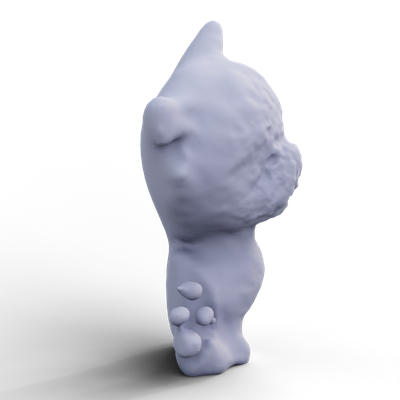} &
\includegraphics[width=0.09\textwidth]{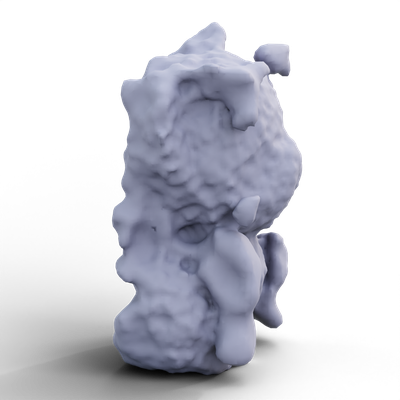} &
\includegraphics[width=0.09\textwidth]{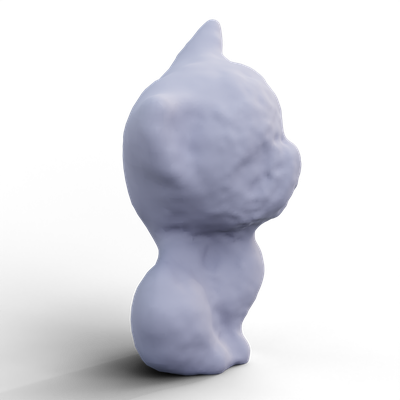} &
\includegraphics[width=0.09\textwidth]{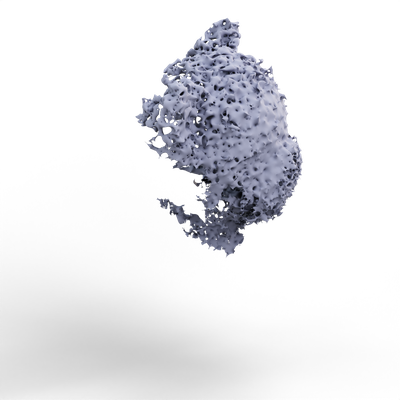} &
\includegraphics[width=0.09\textwidth]{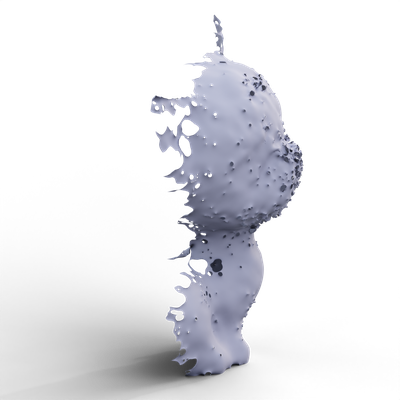} &
\includegraphics[width=0.09\textwidth]{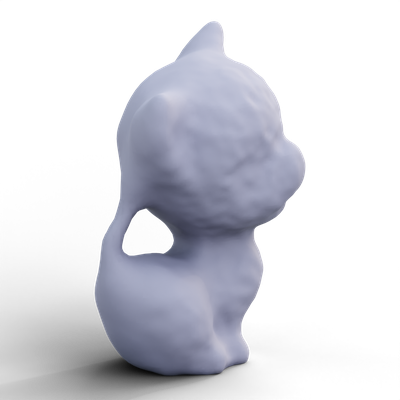} \\

\includegraphics[width=0.09\textwidth]{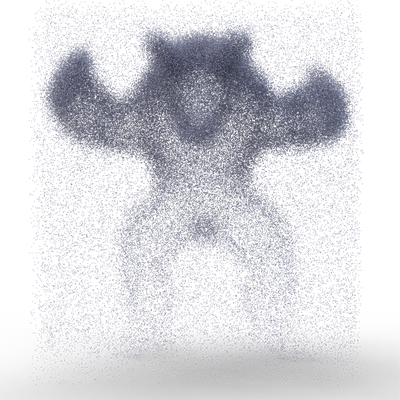} &
\includegraphics[width=0.09\textwidth]{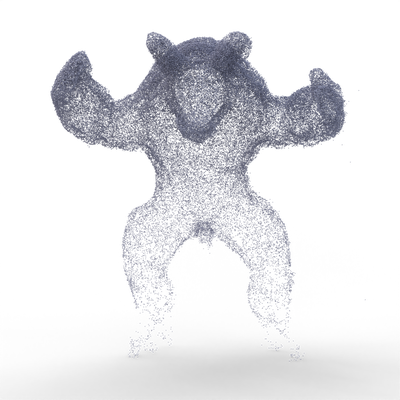} &
\includegraphics[width=0.09\textwidth]{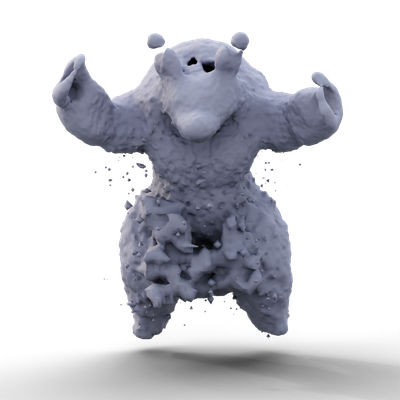} &
\includegraphics[width=0.09\textwidth]{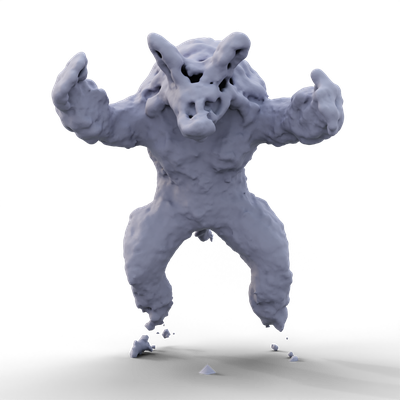} &
\includegraphics[width=0.09\textwidth]{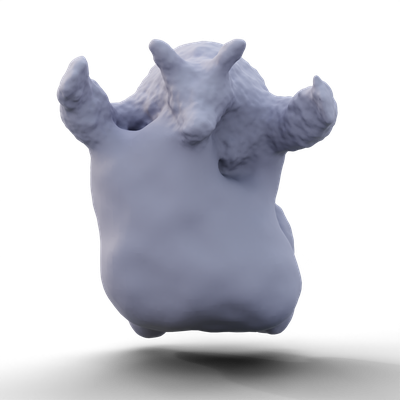} &
\includegraphics[width=0.09\textwidth]{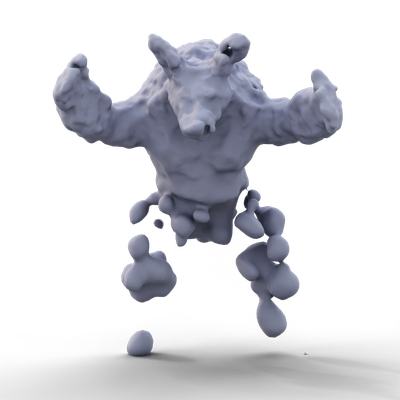} &
\includegraphics[width=0.09\textwidth]{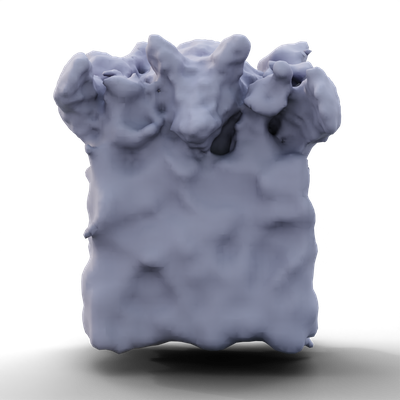} &
\includegraphics[width=0.09\textwidth]{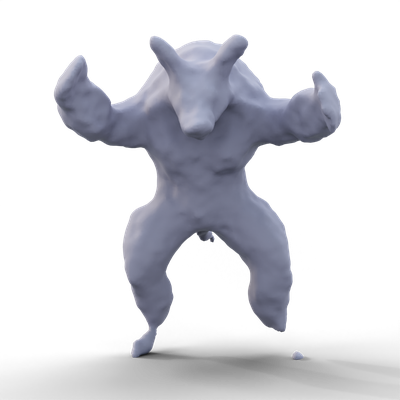} &
\includegraphics[width=0.09\textwidth]{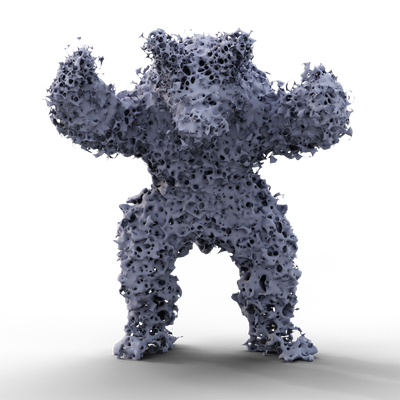} &
\includegraphics[width=0.09\textwidth]{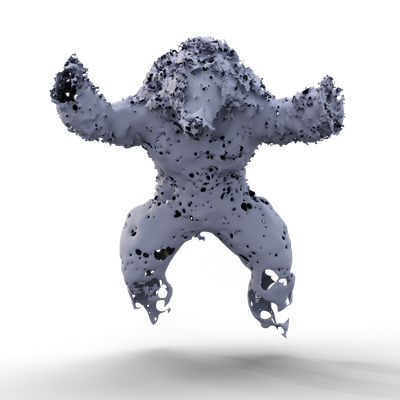} &
\includegraphics[width=0.09\textwidth]{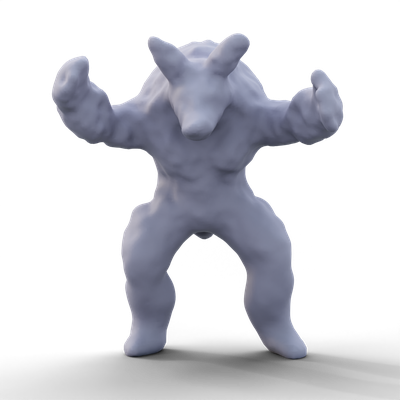} \\
\small Raw pts &
\small Filtered pts &
\small WNNC &
\small WNNC$^\star$ &
\small DWG &
\small DWG$^\star$ &
\small FaCe &
\small FaCe$^\star$ &
\small LoSF &
\small LoSF$^\star$ &
\small Ours 
\end{tabular}

\caption{More results (Top: Kitte$_4$; Bottom: Armadillo$_4$) from common graphics benchmarks degraded with non-uniform sampling, perturbed point positions and injected outliers.}
\label{fig:graphics-benchmark}
\end{figure*}

We provide qualitative and quantitative results for all the test models in the 3DGS and graphics-benchmark categories (see Table~\ref{tab:statistics_all} and Figures~\ref{fig:3dgs} and \ref{fig:graphics-benchmark}). For VGGT point clouds, ground-truth geometry is not available; we therefore report qualitative results only (Figure~\ref{fig:vggt}).

We observe that preprocessing can be overly aggressive under severe corruption. After outlier removal and denoising, the filtered point clouds often contain missing regions or become overly sparse in parts of the shape. Since subsequent orientation and reconstruction stages cannot recover geometry that has been removed, many baselines exhibit a drop in overall performance in terms of NC and CD (due to missing geometry or large distortions), even though the remaining reconstructed regions may appear cleaner locally.

We also observe that LoSF-UDF~\cite{losf} is particularly sensitive to density variation, largely because its inference operates on local patches extracted with a fixed-radius neighborhood. In sparsely sampled regions, a fixed-radius patch may contain only a small number of points (or even be nearly empty), providing insufficient geometric evidence for the pre-trained network to extract meaningful features, from which reliable unsigned distances can be inferred. Consequently, LoSF-UDF tends to produce unstable or incomplete reconstructions in low-density areas, even when it performs well in densely sampled regions.

\section{Limitations}
\label{sec:limitations}

\method{} is built on the generalized winding number field and is therefore primarily suited for reconstructing watertight, manifold surfaces, similar to other GWN-based methods (e.g., GCNO~\cite{xu2023gcno}, BIM~\cite{liu2024bim}, WNNC~\cite{lin2024wnnc}, and DWG~\cite{liu2025diffusing}). Extending the formulation to open surfaces with boundaries or to non-manifold structures remains challenging, especially when the input is heavily affected by non-uniform sampling, noise, and outliers.

Although \method{} improves robustness compared to existing approaches, it can still fail in extreme cases where the point cloud provides too little reliable geometric evidence to guide the winding-field optimization (e.g., very sparse sampling combined with strong noise and a high outlier ratio, like the most challenging cases in our stress tests). Such inputs are inherently ambiguous, and Dirichlet regularization alone is insufficient to recover the correct geometry.

Finally, \method{} regularizes the winding field by minimizing Dirichlet energy, which favors smooth solutions and therefore does not explicitly preserve sharp features or fine-scale details. As a result, CAD-like edges and corners, as well as high-frequency geometric details, may be rounded or attenuated. Since our primary goal is robustness under substantial input corruption, explicitly enforcing feature and detail preservation is challenging when the available geometric evidence near such structures is sparse or unreliable.

\end{document}